\documentclass[preprint,12pt]{elsarticle}

\usepackage{amssymb}
\usepackage{amsmath}

\usepackage{lineno}
\usepackage[utf8]{inputenc} 
\usepackage[T5]{fontenc}    
\usepackage[table]{xcolor}
\usepackage{multirow}
\usepackage{caption}
\usepackage{graphicx}
\usepackage{subcaption}
\usepackage[most]{tcolorbox} 
\usepackage{booktabs}
\usepackage{array}
\usepackage{lipsum}
\usepackage{enumitem}        
\usepackage{float} 

\usepackage{booktabs}    
\usepackage{threeparttable}  
\usepackage{siunitx}     
\usepackage{graphicx}
\journal{Information Fusion}
\usepackage{subcaption}

\usepackage{algorithm}
\usepackage{algpseudocode}
\usepackage{algorithmicx}

\usepackage{algorithm}
\usepackage{algpseudocode}
\usepackage{xcolor}

\usepackage{tcolorbox}        
\usepackage{hyperref}         
\usepackage{cleveref}         

\newcommand{\pos}[1]{{\scriptsize\,\textcolor{teal!80!black}{$\uparrow$#1}}}
\newcommand{\nega}[1]{{\scriptsize\,\textcolor{red!70!black}{$\downarrow$#1}}}

\usepackage{booktabs, bm}
\newcommand{\cmark}{\checkmark}
\newcommand{\xmark}{---}

\algrenewcommand\algorithmiccomment[1]{%
  \hskip1em{\color{gray}\small\textit{// #1}}%
}

\usepackage{titletoc}   
\usepackage{hyperref}   

\usepackage{booktabs, colortbl, xcolor}
\definecolor{ifblue50}{RGB}{230,241,251}
\definecolor{ifblue800}{RGB}{12,68,124}
\definecolor{ifgray50}{RGB}{241,239,232}
\definecolor{ifgray600}{RGB}{95,94,90}
\definecolor{ifgreen50}{RGB}{234,243,222}
\definecolor{ifgreen800}{RGB}{59,109,17}
\definecolor{ifamber50}{RGB}{250,238,218}
\definecolor{ifamber800}{RGB}{133,79,11}

\begin{document}

\begin{frontmatter}



\title{Linguistically Informed Multimodal Fusion for Vietnamese 
Scene-Text Image Captioning: Dataset, Graph Framework, 
and Phonological Attention}


\author[1,2,3]{Nhi Ngoc-Yen Nguyen}
\ead{21521231@gm.uit.edu.vn}

\author[1,2,3]{Anh-Duc Nguyen}
\ead{21520140@gm.uit.edu.vn}

\author[1,2,3]{Nghia Hieu Nguyen}
\ead{nghiangh@uit.edu.vn}

\author[1,2,3]{Kiet Van Nguyen}
\ead{kietnv@uit.edu.vn}

\author[1,2,3]{Ngan Luu-Thuy Nguyen}
\ead{ngannlt@uit.edu.vn}

\affiliation[1]{organization={Faculty of Information Science and Engineering},
}

\affiliation[2]{organization={University of Information Technology}}
\affiliation[3]{
    organization={Vietnam National University},
    city={Ho Chi Minh},
    country={Viet Nam}
}


\begin{abstract}
Scene-text image captioning requires fusing three information
streams---visual features, OCR-detected text, and linguistic
knowledge---to generate descriptions that faithfully integrate text
visible in images. Existing fusion approaches treat text as
language-agnostic, which fails for Vietnamese: a tonal language
where diacritics alter word meaning, OCR errors are pervasive, and
word boundaries are ambiguous. We argue that Vietnamese scene-text
captioning demands \textit{linguistically informed multimodal
fusion}, where language-specific structural knowledge is explicitly
incorporated into the fusion mechanism. Motivated from these insights, we  propose \textbf{HSTFG} (Heterogeneous Scene-Text Fusion
Graph), a general-purpose graph fusion framework with learned
spatial attention bias, and show through topology analysis that
cross-modal graph edges are harmful for scene-text fusion.
Building on this finding, we design \textbf{PhonoSTFG}
(Phonological Scene-Text Fusion Graph) which specializes
graph-level fusion for Vietnamese linguistic reasoning. To support evaluation, we introduce \textbf{ViTextCaps}, the first large-scale Vietnamese scene-text captioning dataset (\textbf{15{,}729} images with \textbf{74{,}970} captions), with comprehensive linguistic analysis showing that 52.8\% of the vocabulary is at risk of diacritic collision. 
\end{abstract}








\begin{keyword}
multimodal fusion \sep
scene-text captioning \sep
Vietnamese NLP \sep
phonological attention \sep
graph neural networks \sep
information fusion
\end{keyword}

\end{frontmatter}



\section{Introduction}
\label{sec:introduction}
A street sign reading \textit{bán đất} (land for sale) and
\textit{bàn đất} (soil table) differ by a single diacritical
mark---yet this distinction, trivial for a human reader, is
systematically mishandled by state-of-the-art multimodal fusion
systems. This failure exemplifies a broader problem:
\textbf{scene-text image captioning} requires fusing three distinct
information streams---visual features describing objects and layouts,
OCR-detected text tokens, and linguistic knowledge governing how to
select and integrate text into fluent natural language---and existing
fusion mechanisms are not equipped to handle the orthographic
complexity of tonal languages~\cite{sidorov2020textcaps,wang2021ocr}.

Existing multimodal fusion approaches for scene-text captioning
treat the text modality as language-agnostic: OCR tokens are encoded
via static word embeddings (e.g., FastText) and fused with visual
features through cross-modal attention or copy
mechanisms~\cite{hu2020iterative,wang2021ocr}. This
\textbf{language-agnostic fusion} is adequate for English, where
OCR errors are predominantly character substitutions with limited
semantic impact. However, it fundamentally fails for
\textbf{tonal languages} such as Vietnamese, where:

\begin{enumerate}
    \item \textbf{Diacritic-sensitive fusion is essential.}
    Vietnamese encodes six lexical tones via diacritical marks that
    completely change word meaning
    (\textit{ma} = ghost, \textit{m\`{a}} = but,
    \textit{m\'{a}} = mother). OCR systems frequently confuse these
    visually similar marks, producing tokens that are
    \textit{orthographically close but semantically distant}. A
    fusion mechanism that treats OCR tokens as opaque symbols
    cannot detect or correct these errors---it needs explicit
    phonological awareness. Our analysis of ViTextCaps reveals
    that 52.8\% of the caption vocabulary is at risk of diacritic
    collision, and one in five caption--OCR token pairs disagree
    on diacritics even at high OCR confidence.

    \item \textbf{Spatial reasoning over OCR tokens is critical.}
    Vietnamese scene text frequently appears in multi-block layouts,
    with 74.3\% of ViTextCaps images containing text in multiple
    spatially separated regions. A fusion mechanism must reason
    about geometric relationships between OCR tokens to determine
    which text blocks are thematically related and which should be
    selected for the caption.

    \item \textbf{Bilingual fusion is the norm.} Vietnamese signage
    frequently mixes Vietnamese and English text (brand names,
    loanwords), with 42.9\% of ViTextCaps images containing both
    languages. The fusion system must handle code-switching between
    two linguistically distinct sources and apply language-specific
    reasoning selectively.
\end{enumerate}

These challenges motivate a new paradigm: \textbf{linguistically
informed multimodal fusion}, where language-specific structural
knowledge is explicitly incorporated into the fusion mechanism
rather than being implicitly learned from data alone. To this end, we make two contributions:

\begin{enumerate}[nosep]
    \item \textbf{HSTFG and PhonoSTFG: a principled progression from
    general to linguistically informed fusion.}
    We propose \textit{HSTFG}, a heterogeneous scene-text fusion
    graph with learned spatial attention bias across configurable
    edge types. Building on this module, we propose \textit{PhonoSTFG},
    which inherits HSTFG's T$\to$T-only topology and specializes
    it for Vietnamese linguistic reasoning.

    \item \textbf{ViTextCaps: the first Vietnamese scene-text
    captioning benchmark.} We introduce a large-scale dataset
    comprising 15{,}729 images with 74{,}970 human-annotated
    captions ($\sim$4.77 captions per image), accompanied by
    comprehensive linguistic analysis covering tonal ambiguity
    (52.8\% vocabulary collision rate), OCR error taxonomy,
    code-mixing patterns, word segmentation, and syntactic
    dependency structure.

\end{enumerate}


The remainder of this paper is organized as follows. Section~\ref{sec:related_work} surveys multimodal information fusion, graph-based fusion, scene-text captioning, and linguistic knowledge integration, positioning our work within the broader fusion literature. Section~\ref{sec:dataset} introduces ViTextCaps, describing the data collection pipeline, annotation protocol, and quality control procedures. Section~\ref{sec:dataset_analysis} presents a comprehensive linguistic analysis of ViTextCaps—covering diacritic collision, OCR error taxonomy, code-mixing, and text selection behavior—establishing the Vietnamese-specific challenges that motivate our fusion designs. Section~\ref{sec:proposed_method} presents HSTFG and PhonoSTFG: we first develop HSTFG as a general heterogeneous graph fusion framework and derive from its topology analysis the harmful role of cross-modal edges, then build PhonoSTFG as a linguistically specialized extension with dual-stream gated fusion and Vietnamese phonological attention bias. Section~\ref{sec:benchmark_setup} describes the benchmark setup, including evaluation metrics, comparison baselines, feature extraction, and the word segmentation standard adopted for ViTextCaps. Section~\ref{sec:experiments} reports experimental results: both models significantly outperform existing baselines ($p < 0.01$, paired bootstrap), with stratified analyses identifying the conditions under which each model excels and characterizing the fluency--fidelity trade-off between task-specific fusion and general-purpose LMMs. Finally, Section~\ref{sec:conclusion} concludes with broader implications for linguistically informed multimodal fusion and directions for future work.

\section{Related Work}
\label{sec:related_work}

\subsection{Scene-Text Understanding Datasets}
\label{sec:related_datasets}

Scene-text understanding benchmarks have evolved from recognition
tasks toward higher-level semantic understanding.
English captioning datasets---TextCaps~\cite{sidorov2020textcaps}
(28{,}408 images) and the web-scale
OCR-CC~\cite{wang2021ocr} (1.4M images)---established the core task
of integrating OCR tokens into natural language descriptions, but
target a non-tonal language where diacritic errors are absent by
definition.
More recently, MTVQA~\cite{tang2024mtvqa} introduced the first
multilingual text-centric VQA benchmark with human expert annotations
across nine languages including Vietnamese, confirming that
multilingual scene-text understanding remains an open problem---yet it
addresses only VQA and provides no mechanism for the tonal
orthographic challenges inherent to Vietnamese captioning.
Vietnamese scene-text understanding has been addressed through VQA
benchmarks: ViTextVQA~\cite{vitextvqa},
ViOCRVQA~\cite{viocrvqa},
ViSignVQA~\cite{visignvqa}, and
OpenViVQA~\cite{tran2023openvivqa}, the latter published in this
journal.
However, VQA differs fundamentally from captioning: the question
specifies which text to reference, bypassing the
\textit{text selection problem}---the challenge of autonomously
determining which OCR tokens are relevant without question guidance.
For image captioning, UIT-ViIC~\cite{lam2020uitviic} and
UIT-OpenViIC~\cite{openviic} provided Vietnamese captioning
benchmarks but contain no scene text.

Table~\ref{tab:dataset_comparison} positions ViTextCaps within
this landscape. The \textit{Scene Text} column reveals the central
gap: no existing resource combines Vietnamese with both the
captioning task and scene-text annotation.
Beyond filling this gap, ViTextCaps is also the largest
Vietnamese captioning dataset by annotation count (74,970 captions)
and the only one that requires models to read, select, and fuse
OCR tokens without question guidance---making it a fundamentally
more demanding benchmark than existing Vietnamese VQA or
captioning resources.

\begin{table*}[htbp]
\centering
\caption{Comparison of ViTextCaps with existing scene-text
understanding datasets. The \textit{Scene Text} column marks
whether the dataset contains scene-text annotations alongside
captions or QA pairs.
$^\dagger$Estimated from published statistics.
$^\ddagger$Covers nine languages including Vietnamese; VQA only.
$^\S$Focuses primarily on book-cover images.}
\label{tab:dataset_comparison}
\resizebox{\textwidth}{!}{
\small
\setlength{\tabcolsep}{9pt}
\renewcommand{\arraystretch}{1.05}
\begin{tabular}{l c r r c c c}
\toprule
\textbf{Dataset}
  & \textbf{Lang.}
  & \textbf{Images}
  & \textbf{Captions / QA}
  & \textbf{Task}
  & \textbf{Ann/Img}
  & \textbf{Scene Text} \\
\midrule

\multicolumn{7}{l}{\textit{English scene-text captioning}} \\[1pt]
TextCaps~\cite{sidorov2020textcaps}
  & EN & 28{,}408 & 142{,}040 & Captioning & 5.0 & \cmark \\
OCR-CC~\cite{wang2021ocr}
  & EN & 1.4M & 1.4M & Captioning & 1.0 & \cmark \\

\midrule
\multicolumn{7}{l}{\textit{Multilingual scene-text VQA}} \\[1pt]
MTVQA$^\ddagger$~\cite{tang2024mtvqa}
  & Multi & 8{,}794 & 28{,}607 & VQA & 3.3 & \cmark \\

\midrule
\multicolumn{7}{l}{\textit{Vietnamese general captioning}} \\[1pt]
UIT-ViIC~\cite{lam2020uitviic}
  & VI & 3{,}850 & 19{,}250 & Captioning & 5.0 & \xmark \\
UIT-OpenViIC~\cite{openviic}
  & VI & 13{,}100 & 61{,}241 & Captioning & 4.7 & \xmark \\

\midrule
\multicolumn{7}{l}{\textit{Vietnamese scene-text VQA}} \\[1pt]
ViTextVQA~\cite{vitextvqa}
  & VI & 16{,}762 & 50{,}342 & VQA & 3.0 & \cmark \\
ViOCRVQA$^\S$~\cite{viocrvqa}
  & VI & 28{,}282 & 123{,}781 & VQA & 4.4 & \cmark \\
ViSignVQA~\cite{visignvqa}
  & VI & 10{,}762 & 25{,}573 & VQA & 2.4 & \cmark \\
OpenViVQA~\cite{tran2023openvivqa}
  & VI & 11{,}199 & 37{,}914 & VQA & 3.4 & \cmark \\

\midrule
\multicolumn{7}{l}{\textit{Vietnamese scene-text captioning}} \\[1pt]
\textbf{ViTextCaps (ours)}
  & \textbf{VI}
  & \textbf{15{,}729}
  & \textbf{74{,}970}
  & \textbf{Captioning}
  & \textbf{4.77}
  & $\bm{\cmark}$ \\
\bottomrule
\end{tabular}
}
\end{table*}

\subsection{Multimodal Information Fusion}
\label{sec:related_fusion}

Multimodal fusion---combining information from heterogeneous sources
to support downstream prediction---has been studied across a wide
range of settings~\cite{baltrusaitis2019multimodal,
zhang2020multimodal, atrey2010multimodal, lahat2015multimodal}.
The fundamental challenge is not merely aggregating signals, but
deciding \emph{when}, \emph{how}, and \emph{at what granularity}
different modalities should interact~\cite{lahat2015multimodal}.
Fusion methods are broadly categorised by the stage at which
information is combined.
\textbf{Early fusion} concatenates raw features from different
modalities before processing, preserving fine-grained interactions
but requiring the model to learn cross-modal alignment from
scratch~\cite{ngiam2011multimodal}.
\textbf{Late fusion} processes each modality independently and
combines predictions at the decision level, offering modularity
but sacrificing inter-modal
interactions~\cite{snoek2005early}.
\textbf{Attention-based fusion} dynamically weights information
from different sources and has become the dominant paradigm in
vision-language
models~\cite{vaswani2017attention, lu2019vilbert, tan2019lxmert}.
\textbf{Gated fusion} mechanisms employ learnable gates to control
information flow between modalities, enabling adaptive fusion that
responds to input quality~\cite{arevalo2017gated, wu2021gated}.

Surveys in this journal have systematically characterised these
paradigms: Zhang et al.~\cite{zhang_if_vqa_2019} distinguish
attention-based, bilinear-pooling, and memory-augmented approaches
for visual question answering; Gandhi et
al.~\cite{gandhi2023multimodal} provide a systematic review of
multimodal fusion methods across modalities and applications; and
Gkoumas et al.~\cite{gkoumas2021difference} offer an empirical
comparison of fusion strategies for multimodal language analysis,
demonstrating that the optimal fusion design is
task-dependent---a finding that motivates empirical topology
validation rather than adopting fully connected structures by
default. More recent work in this journal---including relational
reasoning combined with bilinear
attention~\cite{zhang_if_relational_2020} and co-attention
mechanisms for fine-grained multimodal
alignment~\cite{zhang_if_coattention_2021}---has shown that
jointly modelling pairwise relations and attentive feature
selection yields finer multimodal representations than either
mechanism alone.

A critical limitation shared by all these fusion paradigms is
their \textbf{language-agnosticism}: text tokens are treated as
generic vectors regardless of language-specific orthographic
structure. This agnosticism is unproblematic for languages where
OCR errors are predominantly character substitutions with limited
semantic impact, but systematically fails for tonal languages
where diacritics encode lexical distinctions that cannot be
recovered without explicit phonological knowledge.

\subsection{Graph-Based Multimodal Fusion}
\label{sec:related_graph}

Graph neural networks (GNNs) provide a natural framework for
modelling structured relationships among heterogeneous entities
in multimodal
settings~\cite{yao2018exploring, yang2019auto}.
For scene-text tasks, heterogeneous graphs distinguish visual
object nodes (V-nodes) from OCR token nodes (T-nodes) to enable
modality-specific reasoning.
\citet{mm-gnn} proposed MM-GNN, constructing three sub-graphs
for visual, semantic, and numeric modalities with attention-based
cross-graph aggregators for scene-text VQA.
\citet{wang2021ocr} extended this to scene-text captioning via
graph convolutions over visual and textual nodes.
Graph Transformer Networks~\cite{yun2019graph} allow heterogeneous
edge types to carry relational semantics during message passing.
Deep multimodal reasoning networks published in this
journal~\cite{dmrfnet2021} demonstrate that graph-based
multi-relational reasoning over visual objects yields more
discriminative representations than flat attention, through
explicit modelling of spatial and semantic relations between
detected regions. A deeper theoretical concern in all GNN-based
approaches is the risk of over-smoothing, where stacking multiple
graph convolution layers causes node representations to converge
toward uniform vectors~\cite{li2018deeper}; shallow designs
with residual connections are the standard mitigation.

Prior graph-based fusion work has uniformly assumed that richer
cross-modal connectivity improves performance. However, this
assumption has not been empirically validated for the specific
case of scene-text captioning, where a dedicated copy-mechanism
decoder already handles cross-modal fusion at the decoding
stage---creating potential architectural redundancy that motivates
systematic topology analysis.

\subsection{Scene-Text Image Captioning}
\label{sec:related_captioning}

Scene-text captioning requires models to both read and semantically
integrate text visible in images, extending standard
encoder-decoder
captioning~\cite{vinyals2015show, xu2015show, butd}.
TextCaps~\cite{sidorov2020textcaps} introduced the first
large-scale English benchmark, accompanied by
M4C~\cite{hu2020iterative}, a multimodal multi-copy mechanism
that iteratively attends to OCR tokens and copies them into the
output via a pointer network.
Subsequent works improved text integration through spatial
layout-aware attention~\cite{mma-sr} and pre-training objectives
for joint visual--textual grounding~\cite{yang2021tap}.
Meshed-Memory Transformer~\cite{m2} and
Attention-on-Attention~\cite{huang2019aoa} improved
encoder-decoder interactions for standard image captioning and
have been adopted as decoder backbones in scene-text systems.
Context-aware question embedding fusion~\cite{contextvqa_if2023}
and feature-level co-attention approaches published in this
journal further show that explicitly modelling multi-level
relationships between text and visual queries yields more
selective and accurate content grounding.

Large vision-language models have more recently demonstrated
scene-text reading capabilities through emergent OCR skills and
massive pre-training.
GPT-4o~\cite{openai2024gpt4o} represents the current frontier
for multilingual scene-text captioning, yet even at this scale,
the model lacks the explicit pointer-copy mechanism required for
selective OCR token integration---generating fluent text that
does not faithfully reproduce the specific scene text visible
in the image.
This \textit{fluency--fidelity gap} between general-purpose and
task-specific models remains an open problem that standard
captioning metrics are not designed to characterise.

From a fusion perspective, both traditional and large-model
approaches rely on a \textit{single-stream OCR representation}
(e.g., FastText), providing no mechanism for fusing multiple
views of the same text token. When the token representation
itself is unreliable---as is the case with Vietnamese diacritic
errors---single-stream encoding propagates the error directly
into the fusion stage without any opportunity for correction.

\subsection{Vietnamese Vision-Language Understanding}
\label{sec:related_vietnamese}

Vietnamese NLP has benefited from pre-trained language models
such as PhoBERT~\cite{nguyen2020phobert} and
ViT5~\cite{phan2022vit5}.
In the vision-language space,
OpenViVQA~\cite{tran2023openvivqa}---published in this
journal---introduced the first large-scale open-ended Vietnamese
VQA dataset together with novel multimodal fusion architectures
for answer generation in this low-resource setting.
Vietnamese scene-text understanding has been addressed through
VQA benchmarks~\cite{vitextvqa, viocrvqa, visignvqa} and
document QA~\cite{receiptvqa}. More recently, Vietnamese
multimodal large language models have adapted general
vision-language architectures to Vietnamese visual understanding;
however, these models are not designed for scene-text captioning
and lack the pointer-copy mechanism required for selective OCR
token integration.

A consistent limitation across all Vietnamese vision-language
work is the absence of language-specific fusion adaptations:
cross-attention and feature concatenation are applied uniformly
without accounting for Vietnamese's tonal orthography, diacritic
sensitivity, or syllable-level tokenization. The scene-text
captioning setting---where these properties directly affect both
OCR output quality and caption generation---has not previously
been studied.

\subsection{Linguistic Knowledge in Multimodal Fusion}
\label{sec:related_linguistic}

Incorporating linguistic structure into multimodal systems has
been explored across several modalities and tasks.
\citet{li2020universal} used phonological features for
cross-lingual speech processing.
\citet{banchs-etal-2015-report} leveraged phonological similarity
for transliteration.
In NLP, character-level and subword representations capture
morphological structure
implicitly~\cite{bojanowski2017enriching}.
Relation networks for object
detection~\cite{hu2018relation} encode pairwise geometric
relations between region proposals as explicit attention
biases---a design philosophy we adapt for phonological rather
than geometric token relations.
Knowledge-guided attention has been shown effective in related
multimodal tasks: explicit domain knowledge encoded as structural
bias in attention mechanisms improves feature selection in
settings where the input signal is noisy or
ambiguous~\cite{skeafn2023, dmrfnet2021}.

Despite this breadth, \textbf{explicit phonological knowledge
has never been incorporated into vision-language fusion}.
Existing approaches rely either on implicit knowledge from
pre-trained language models or on language-agnostic string
similarity features such as edit distance and character overlap,
neither of which captures the structured relationship between
diacritically related token pairs.
The specific challenges of tonal orthography---where sub-pixel
diacritical marks determine lexical identity and OCR errors are
concentrated in the diacritical dimension---remain unaddressed
in any multimodal fusion framework.
This gap is particularly acute in scene-text captioning for
tonal languages, where standard fusion mechanisms propagate
rather than correct diacritic errors, and where the absence of
phonological awareness leads to systematic semantic distortion
in the generated output.

\section{ViTextCaps: Vietnamese Scene-Text Captioning Dataset}
\label{sec:dataset}
\begin{figure}[ht] 
    \centering 
    \includegraphics[width=\textwidth]{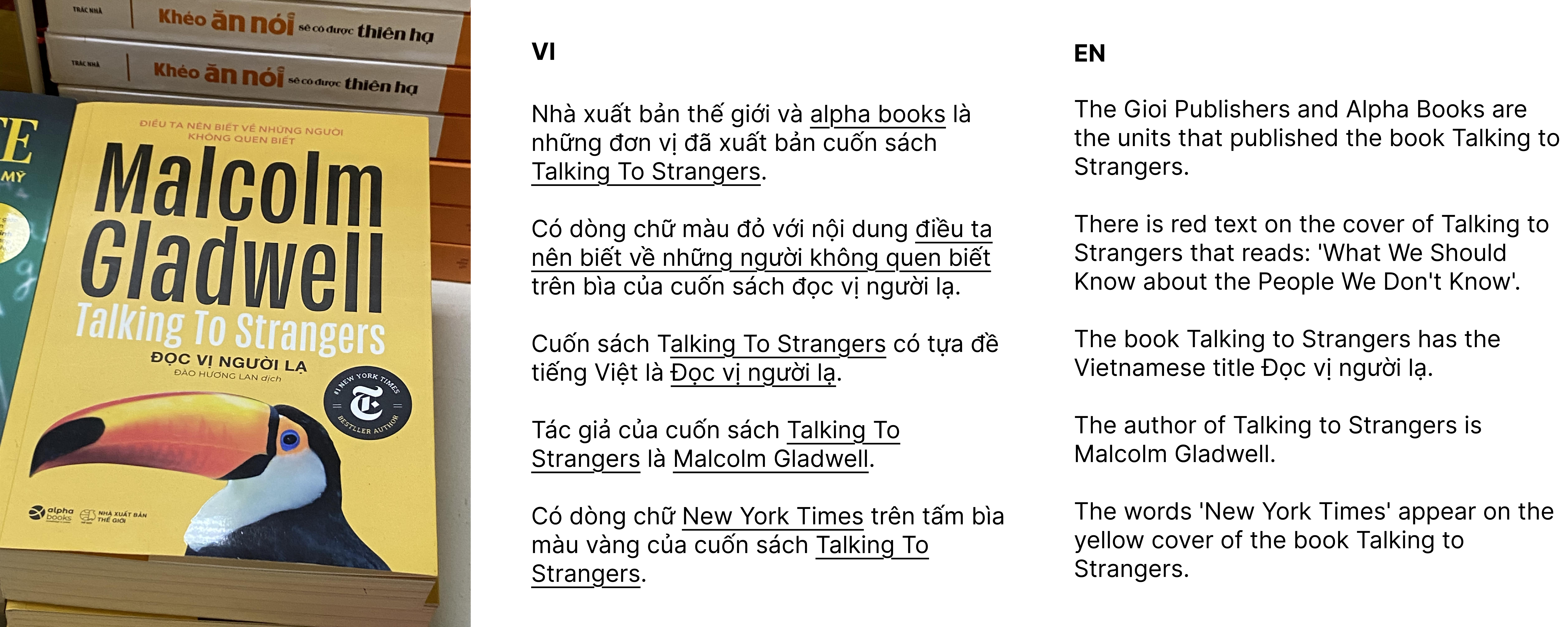} 
  \caption{
    A representative ViTextCaps sample comprising one image 
    and five human-annotated Vietnamese captions. 
    \underline{Underlined} tokens denote scene-text strings 
    copied verbatim from the image, illustrating how 
    annotators selectively incorporate visible text into 
    fluent captions.
  }
  \label{fig:sample} 
\end{figure}

Vietnamese scene-text captioning requires fusing three information
streams visual features, OCR-detected text, and linguistic
knowledge within a tonal orthography where a single diacritic
changes word meaning entirely. Existing scene-text captioning
datasets address none of these challenges, as they target
English~\cite{sidorov2020textcaps,wang2021ocr}, a non-tonal
language where diacritic errors are absent by definition.
Vietnamese scene-text VQA datasets~\cite{vitextvqa,
viocrvqa,visignvqa} address related tasks but
not captioning, which requires solving the text selection problem
without question guidance. General Vietnamese captioning datasets~\cite{lam2020uitviic}
contain no scene text. We address this gap with \textbf{ViTextCaps},
the first large-scale Vietnamese \textit{scene-text} captioning
dataset.

\begin{figure}[htbp]
  \centering
  \includegraphics[width=\textwidth]{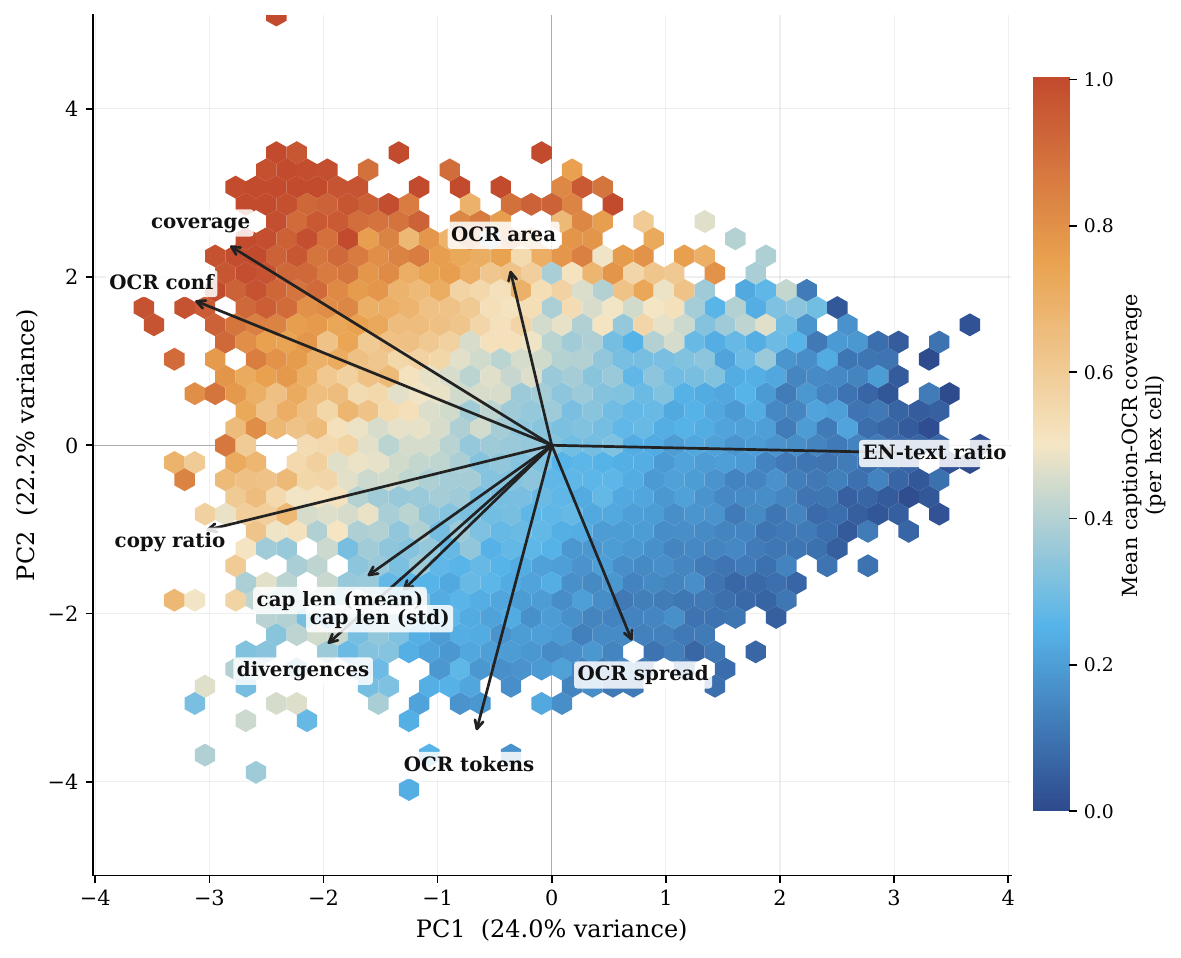}
  \caption{%
    PCA biplot of ten per-image statistics computed over all
    15{,}729 ViTextCaps images.
    Each hexagonal bin aggregates neighbouring images; colour encodes
    mean caption--OCR coverage within the bin.
    \textbf{PC1} (24.0\% variance) is dominated by \textit{EN-text
    ratio}: images shifting rightward contain predominantly
    English scene text and exhibit low annotator coverage (blue),
    whereas leftward images carry high-confidence Vietnamese OCR
    and tight coverage (orange--red).
    \textbf{PC2} (22.2\% variance) separates images by
    \textit{text density}: upward images have large OCR footprints
    and high confidence; downward images have many spatially
    dispersed tokens and longer, more varied captions
    (\textit{OCR tokens}, \textit{cap~len}).
    The two principal components together explain 46.2\% of
    dataset variance, confirming that OCR salience---not caption
    length or visual complexity---is the primary axis of
    difficulty in ViTextCaps.%
  }
  \label{fig:key_stats_pca}
\end{figure}

ViTextCaps comprises \textbf{15{,}729 images} paired with
\textbf{74{,}970 human-written Vietnamese captions}
($\sim$4.77 captions per image), sourced from diverse real-world
Vietnamese environments. Table~\ref{tab:dataset_comparison}
positions ViTextCaps among existing scene-text understanding
datasets. Figure~\ref{fig:sample} illustrates
representative examples. The construction pipeline consists of
four phases: image collection, annotation, quality control, and
validation.
Figure~\ref{fig:key_stats_pca} characterises the joint distribution
of OCR and caption statistics across all 15{,}729 images.

\subsection{Task Definition}
\label{sec:task_definition}

Formally, given an image $\mathcal{I}$ containing $N_t$
OCR-detectable text regions
$\mathcal{T} = \{(s_i, \mathbf{b}_i)\}_{i=1}^{N_t}$,
where $s_i$ is the recognized string and
$\mathbf{b}_i \in \mathbb{R}^4$ its bounding box, an annotator
produces a Vietnamese caption $Y = (y_1, \ldots, y_T)$ satisfying:
\begin{enumerate}[nosep]
    \item \textbf{(C1) Text fidelity.} Relevant scene-text tokens
    appear in $Y$ verbatim, preserving original spelling, diacritics,
    and abbreviations.
    \item \textbf{(C2) Linguistic correctness.} $Y$ is grammatically
    correct standard Vietnamese, using basic color terms and avoiding
    colloquialisms.
    \item \textbf{(C3) Descriptive completeness.} $Y$ integrates
    both visual content and textual content, embedding scene text
    naturally into the caption narrative.
\end{enumerate}
Unlike VQA, where the question specifies which text to reference,
captioning requires annotators to autonomously determine text
relevance---the text selection problem quantified in
Section~\ref{sec:text_taxonomy}.

\subsection{Data Collection}
\label{sec:data_collection}

\subsubsection{Image Collection and Filtering}

Images were collected from open web platforms (Google Images,
Facebook, Pinterest) and through manual field photography across
Vietnamese urban environments. All candidates underwent a
two-stage filtering pipeline: automated pre-screening via
SwinTextSpotter~\cite{swintextspotter} to retain only images with at
least one detectable text region, followed by manual review to
remove images with insufficient text readability (blur, occlusion),
artificially overlaid text (e.g., internet memes), or resolution
below $400 \times 400$ pixels. The final dataset contains
\textbf{15{,}729 images}.

\subsubsection{Dataset Format}
\label{sec:data_format}

Each entry in ViTextCaps links a unique image to multiple free-form
human captions. The dataset is serialized in \textbf{JSON} for
compatibility with standard vision-language frameworks:

\begin{tcolorbox}[title=Dataset Entry Format,
  colback=blue!5!white, colframe=black!75!black,
  fontupper=\small\ttfamily]
\begin{verbatim}
{
  "image_id": "<unique_identifier>",
  "file_name": "<image_filename>.jpg",
  "captions": [
    {
      "id": i,          // i = 1, ..., N; N <= 5
      "caption": "<Vietnamese free-form caption
                   integrating visible scene text>"
    }
  ]
}
\end{verbatim}
\end{tcolorbox}

Each entry contains: (1)~\texttt{image\_id}, a unique dataset
identifier; (2)~\texttt{file\_name}, the corresponding image file;
and (3)~\texttt{captions}, a list of free-form annotations each
with its own \texttt{id} and \texttt{caption} string. Multiple
captions per image capture diverse phrasings of the same visual
content, enabling robust evaluation against varied reference texts.

\subsubsection{Dataset Splits}

Splits were constructed at the \textbf{image-source level}: images
from the same physical location or storefront are assigned to the
same split, preventing visual scene overlap across train, dev, and
test. The test split is intentionally larger than the development
split (20\% vs.\ 10\%, yielding a 7:1:2 ratio) to support reliable
stratified evaluation across OCR confidence levels, text density,
and language profiles (\ref{app:stratified}), where
per-stratum sample sizes must remain sufficient for bootstrap
confidence interval estimation. Table~\ref{tab:dataset_splits}
summarizes the split statistics. Full annotation protocol and quality control procedures 
are provided in ~\ref{app:app_dataset}.

\begin{table}[htbp]
\centering
\caption{ViTextCaps dataset splits.}
\label{tab:dataset_splits}
\small
\begin{tabular}{lrrrr}
\toprule
\textbf{Split} & \textbf{Images} & \textbf{Captions}
  & \textbf{Cap/Img} & \textbf{Vocab} \\
\midrule
Train & 10{,}996 & 52{,}359 & 4.76 & 24{,}940 \\
Dev   &  1{,}641 &  7{,}883 & 4.80 &  7{,}718 \\
Test  &  3{,}092 & 14{,}728 & 4.76 & 11{,}395 \\
\midrule
Total & 15{,}729 & 74{,}970 & 4.77 & --- \\
\bottomrule
\end{tabular}
\end{table}

\subsection{Annotation Quality}
\label{sec:annotation_quality}

\subsubsection{Expert Quality Audit}
\label{sec:expert_audit}

Five research team members independent of the 30-person annotation
pool evaluated a random sample of \textbf{500 finalized
image--caption pairs} ($\sim$1\% of the corpus) against three
criteria: textual accuracy (OCR alignment), visual grounding
(absence of hallucination), and grammatical consistency.
Reviewers first assessed each pair independently, then resolved
disagreements through structured discussion to reach a consensus
judgment. At the observed pass rate of 86.8\%, a sample of 500
yields a 95\% confidence interval of $\pm$3.0\% (Wilson score
interval), providing sufficient precision to characterize
corpus-level quality.

\begin{table}[t]
\centering
\caption{Expert quality audit on 500 randomly sampled
image--caption pairs. Five research team members assessed
each pair independently before reaching consensus.
Overall Perfect requires simultaneous passage of all
three criteria.}
\label{tab:quality_audit}
\small
\begin{tabular}{lc}
\toprule
\textbf{Criterion} & \textbf{Pass Rate} \\
\midrule
Textual Accuracy (OCR Alignment)    & 91.4\% \\
Visual Grounding (No Hallucination) & 93.6\% \\
Grammatical Consistency             & 89.2\% \\
\midrule
\textbf{Overall Perfect}            & \textbf{86.8\%} \\
\bottomrule
\end{tabular}
\end{table}

Table~\ref{tab:quality_audit} shows that 86.8\% of captions
pass all three criteria simultaneously. Textual accuracy (91.4\%)
confirms reliable capture of in-the-wild scene text. Visual
grounding (93.6\%) is consistent with guideline P2
(omission over guessing). Grammatical consistency (89.2\%)
proved most demanding, reflecting the difficulty of enforcing
Vietnamese orthographic standards across a 30-person pool.

\subsubsection{Inter-Annotator Analysis}
\label{sec:iaa}

With $\sim$4.77 captions per image, ViTextCaps provides a
natural basis for inter-annotator analysis. We evaluate
agreement, diversity, and consistency via a \textbf{random
baseline}: each metric is computed under two conditions---
\textit{same-image} (captions from the same image) and
\textit{cross-image} (captions randomly paired from different
images, 1{,}000 permutations on 500 images with $\geq$4 captions).
A large same-image/cross-image ratio confirms that observed
agreement reflects genuine content overlap rather than
corpus-level vocabulary sharing.

\begin{table}[t]
\centering
\caption{Inter-annotator analysis with random baseline.
Same-image: each caption evaluated against remaining
captions for the same image. Cross-image: captions
evaluated against randomly sampled captions from different
images (1,000 permutations). Ratio = same-image /
cross-image. Dist-1 and Dist-3 computed corpus-level and
per-image. CV = coefficient of variation of caption length.
Computed on training split ($\sim$4.77 captions/image).}
\label{tab:iaa}
\small
\begin{tabular}{llccc}
\toprule
\textbf{Category} & \textbf{Metric}
  & \textbf{Same-image} & \textbf{Cross-image}
  & \textbf{Ratio} \\
\midrule
\multirow{3}{*}{Agreement}
  & BLEU-1  & 0.369 & 0.057 & $6.5\times$ \\
  & BLEU-4  & 0.189 & 0.001 & $209\times$ \\
  & ROUGE-L & 0.347 & 0.063 & $5.5\times$ \\
\midrule
\multirow{3}{*}{Diversity}
  & Dist-1 (corpus)    & \multicolumn{2}{c}{0.021} & --- \\
  & Dist-1 (per-image) & \multicolumn{2}{c}{0.490} & --- \\
  & Dist-3 (corpus)    & \multicolumn{2}{c}{0.332} & --- \\
\midrule
\multirow{2}{*}{Consistency}
  & Vocab.\ Jaccard & \multicolumn{2}{c}{0.326 ($\sigma$=0.158)} & --- \\
  & Length CV       & \multicolumn{2}{c}{0.191 ($\sigma$=0.101)} & --- \\
\midrule
Informativeness
  & Generic openings & \multicolumn{2}{c}{7.1\%} & --- \\
\bottomrule
\end{tabular}
\end{table}

Table~\ref{tab:iaa} reveals three properties. First, same-image
BLEU-4 exceeds the cross-image chance level by $\mathbf{209\times}$,
confirming that annotators consistently reference the same
scene-text content---specific $n$-grams such as shop names,
addresses, and brand names---rather than sharing only generic
Vietnamese vocabulary. Second, the moderate absolute BLEU-4
(0.189) reveals that annotators achieve this content agreement
through \textit{diverse phrasing}, expressing the same text and
objects through substantially different word sequences. This
combination---high content agreement ($209\times$ above chance)
with low surface-form overlap---rewards models that capture
semantic meaning rather than memorizing specific phrasings.
Third, only 7.1\% of captions use generic openings, confirming
that annotators engaged meaningfully with each image rather than
applying formulaic descriptions.

\section{Dataset Analysis}
\label{sec:dataset_analysis}
The following analyses characterize the linguistic and visual
properties of ViTextCaps, establishing the Vietnamese-specific
challenges that motivate the design of HSTFG and PhonoSTFG.

\subsection{Dataset Overview}
\label{sec:dataset_overview}

ViTextCaps comprises \textbf{15{,}729 images} paired with
\textbf{74{,}970 human-written captions} (average 4.77 captions
per image). The dataset is characterized by three properties
that define the Vietnamese-specific challenges developed in
Sections~\ref{sec:tone_diacritic_ambiguity}--\ref{sec:text_dependency}:
high text density, tonal linguistic complexity, and a demanding
text selection environment. Table~\ref{tab:key_stats} consolidates
the core statistics on the training split; linguistic properties
marked $^\dagger$ are derived from the analyses that follow.

\begin{table}[t]
\centering
\caption{Key dataset statistics for ViTextCaps (training set).
Linguistic properties ($^\dagger$) are consolidated from
Sections~\ref{sec:tone_diacritic_ambiguity}--\ref{sec:text_dependency}.}
\label{tab:key_stats}
\small
\begin{tabular}{lr}
\toprule
\textbf{Statistic} & \textbf{Value} \\
\midrule
\multicolumn{2}{l}{\textit{Caption Properties}} \\
\quad Mean caption length (words) & 22.1 \\
\quad Median caption length       & 19.0 \\
\quad Max caption length          & 289 \\
\quad Vocabulary size             & 24{,}940 \\
\quad Type-token ratio            & 0.022 \\
\quad Caption diversity (mean)    & 0.674 \\
\midrule
\multicolumn{2}{l}{\textit{OCR Properties}} \\
\quad Mean OCR tokens / image     & 31.8 \\
\quad Median OCR tokens / image   & 23.0 \\
\quad Mean OCR confidence         & 0.638 \\
\quad Dense images ($>$10 tokens) & 75.1\% \\
\quad Multi-block layout          & 74.3\% \\
\midrule
\multicolumn{2}{l}{\textit{Visual Properties}} \\
\quad Mean visual regions / image   & 31.1 \\
\quad Median visual regions / image & 29.0 \\
\midrule
\multicolumn{2}{l}{\textit{Linguistic Properties}$^\dagger$} \\
\quad Diacritic collision rate (caption vocab) & 52.8\% \\
\quad Caption--OCR diacritic divergence        & 20.6\% \\
\quad Images with bilingual text               & 42.9\% \\
\quad Copy ratio (caption tokens from OCR)     & 28.9\% \\
\quad Text coverage rate (mean)                & 35.9\% \\
\bottomrule
\end{tabular}
\end{table}

\begin{figure}[htbp]
  \centering
  \includegraphics[width=\linewidth]{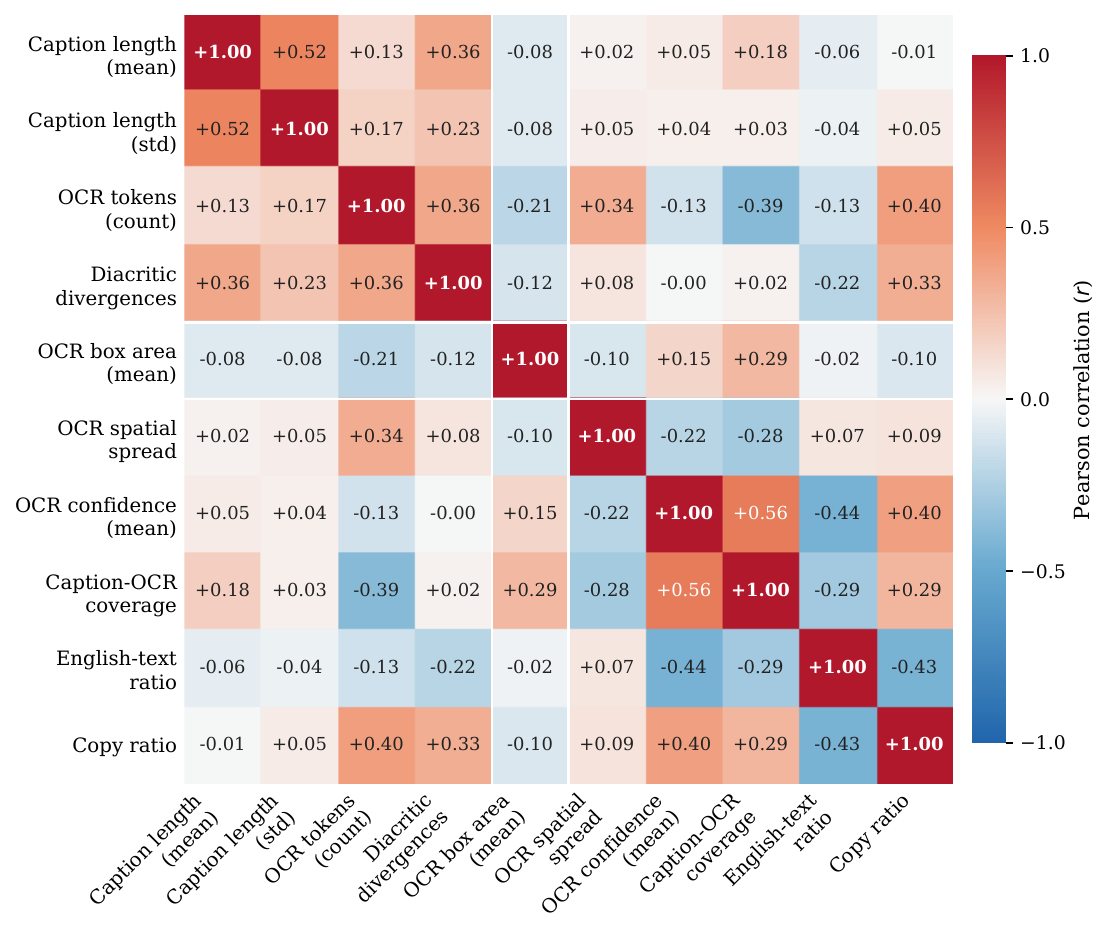}
  \caption{%
    Hierarchical-cluster-ordered Pearson correlation heatmap of
    10 per-image statistics (training split). Clusters reveal three
    groups: OCR output properties (density, spread, area),
    caption--OCR interaction (copy ratio, coverage, divergences),
    and quality indicators (confidence, caption length).
    Diacritic divergences correlate strongly with OCR density
    ($r = 0.62$) but weakly with confidence ($r = -0.21$),
    confirming that exposure to ambiguous tokens---not low
    detection quality---drives most diacritic errors.%
  }
  \label{fig:key_stats_corr}
\end{figure}

Figure~\ref{fig:key_stats_corr} complements the PCA view with
a cluster-ordered Pearson correlation matrix of the same 10
per-image statistics, making the inter-variable structure
directly legible without requiring dimension reduction.

\subsection{Tone and Diacritic Ambiguity}
\label{sec:tone_diacritic_ambiguity}

Vietnamese is a tonal language with six lexical tones, each realized
orthographically through diacritical marks placed above or below vowel
characters. A single base syllable such as \textit{ma} can correspond
to up to six semantically unrelated words: \textit{ma} (ghost),
\textit{mà} (but), \textit{má} (mother), \textit{mả} (grave),
\textit{mã} (code), and \textit{mạ} (rice seedling). This property
is unique among Latin-script languages and introduces a class of
ambiguity that does not exist in English scene-text understanding:
a single diacritical error---often caused by the misrecognition of
a sub-pixel mark---can change the meaning of a word entirely.

We quantify this vulnerability through three complementary analyses:
\textit{vocabulary-level collision rate}, \textit{caption--OCR
diacritic divergence}, and a study of the \textit{most dangerous
collision groups} in ViTextCaps.

\subsubsection{Diacritic Collision Rate}
\label{sec:collision_rate}

We define a \textit{diacritic collision group}
$\mathcal{G}(b) = \{w \in \mathcal{V}_{\mathrm{VN}} :
\mathrm{base}(w) = b\}$ as the set of distinct Vietnamese words
sharing the same base form $b$, obtained by Unicode NFD
decomposition followed by removal of combining diacritical
marks (U+0300--U+036F). The \textit{collision rate} of a
vocabulary $\mathcal{V}$ is the fraction of Vietnamese words
belonging to at least one collision group:
\begin{equation}
\mathrm{CR}(\mathcal{V}) =
\frac{|\{w \in \mathcal{V}_{\mathrm{VN}} :
  |\mathcal{G}(\mathrm{base}(w))| \geq 2\}|}
  {|\mathcal{V}_{\mathrm{VN}}|}
\end{equation}
where $\mathcal{V}_{\mathrm{VN}} \subset \mathcal{V}$ retains
only Vietnamese words, excluding English brand names, numeric
codes, and URLs, to isolate the effect of Vietnamese diacritics
from non-Vietnamese noise.

\begin{figure}[htbp]
\centering
\includegraphics[width=\textwidth]{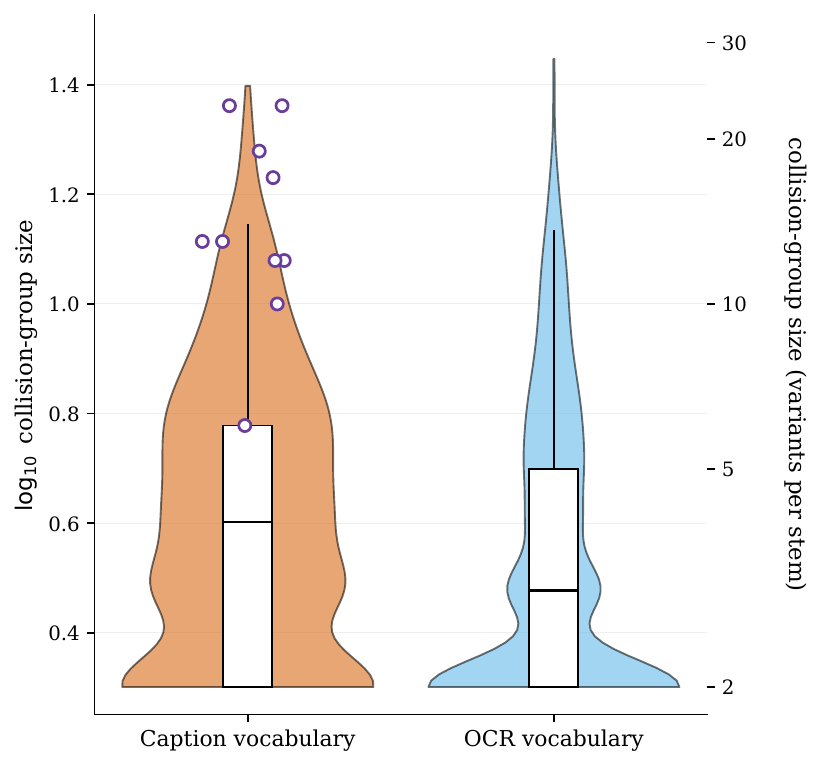}
\caption{Collision-group size distribution in the Vietnamese caption
vocabulary ($n{=}912$ groups) and the OCR vocabulary
($n{=}2{,}855$). Group size is the number of diacritic variants
that share one base form. Violins show the full distribution on a
$\log_{10}$ scale (raw counts on the right axis); inner
box-and-whisker marks the interquartile range and median. Purple
rings overlay the ten highest-\textit{danger-score} stems in the
caption vocabulary (group size $\times$ total frequency): nine of
ten fall beyond the 75th percentile of the caption distribution.
The caption vocabulary carries a heavier upper tail than the OCR
vocabulary, quantifying the ``ambiguity amplification'' between
input and output vocabularies.}
\label{fig:motivation}
\end{figure}

Table~\ref{tab:collision_stats} and Figure~\ref{fig:motivation}
report the collision statistics for both caption and OCR
vocabularies on the training split.

\begin{table}[ht]
\centering
\caption{Diacritic collision statistics for Vietnamese-only words
in caption and OCR vocabularies. \textit{Collision rate} measures
the proportion of unique words that share a base form with at
least one other word when diacritics are removed.}
\label{tab:collision_stats}
\begin{tabular}{lcc}
\toprule
\textbf{Metric} & \textbf{Caption} & \textbf{OCR} \\
\midrule
Total Vietnamese words  & 8{,}565  & 31{,}195 \\
Unique base forms       & 4{,}956  & 22{,}160 \\
Collision groups        & 912      & 2{,}855  \\
Words in collision      & 4{,}521  & 11{,}890 \\
\textbf{Collision rate} & \textbf{52.8\%} & \textbf{38.1\%} \\
\bottomrule
\end{tabular}
\end{table}

Two observations emerge. First, the caption vocabulary exhibits a
collision rate of 52.8\%---meaning that more than half of the
Vietnamese words a captioning model must generate have at least
one diacritic-ambiguous counterpart. This poses a fundamental
challenge for generation: even with perfect visual understanding,
selecting the correct diacritical variant from a collision group
requires robust linguistic reasoning.

Second, the OCR vocabulary shows a lower but still substantial rate
of 38.1\%. The gap between caption and OCR collision rates reflects
the compositional difference between the two vocabularies: annotators
write fluent Vietnamese using common, high-frequency words that
overlap heavily in their base forms, whereas OCR output includes a
long tail of rare tokens---proper names, partial words, and
mixed-script fragments---that are less likely to participate in
collisions. Critically, this means that the \textit{target output
space} is more ambiguous than the input space, making Vietnamese
scene-text captioning a problem where the generation difficulty
exceeds the recognition difficulty.

\subsubsection{Dangerous Collision Groups}
\label{sec:collision_groups}

Not all collision groups carry equal risk. Table~\ref{tab:top_collisions}
lists the most dangerous groups, ranked by a \textit{danger score}
defined as $|\mathcal{G}| \times \sum_{w \in \mathcal{G}} f(w)$,
where $|\mathcal{G}|$ is the group size and $f(w)$ is the frequency
of word $w$ in the caption corpus.

\begin{table}[ht]
\centering
\caption{Top diacritic collision groups in the ViTextCaps caption
vocabulary, ranked by danger score (group size $\times$ total
frequency). Each row shows a base form and the distinct Vietnamese
words that collapse to it when diacritics are stripped.}
\label{tab:top_collisions}
\begin{tabular}{clcc}
\toprule
\textbf{Base Form} & \textbf{Collision Words (examples)}
  & \textbf{$|\mathcal{G}|$} & \textbf{$\sum f$} \\
\midrule
\texttt{co}
  & có, cô, cổ, cơ, cờ, cỏ, cố, \ldots   & 19 & 37{,}601 \\
\texttt{mau}
  & màu, mẫu, mau, máu, mầu, mẩu, \ldots  & 12 & 36{,}311 \\
\texttt{dong}
  & dòng, đồng, đông, động, đóng, \ldots   & 17 & 25{,}076 \\
\texttt{chu}
  & chữ, chủ, chú, chứ, chư, \ldots        & 13 & 28{,}193 \\
\texttt{do}
  & đỏ, đồ, đó, do, đô, độ, đổ, \ldots    & 23 & 11{,}531 \\
\texttt{ban}
  & bán, bản, bàn, bạn, ban, bắn, \ldots   & 13 &  9{,}147 \\
\texttt{dung}
  & đứng, dung, dựng, dụng, dùng, \ldots   & 20 &  5{,}859 \\
\bottomrule
\end{tabular}
\end{table}

Several of these groups are directly relevant to scene-text
understanding. The base form \texttt{ban} collapses four
semantically distinct high-frequency words: \textit{bán}
(to sell, $f$=3{,}969), \textit{bản} (copy/version, $f$=3{,}279),
\textit{bàn} (table, $f$=884), and \textit{bạn} (friend,
$f$=568)---all of which appear frequently on Vietnamese shop signs
and are crucial for describing commercial scenes. Similarly,
\texttt{dong} conflates \textit{dòng} (line/row, $f$=20{,}715)
with \textit{đồng} (currency unit, $f$=2{,}227) and \textit{đông}
(east, $f$=1{,}011), where confusing the currency with a direction
fundamentally changes the semantics of a caption.

The largest group, \texttt{do}, contains 23 distinct words. While
not all of these are high-frequency, the sheer size of the group
means that any base-form match during generation presents the model
with 23 possible surface realizations, each carrying a different
meaning.

\subsubsection{Caption--OCR Diacritic Divergence}
\label{sec:diacritic_divergence}

The collision analysis above characterizes vulnerability at the
vocabulary level. To measure how often diacritic ambiguity
manifests \textit{in practice}, we perform an instance-level
analysis. For each image, we match caption tokens to OCR tokens
by their stripped base forms. When a match is found, we check
whether the diacritics agree (exact match) or differ (divergence).
This comparison does not require ground-truth text annotations:
the caption reflects human reading of the original image, and the
OCR output reflects machine reading of the same image. Any
divergence in diacritics between the two indicates that at least
one of them has resolved the diacritic ambiguity differently.

Table~\ref{tab:diacritic_divergence} reports the divergence rates
stratified by OCR confidence.

\begin{table}[ht]
\centering
\caption{Caption--OCR diacritic divergence stratified by OCR
confidence. The strong inverse relationship between confidence
and divergence rate---2.8$\times$ higher at low confidence than
at high---motivates the confidence gate in HSTFG
(Section~\ref{sec:hstfg_graph}), which down-weights attention
toward detections where diacritic errors are most prevalent.}
\label{tab:diacritic_divergence}
\begin{tabular}{lccc}
\toprule
\textbf{OCR Confidence} & \textbf{Matches}
  & \textbf{Divergences} & \textbf{Rate} \\
\midrule
Low ($< 0.5$)         &  52{,}686 & 21{,}471 & 40.8\% \\
Medium ($0.5$--$0.8$) & 108{,}281 & 27{,}760 & 25.6\% \\
High ($\geq 0.8$)     & 273{,}977 & 40{,}499 & 14.8\% \\
\midrule
\textbf{Overall}
  & \textbf{434{,}944} & \textbf{89{,}730} & \textbf{20.6\%} \\
\bottomrule
\end{tabular}
\end{table}

One in five caption--OCR token pairs that share the same base form
disagree on diacritics overall. The divergence rate exhibits a
strong inverse correlation with OCR confidence: at low confidence
($< 0.5$), the rate reaches 40.8\%, nearly three times the rate at
high confidence ($\geq 0.8$, 14.8\%). This gradient confirms that
diacritic ambiguity is not merely a theoretical concern but a
pervasive empirical phenomenon that worsens precisely when
recognition is uncertain.

Concrete examples illustrate the practical impact:
\begin{itemize}[nosep]
    \item \textit{hóa} (chemistry) $\leftrightarrow$ \textit{hoa}
    (flower): OCR drops the acute accent at confidence 0.31,
    changing the semantic category entirely.
    \item \textit{nguyên} (original) $\leftrightarrow$
    \textit{nguyễn} (a surname): OCR substitutes a tilde for a
    circumflex-acute at confidence 0.81, converting a common
    adjective into Vietnam's most frequent family name.
    \item \textit{quan} (official/relation) $\leftrightarrow$
    \textit{quán} (shop) $\leftrightarrow$ \textit{quản}
    (to manage): a single base form matched to three different
    OCR readings of the same image, each with confidence above
    0.76.
\end{itemize}

These findings have three direct consequences for model design.
Standard metrics such as BLEU and CIDEr treat \textit{bán}
(to sell) and \textit{bàn} (table) as a routine one-token
mismatch---the same penalty applied to substituting \textit{large}
for \textit{big} in English---underestimating the semantic impact
of diacritic errors; this motivates the Text Hallucination Rate
(THR) metric proposed in Section~\ref{sec:main_results}.
With 52.8\% of the caption vocabulary at risk of diacritic
collision, a model that treats Vietnamese characters as opaque
symbols will frequently select incorrect collision variants,
motivating the phonological attention bias in PhonoSTFG
(Section~\ref{sec:phonological_bias}) which explicitly encodes
tone and diacritic features to guide text fusion.
Finally, even at high OCR confidence ($\geq 0.8$), 14.8\% of
matched tokens still exhibit diacritic divergence---the ambiguity
is rooted in Vietnamese orthography itself, not in any particular
recognizer's limitations, positioning ViTextCaps as a benchmark
that will remain challenging regardless of future OCR improvements.

\subsection{Text Usage and Selection}
\label{sec:text_dependency}

The ViTextCaps annotation guideline requires every caption to reference
text visible in the image while allowing paraphrasing for natural
expression. This design choice means that the relevant question is
not \textit{whether} captions depend on scene text---they all do by
construction---but rather \textit{how} annotators select and
incorporate text when faced with text-rich images averaging 31 OCR
tokens per image.

\subsubsection{Text Usage Taxonomy}
\label{sec:text_taxonomy}

We classify each caption into one of four categories based on how its
tokens relate to the image's OCR output. For each OCR token, we check
for exact match, base-form match (diacritics stripped), or substring
containment against caption tokens. The caption is then classified
by the proportion and type of matches:

\begin{itemize}[nosep]
    \item \textbf{Verbatim heavy} ($>$50\% of matched OCR tokens are
    exact copies): The caption quotes scene text directly.
    \item \textbf{Paraphrase heavy} ($>$30\% coverage, majority are
    approximate matches): The caption references text but rephrases it.
    \item \textbf{Partial reference} (15--30\% coverage): The caption
    mentions a small subset of the available text.
    \item \textbf{Contextual inference} ($<$15\% coverage): The caption
    uses scene text to infer context rather than quoting it.
\end{itemize}

\begin{figure}[htbp]
\centering
\includegraphics[width=\linewidth]{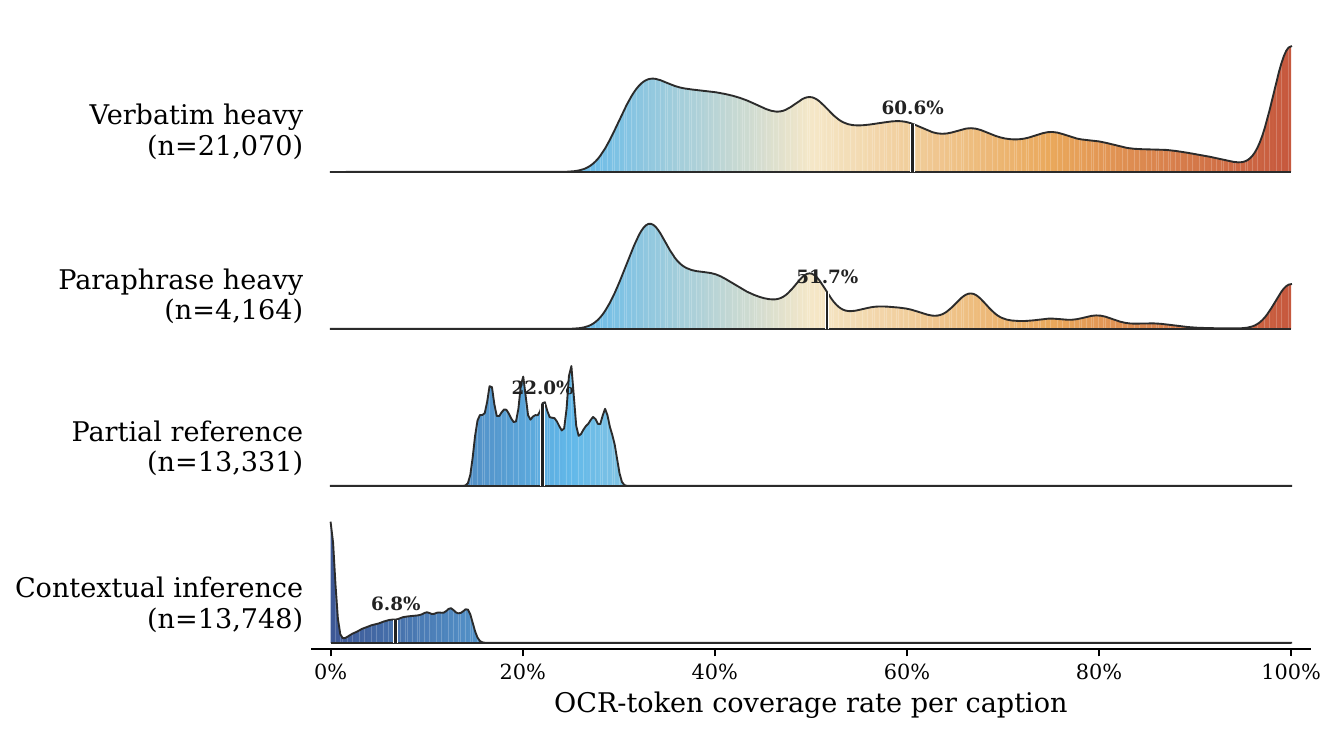}
\caption{Per-caption OCR-token coverage rate, stratified by text
usage type (KDE ridges over $n{=}52{,}313$ training captions,
ordered bottom-to-top by mean coverage). Ridge amplitude scales
with $\log_{10}$ of the per-category count, so prevalence reads
from vertical height; gradient fill encodes coverage on the x-axis
(cool $\to$ warm). Vertical ticks mark the per-category mean.
Beyond the aggregates in
Table~\ref{tab:text_usage_taxonomy}, the figure surfaces a
\textbf{bimodal} paraphrase-heavy distribution (peaks near 0.2 and
1.0) that the single-mean summary does not capture, and shows a
contextual-inference distribution almost disjoint from the other
three categories---a regime a pure pointer-copy mechanism cannot
reach.}
\label{fig:text_usage}
\end{figure}

Table~\ref{tab:text_usage_taxonomy} reports the distribution across
all four categories on the training set;
Figure~\ref{fig:text_usage} shows the corresponding per-caption
coverage distributions on raw data.

\begin{table}[ht]
\centering
\caption{Text usage taxonomy of ViTextCaps captions. Despite the
annotation requirement to reference OCR text, usage patterns vary
substantially. \textit{Avg.\ coverage} is the mean fraction of
valid OCR tokens mentioned in captions of each type.}
\label{tab:text_usage_taxonomy}
\begin{tabular}{lrrr}
\toprule
\textbf{Usage Type} & \textbf{Captions} & \textbf{\%}
  & \textbf{Avg.\ Cov.} \\
\midrule
Verbatim heavy       & 21{,}070 & 40.2\% & 60.6\% \\
Partial reference    & 13{,}331 & 25.5\% & 22.0\% \\
Contextual inference & 13{,}748 & 26.3\% &  6.8\% \\
Paraphrase heavy     &  4{,}164 &  8.0\% & 51.7\% \\
\midrule
\textbf{Overall} & \textbf{52{,}328} & \textbf{100\%}
  & \textbf{35.9\%} \\
\bottomrule
\end{tabular}
\end{table}

While 40.2\% of captions directly quote OCR tokens (verbatim heavy),
over half (51.8\%) reference text only partially or inferentially.
Notably, 26.3\% fall into the \textit{contextual inference} category,
where the average coverage is only 6.8\%---annotators use the text to
understand the scene (e.g., reading ``nhà thuốc'' [pharmacy] to write
``cửa hàng bán thuốc'' [a shop selling medicine]) rather than quoting
it. This distribution demonstrates that the \textit{text selection
problem} is not an edge case but a central characteristic of the
dataset: annotators must actively decide which text to mention, how
to rephrase it, and what to infer from it.

\subsubsection{Text Coverage Rate}
\label{sec:coverage_rate}

We define \textit{text coverage rate} as the fraction of valid OCR
tokens (excluding stopwords and very short tokens) that are referenced
in a given caption. Table~\ref{tab:coverage_stats} reports the
distribution across the training set.

\begin{table}[ht]
\centering
\caption{Distribution of text coverage rate across training
captions. 43.3\% of captions reference fewer than 25\% of
available OCR tokens, confirming pervasive selective reporting.}
\label{tab:coverage_stats}
\begin{tabular}{lrr}
\toprule
\textbf{Coverage Bucket} & \textbf{Captions} & \textbf{\%} \\
\midrule
Very low ($<$10\%)       &  8{,}845 & 16.9\% \\
Low (10--25\%)           & 13{,}797 & 26.4\% \\
Medium (25--50\%)        & 15{,}009 & 28.7\% \\
High (50--75\%)          &  7{,}786 & 14.9\% \\
Very high ($\geq$75\%)   &  6{,}876 & 13.1\% \\
\bottomrule
\end{tabular}
\end{table}

The mean coverage is \textbf{35.9\%} (median 28.6\%), meaning captions
reference roughly one-third of available scene text. 43.3\% of captions
reference fewer than 25\% of OCR tokens, while only 13.1\% achieve
coverage above 75\%, confirming that selective text reporting is the
norm rather than the exception.

Coverage correlates negatively with text density ($r = -0.352$): as
OCR token count increases, annotators reference a progressively smaller
fraction. This is the quantitative signature of the text selection
problem---in text-dense images, the model must learn not only to read
text but to judge which text is worth mentioning.

\subsubsection{What Gets Selected?}
\label{sec:selection_factors}

\begin{figure}[htbp]
\centering
\includegraphics[width=\linewidth]{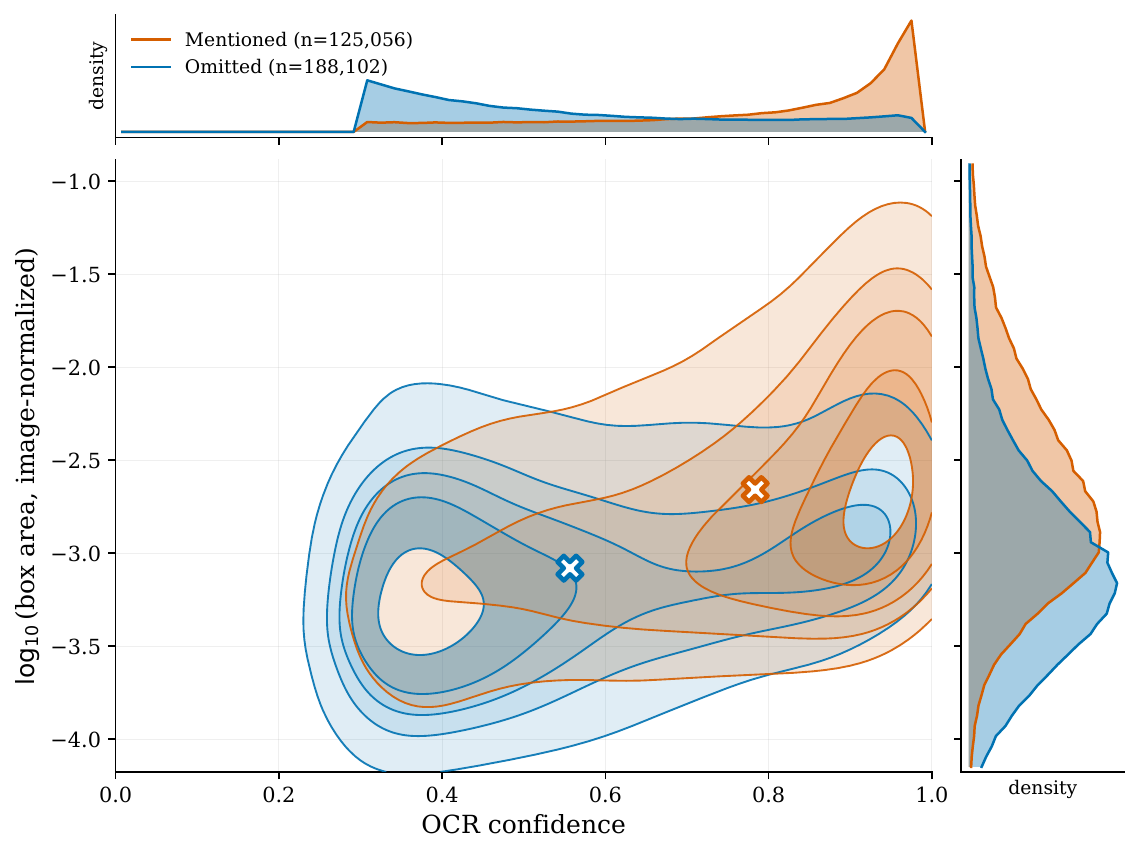}
\caption{Joint distribution of OCR tokens in salience space
(confidence $\times$ $\log_{10}$ box area) for the train split
($n{=}313{,}158$). Two overlaid KDE contour sets at 10/30/50/70/90\%
cumulative density: mentioned tokens (vermilion, $n{=}125{,}056$)
versus omitted tokens (sky blue, $n{=}188{,}102$).
Centroid ``$\times$'' markers show per-group means. Top and right
margins give the 1D densities along each axis. The two clouds are
systematically offset in the same direction on both axes, so while
neither dimension alone separates them, the joint structure does
---motivating architectures that integrate spatial reasoning with
linguistic features, as in PhonoSTFG.}
\label{fig:text_selection}
\end{figure}

Table~\ref{tab:text_selection} and Figure~\ref{fig:text_selection}
compare the properties of OCR tokens that annotators choose to
mention against those they omit.

\begin{table}[ht]
\centering
\caption{Properties of OCR tokens mentioned in captions vs.\ those
omitted. Mentioned tokens have substantially higher OCR confidence
and larger bounding boxes, identifying visual salience as the
primary selection criterion.}
\label{tab:text_selection}
\begin{tabular}{lcc}
\toprule
\textbf{Property} & \textbf{Mentioned} & \textbf{Not Mentioned} \\
\midrule
Count            & 125{,}056 & 188{,}102 \\
Mean confidence  & \textbf{0.784}  & 0.557  \\
Mean box area    & \textbf{0.0064} & 0.0023 \\
Median box area  & \textbf{0.0024} & 0.0008 \\
Mean Y position  & 0.464           & 0.483  \\
\bottomrule
\end{tabular}
\end{table}

Mentioned tokens have mean OCR confidence 0.784 versus 0.557 for
omitted tokens (+0.227), indicating that annotators reference text
they can read clearly. More strikingly, mentioned tokens occupy
bounding boxes \textbf{2.8$\times$ larger} in area (0.0064
vs.\ 0.0023), confirming that physical prominence is a primary
selection criterion. Vertical position shows a weaker effect:
mentioned tokens are marginally higher (mean $y = 0.464$
vs.\ $0.483$), suggesting a slight top-to-bottom reading preference.

These findings establish three properties of the text selection
problem. Unlike VQA, where the question specifies which text to
focus on, scene-text captioning requires the model to autonomously
identify relevant text from a cluttered scene---with mean coverage
of 35.9\%, models must generate informative captions while ignoring
approximately two-thirds of available text. The 2.8$\times$ box area
gap and +0.23 confidence gap confirm that selection is driven by
visual salience rather than random sampling, motivating architectures
that integrate spatial reasoning with linguistic features through
graph-based fusion---precisely the design of PhonoSTFG
(Section~\ref{sec:phonostfg}). Finally, the negative
coverage--density correlation ($r = -0.35$) defines a natural
difficulty gradient: text-dense images, which constitute 75\% of
ViTextCaps, present the hardest selection challenge, enabling
fine-grained model evaluation through density-stratified
benchmarking (\ref{sec:text_density_1}).

Further dataset analyses—including domain distribution, 
vocabulary statistics, OCR error taxonomy, code-mixing 
patterns, and linguistic structure—are provided in 
\ref{app:further_analysis}.

\section{Our Proposed Method}
\label{sec:proposed_method}

We propose a two-stage approach to scene-text fusion for
Vietnamese image captioning. First, we introduce \textbf{HSTFG}
(Section~\ref{sec:hstfg}), a heterogeneous graph fusion framework
that integrates visual regions and OCR tokens through configurable
edge types with learned spatial attention bias. Second, we propose
\textbf{PhonoSTFG} (Section~\ref{sec:phonostfg}), which
specializes graph-level fusion for Vietnamese linguistic reasoning
with four components: (1)~PhoBERT Dual-Stream OCR Embedding with
Gated Fusion, (2)~T$\to$T-only graph topology,
(3)~Linguistic Residual Preservation, and (4)~Vietnamese
Phonological Attention Bias---to our knowledge, the first work
to integrate explicit Vietnamese phonological knowledge into the
attention mechanism for scene-text image captioning
(Section~\ref{sec:phonological_bias}).

\subsection{Problem Formulation}
\label{sec:problem_formulation}

Given an input image $I$, a visual encoder extracts $N_v$ region
features $\mathcal{V} = \{\mathbf{x}_1^v, \ldots,
\mathbf{x}_{N_v}^v\}$ with bounding boxes
$\mathcal{B}^v = \{\mathbf{b}_1^v, \ldots, \mathbf{b}_{N_v}^v\}$,
and an OCR detector extracts $N_t$ text tokens
$\mathcal{T} = \{(s_i,\, \mathbf{x}_i^t,\, \mathbf{b}_i^t,\,
c_i)\}_{i=1}^{N_t}$, where $s_i$ is the recognized string,
$\mathbf{x}_i^t$ the feature vector,
$\mathbf{b}_i^t \in \mathbb{R}^4$ the bounding box, and
$c_i \in [0,1]$ the detection confidence.

The task is to generate a Vietnamese caption
$Y = (y_1, \ldots, y_T)$ that faithfully integrates relevant
scene text. At each step $t$, the model predicts:
\begin{equation}
p(y_t \mid y_{<t}, \mathcal{V}, \mathcal{T}) =
(1 - p_t^{\mathrm{copy}}) \cdot p_t^{\mathrm{vocab}}(y_t)
+ p_t^{\mathrm{copy}} \cdot p_t^{\mathrm{ocr}}(y_t)
\end{equation}
where $p_t^{\mathrm{vocab}}$ is a distribution over a fixed
vocabulary $\mathcal{W}$, $p_t^{\mathrm{ocr}}$ is a pointer
distribution over OCR tokens via
OcrPtrNet~\cite{hu2020iterative}, and $p_t^{\mathrm{copy}}$
is a learned mixing weight. Training minimises:
\begin{equation}
\mathcal{L} = -\sum_{t=1}^{T}
\log\, p(y_t^* \mid y_{<t}^*,\, \mathcal{V},\, \mathcal{T})
\end{equation}
where $Y^*$ is the ground-truth caption. All notation introduced
here is used consistently throughout both proposed models.

\subsection{HSTFG: Heterogeneous Scene-Text Fusion Graph}
\label{sec:hstfg}

\begin{figure*}[htbp]
  \centering
  \includegraphics[width=\textwidth]{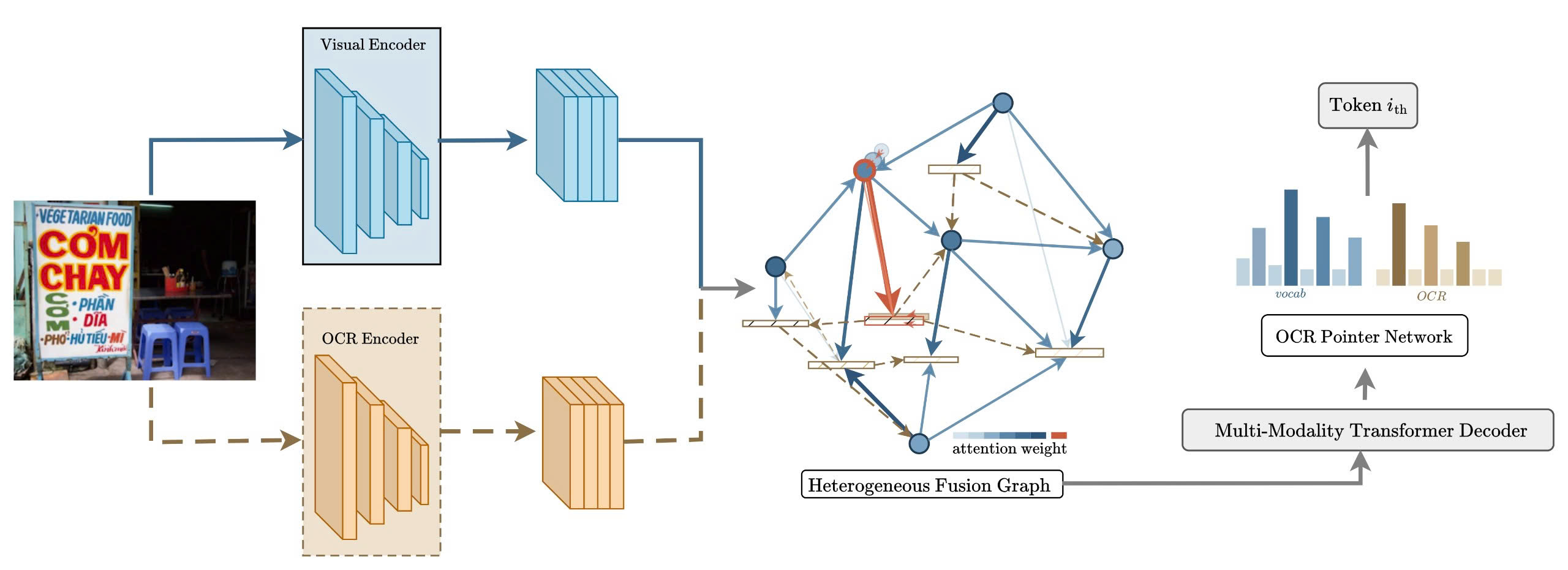}
  \caption{%
    \textbf{HSTFG: Heterogeneous Scene-Text Fusion Graph.}
    \textbf{Full pipeline:} a Faster~R-CNN/VinVL visual
    encoder and a SwinTextSpotter OCR detector produce visual
    region embeddings (V~nodes) and OCR text token embeddings
    (T~nodes), both projected to $d{=}768$. Three configurable
    edge types (V$\to$T, T$\to$V, T$\to$T) with independent
    $W_Q,W_K,W_V$ projections fuse all modalities across
    $L{=}3$ spatial graph attention layers, after which an MMT
    decoder generates captions via a mixture distribution over
    the fixed vocabulary and the OCR token pool.
  }
  \label{fig:hstfg}
\end{figure*}

HSTFG is a general-purpose graph fusion framework for scene-text
image captioning (Figure~\ref{fig:hstfg}). An input image
is processed in parallel by two feature extractors: a visual
encoder (Faster R-CNN with VinVL backbone~\cite{zhang2021vinvl})
produces $N_v$ visual regions (V~nodes), and an OCR detector
(SwinTextSpotter~\cite{swintextspotter}) produces $N_t$ text tokens
(T~nodes). These heterogeneous nodes are processed through
$L$~spatial graph attention layers, then fed into a multimodal
decoder.

\subsubsection{Node Embeddings}
\label{sec:node_embed}

Visual regions and OCR tokens reside in different feature spaces
(2048-d and 812-d respectively). We project both to a shared
dimension $d = 768$, matching the hidden size of the downstream
MMT decoder, so that graph attention operates over a unified
representation space.

\paragraph{V~nodes.}
Each visual region is projected to $d$ with additive position
encoding:
\begin{equation}
\mathbf{v}_i = \mathrm{LN}(W_v\, \mathbf{x}_i^v)
+ \mathrm{LN}(W_b\, \mathbf{b}_i^v)
\end{equation}
where $W_v \in \mathbb{R}^{d \times 2048}$,
$W_b \in \mathbb{R}^{d \times 4}$, and $\mathrm{LN}$ denotes
Layer Normalization.

\paragraph{T~nodes.}
Each OCR token is represented by concatenating a FastText
embedding ($\mathbf{e}_i^{\mathrm{ft}} \in \mathbb{R}^{300}$),
recognition features ($\mathbf{r}_i \in \mathbb{R}^{256}$), and
detection features ($\mathbf{d}_i \in \mathbb{R}^{256}$),
totaling 812 dimensions:
\begin{equation}
\mathbf{t}_i = \mathrm{LN}\!\left(W_t\,
[\mathbf{e}_i^{\mathrm{ft}};\,\mathbf{r}_i;\,\mathbf{d}_i]
\right) + \mathrm{LN}(W_b\, \mathbf{b}_i^t)
\end{equation}
where $W_t \in \mathbb{R}^{d \times 812}$.

\subsubsection{Heterogeneous Graph with Spatial Attention Bias}
\label{sec:hstfg_graph}

HSTFG constructs a heterogeneous graph over V and T~nodes with
three configurable edge types, each with its own learned
$W_Q, W_K, W_V$ projections:
\begin{itemize}[nosep]
    \item \textbf{V$\to$T}: T~queries attend to V~keys
    \item \textbf{T$\to$V}: V~queries attend to T~keys
    \item \textbf{T$\to$T}: T self-attention
\end{itemize}
We omit V$\to$V edges: visual regions are already contextualized
by Faster R-CNN's multi-layer feature pyramid and region proposal
network, making additional V$\to$V graph attention redundant.

\paragraph{Spatial attention bias.}
Standard dot-product attention is agnostic to the spatial
arrangement of nodes in the image. To encode geometric
relationships, we add a learnable spatial bias that influences
attention independently of content similarity.

For each node pair $(i, j)$, we compute scale-invariant spatial
features following Hu~et~al.~\cite{hu2018relation}:
\begin{equation}
\label{eq:spatial_feat}
\mathbf{s}_{ij} = \left[\frac{c_x^j - c_x^i}{w^i},\;
\frac{c_y^j - c_y^i}{h^i},\;
\log\frac{w^j}{w^i},\;
\log\frac{h^j}{h^i}\right] \in \mathbb{R}^4
\end{equation}
where $(c_x, c_y, w, h)$ are the center coordinates and
dimensions of each bounding box. The log-ratio terms ensure scale
invariance. A two-layer MLP with ReLU non-linearity maps these
features to $H$ per-head attention biases:
\begin{equation}
\label{eq:spatial_attn}
A_{ij}^{(h)} =
\frac{{\mathbf{q}_i^{(h)}}^\top \mathbf{k}_j^{(h)}}{\sqrt{d_h}}
+ \underbrace{{w_{\mathrm{sp}}^{(h)}}^\top
  \phi(\mathbf{s}_{ij})}_{\text{spatial bias}}
\end{equation}
where $\phi: \mathbb{R}^4 \xrightarrow{W_1^{\mathrm{sp}}}
\mathbb{R}^{32} \xrightarrow{\mathrm{ReLU}} \mathbb{R}^{32}$
is the shared MLP hidden layer,
$w_{\mathrm{sp}}^{(h)} \in \mathbb{R}^{32}$ are per-head output
weights, and $d_h = d/H$.

\paragraph{Confidence gate.}
OCR detectors assign each detection a confidence score
$c_j \in [0,1]$. We incorporate this as a learnable post-softmax
gate:
\begin{equation}
\label{eq:confidence_gate}
\tilde{\alpha}_{ij}^{(h)} =
\mathrm{softmax}\!\left(\mathbf{A}^{(h)}\right)_{ij}
\cdot \sigma\!\left(W_c^{(h)}\, c_j + b_c^{(h)}\right)
\end{equation}
where $W_c^{(h)} \in \mathbb{R}$ and $b_c^{(h)} \in \mathbb{R}$
produce a per-head gate value, and $\sigma$ is the sigmoid
function. The score $c_j$ is fixed input from the OCR detector,
not updated during training.

\paragraph{Graph layer update.}
At each layer $\ell \in \{1, \ldots, L\}$, each node aggregates
information from its neighbours using the gated attention weights.
For T~nodes under T$\to$T attention:
\begin{equation}
\label{eq:graph_update}
\mathbf{t}_i^{(\ell)} = \mathrm{LN}\!\left(
  \mathbf{t}_i^{(\ell-1)} +
  W_O^{(\ell)}\,\mathrm{Concat}_{h=1}^{H}\!\left(
    \sum_{j=1}^{N_t} \tilde{\alpha}_{ij}^{(h,\ell)}\,
    W_V^{(h,\ell)}\, \mathbf{t}_j^{(\ell-1)}
  \right)
\right)
\end{equation}
where $W_V^{(h,\ell)} \in \mathbb{R}^{d_h \times d}$ is the
value projection for head $h$ at layer $\ell$, and
$W_O^{(\ell)} \in \mathbb{R}^{d \times d}$ projects the
concatenated multi-head output. V-T and T-V edge types follow
the same update structure with their respective node sets and
projections. The final output of each layer also passes through a
position-wise feed-forward network with residual connection,
following the standard transformer block~\cite{vaswani2017attention}.

\paragraph{Graph depth.}
We use $L = 3$ spatial heterogeneous attention layers, balancing
two opposing forces inherent to GNNs: greater depth enables
multi-hop reasoning---after $L$ layers, each T~node has integrated
information from all T~nodes within $L$ hops via
Eq.~\eqref{eq:graph_update}---while excessive depth causes
over-smoothing, where repeated neighbourhood aggregation drives
all embeddings toward a uniform
representation~\cite{li2018deeper}. The 2--3 layer regime is
consistent with the broader heterogeneous GNN
literature~\cite{yun2019graph}. After $L$ layers, the updated
V and T~embeddings are concatenated and fed into the decoder.

\subsubsection{Decoder}
\label{sec:decoder}

We adopt the MMT decoder from M4C~\cite{hu2020iterative}, which
attends over the concatenated V and T~embeddings to generate
captions auto-regressively via the mixture distribution in
Eq.~(1). OcrPtrNet enables the model to copy OCR tokens directly,
essential for faithfully reproducing scene text.

\subsection{PhonoSTFG: Phonologically-Enhanced Fusion}
\label{sec:phonostfg}

\begin{figure*}[htbp]
  \centering
  \includegraphics[width=\textwidth]{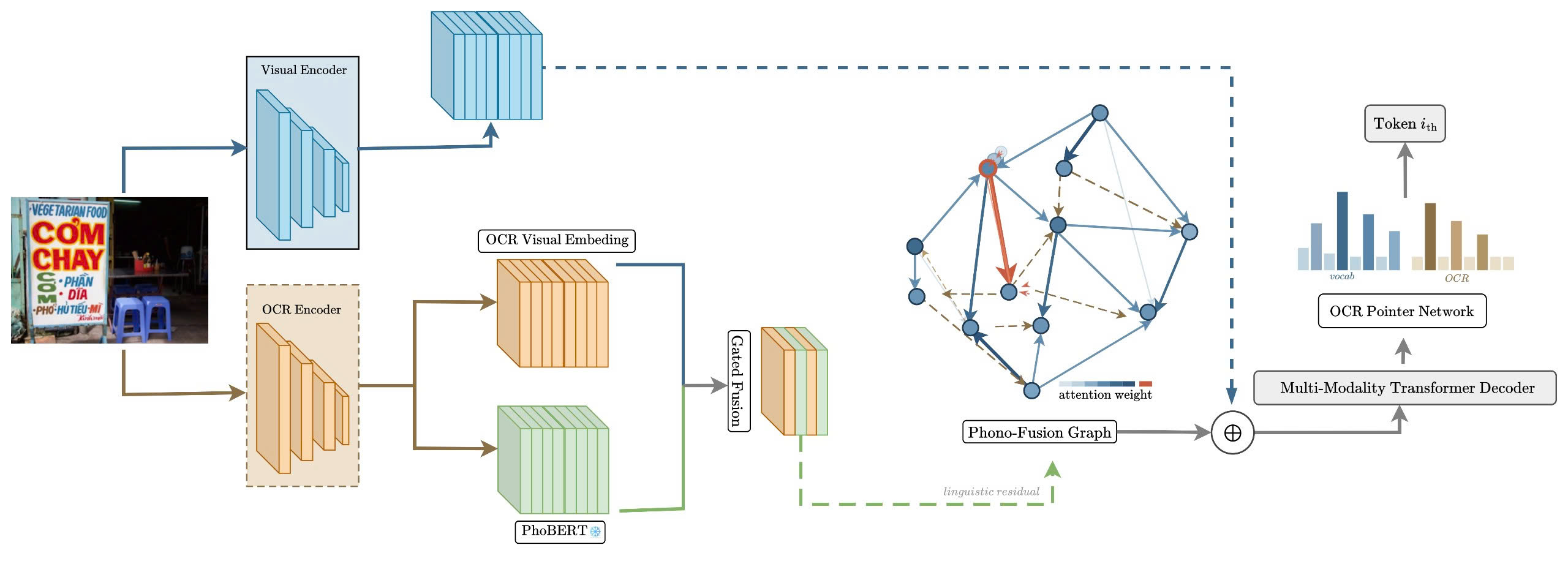}
  \caption{%
    \textbf{PhonoSTFG: Phonologically-Enhanced T$\to$T Fusion.}
    \textbf{Full pipeline:} the OCR token embedding is
    replaced by a dual-stream architecture combining a visual
    OCR stream ($\mathbf{v}^{\mathrm{vis}}$, L2-normalised
    recognition and detection features) and a frozen PhoBERT
    linguistic stream ($\mathbf{v}^{\mathrm{pho}}$,
    ${\sim}135$M parameters); a learned gate
    $\mathbf{g}{=}\sigma(W_g[\mathbf{v}^{\mathrm{vis}};
    \mathbf{v}^{\mathrm{pho}}])$ fuses them per-dimension.
    The graph is restricted to T$\to$T edges; V~node
    embeddings bypass graph processing and enter the MMT
    decoder directly. A learnable scalar residual
    $\sigma(\alpha)\,\mathbf{t}_i$ preserves pre-graph
    PhoBERT signal through all layers
    (Eq.~\eqref{eq:residual}).
  }
  \label{fig:phonostfg}
\end{figure*}

While HSTFG fuses all available modalities at the graph level
(Figure~\ref{fig:phonostfg}), cross-modal V$\leftrightarrow$T
edges create a duplicate fusion path: the MMT decoder already
attends jointly over all V and T~embeddings via cross-attention,
making graph-level V-T interaction architecturally redundant. Our
fusion topology analysis
(Section~\ref{sec:ablation}) empirically confirms this
redundancy. By contrast, T$\to$T edges serve a function the
decoder cannot fulfill: structured inter-token reasoning that
exploits spatial layout and linguistic relationships between OCR
tokens.

PhonoSTFG therefore restricts the graph to a
\textbf{T$\to$T-only topology}, and enriches these edges with
Vietnamese linguistic knowledge through four components.

\subsubsection{PhoBERT Dual-Stream OCR Embedding}
\label{sec:phobert_dualstream}

\paragraph{Motivation.}
HSTFG represents each OCR token using FastText, a static word
embedding that assigns a single fixed vector per word form
regardless of context. For Vietnamese---an isolating language
where identical syllables carry different meanings depending on
context (e.g., ``b\d{a}c'' as \textit{silver}
vs.\ ``b\d{a}c'' as \textit{ungrateful})---this is a fundamental
limitation. Moreover, OCR outputs frequently contain misspelled
tokens absent from FastText's vocabulary.

\paragraph{Dual-Stream Architecture.}
We replace FastText with a dual-stream architecture where each
stream captures a different aspect of the OCR token.

\paragraph{Stream 1: Visual OCR Embedding.}
Recognition features $\mathbf{r}_i \in \mathbb{R}^{256}$ encode
character shapes, while detection features
$\mathbf{d}_i \in \mathbb{R}^{256}$ encode the image region
containing text. We L2-normalise each vector to equalise scales
before concatenation:
\begin{equation}
\mathbf{v}_i^{\mathrm{vis}} = \mathrm{LN}\!\left(W_{\mathrm{vis}}
\left[\frac{\mathbf{r}_i}{\|\mathbf{r}_i\|_2};\,
\frac{\mathbf{d}_i}{\|\mathbf{d}_i\|_2}\right]\right)
\in \mathbb{R}^d
\end{equation}
where $W_{\mathrm{vis}} \in \mathbb{R}^{d \times 512}$.

\paragraph{Stream 2: PhoBERT Linguistic Embedding.}
Each token string $s_i$ is tokenized with the PhoBERT BPE
tokenizer, encoded by a frozen PhoBERT
encoder~\cite{nguyen2020phobert} ($\sim$135M parameters), and
mean-pooled over subword embeddings:
\begin{equation}
\mathbf{h}_i = \mathrm{MeanPool}\!\left(
\mathrm{PhoBERT}\!\left(\mathrm{Tokenize}(s_i)\right)
\right) \in \mathbb{R}^{768}
\end{equation}
\begin{equation}
\mathbf{v}_i^{\mathrm{pho}} =
\mathrm{LN}(W_{\mathrm{pho}}\, \mathbf{h}_i) \in \mathbb{R}^d
\end{equation}
where $W_{\mathrm{pho}} \in \mathbb{R}^{d \times 768}$.
PhoBERT (vinai/phobert-base-v2), pre-trained on 20\,GB of
Vietnamese text, provides contextual semantics and morphological
reasoning via BPE subwords. All PhoBERT parameters are frozen:
ViTextCaps spans diverse everyday domains (shops, streets,
products), and PhoBERT, pre-trained on similarly diverse
Vietnamese web text, already captures the relevant linguistic
distributions. Fine-tuning 135M parameters on our 15K-image
training set risks specialising these general representations to
training-set artefacts without meaningful improvement.

\paragraph{Gated Fusion.}
We require a mechanism to resolve per-dimension conflicts between
the two streams. Simple addition assumes equal contributions,
masking noise from unreliable OCR. Concatenation doubles
dimensionality and delegates balancing entirely to downstream
layers. Gated fusion provides input-dependent, per-dimension
weighting: each dimension independently selects the more
informative stream. Formally:
\begin{equation}
\mathbf{g}_i = \sigma\!\left(W_g\,
[\mathbf{v}_i^{\mathrm{vis}};\,\mathbf{v}_i^{\mathrm{pho}}]
+ \mathbf{b}_g\right) \in (0,1)^d
\end{equation}
\begin{equation}
\mathbf{f}_i = \mathbf{g}_i \odot \mathbf{v}_i^{\mathrm{vis}}
+ (1 - \mathbf{g}_i) \odot \mathbf{v}_i^{\mathrm{pho}}
\end{equation}
where $W_g \in \mathbb{R}^{d \times 2d}$,
$\mathbf{b}_g \in \mathbb{R}^d$ ($\sim$1.18M parameters),
and $\odot$ denotes element-wise multiplication. We hypothesise
that when OCR recognition is reliable, the gate favours PhoBERT
(low values); when OCR output is noisy, the gate shifts toward
the more stable visual stream (high values). The final T~node
embedding adds bounding box position:
\begin{equation}
\mathbf{t}_i = \mathbf{f}_i + \mathrm{LN}(W_{\mathrm{bbox}}\,
\mathbf{b}_i), \quad \mathbf{b}_i \in \mathbb{R}^4
\end{equation}

The dual-stream embeddings provide richer token-level
representations; we next address how to preserve this linguistic
information through graph processing.

\subsubsection{Linguistic Residual Preservation}
\label{sec:linguistic_residual}

PhoBERT produces high-quality Vietnamese embeddings, but after
$L$ graph attention layers with T$\to$T self-attention, the
original linguistic information may be diluted---a well-known
phenomenon termed over-smoothing, where repeated neighbourhood
aggregation drives node embeddings toward similar
representations~\cite{li2018deeper}.

We address this with a learnable scalar residual. Let
$\mathbf{t}$ denote the T~node embeddings before graph processing.
The final representation is:
\begin{equation}
\label{eq:residual}
\mathbf{t}_i^{\mathrm{final}} =
\mathrm{GraphLayers}(\mathbf{t}_i) + \sigma(\alpha)\, \mathbf{t}_i
\end{equation}
where $\alpha \in \mathbb{R}$ is a \textbf{single scalar
parameter} shared across all tokens and layers. We initialise
$\alpha_0 = 0.5$, the midpoint of the sigmoid input range, so
that $\sigma(0.5) \approx 0.62$ provides a moderate initial bias
toward preserving pre-graph representations. The sigmoid ensures
$\sigma(\alpha) \in (0,1)$: gradient-based optimisation increases
$\alpha$ if graph layers cause information loss, or decreases it
if the original embeddings become redundant. Unlike a fixed skip
connection (weight $= 1.0$), this mechanism lets the model
self-regulate residual strength during training.

With linguistic information preserved through the graph layers, we
now enrich T$\to$T attention with explicit Vietnamese phonological
knowledge.

\subsubsection{Vietnamese Phonological Attention Bias}
\label{sec:phonological_bias}

\begin{figure}[t]
    \centering
    \includegraphics[width=\linewidth]{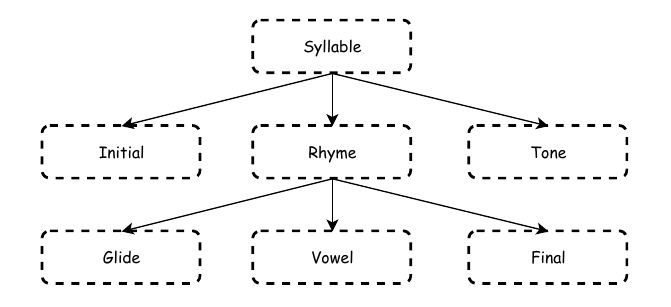}
\caption{Phonological structure of a Vietnamese syllable, following the template $C_1 w V C_2 T$ (onset, medial, nucleus, coda, and tone).}    \label{fig:syllable}
\end{figure}

\paragraph{Motivation.}
Spatial bias in T$\to$T attention encodes \textbf{geometric
relationships} between text regions but completely lacks
\textbf{linguistic relationships}---particularly important for
Vietnamese due to three characteristics of the
language~\cite{doan1977,thompson1965}:
\begin{enumerate}[nosep]
    \item \textbf{Fixed syllable structure.} Each Vietnamese
    syllable follows the strict template $C_1 w V C_2 T$
    (onset--medial--nucleus--coda--tone), with a closed inventory
    of 26~onsets, 2~medials, 23~nuclei, and 12~codas~\cite{doan1977}.
    This regularity enables precise rule-based phonological
    decomposition.
    \item \textbf{Six-tone system.} Vietnamese distinguishes six
    lexical tones that completely change word meaning
    (e.g., ``ma'' (ghost) $\neq$ ``m\`{a}'' (but) $\neq$
    ``m\'{a}'' (mother) $\neq$ ``m\~{a}'' (horse)).
    OCR systems frequently confuse visually similar tone marks.
    \item \textbf{Diacritic system.} Vietnamese uses 12~diacritics
    on vowels (\textit{\ohorn/o, \ecircumflex/e, \uhorn/u,
    \acircumflex/a, \abreve/a}, plus 6~tone marks). Diacritic
    confusion is the most common OCR error---e.g.,
    ``Tr\uhorn\`{o}ng'' (school) misread as
    ``Tr\uhorn\ohorn ng'' (a non-word).
\end{enumerate}
To our knowledge, no prior work integrates explicit phonological
features into vision-language fusion; existing uses are limited to
speech processing~\cite{li2020universal} and
transliteration~\cite{banchs-etal-2015-report}
(see Section~\ref{sec:related_linguistic} for a survey). We
address this gap by encoding Vietnamese phonological structure as
a pairwise attention bias.

\paragraph{Unified Dual-Bias Attention.}
We augment T$\to$T attention with a phonological bias added in
parallel with the existing spatial bias. The combined attention
score for head $h$ between tokens $i$ and $j$ is:
\begin{equation}
\label{eq:dual_bias}
A_{ij}^{(h)} =
\frac{{\mathbf{q}_i^{(h)}}^\top \mathbf{k}_j^{(h)}}{\sqrt{d_h}}
+ \underbrace{{w_{\mathrm{sp}}^{(h)}}^\top
  \phi(\mathbf{s}_{ij})}_{\text{spatial bias}}
+ \underbrace{{w_{\mathrm{ph}}^{(h)}}^\top
  \psi(\mathbf{p}_{ij})}_{\text{phonological bias}}
\end{equation}
where $\phi: \mathbb{R}^4 \xrightarrow{W_1^{\mathrm{sp}}}
\mathbb{R}^{32} \xrightarrow{\mathrm{ReLU}} \mathbb{R}^{32}$
and $\psi: \mathbb{R}^8 \xrightarrow{W_1^{\mathrm{ph}}}
\mathbb{R}^{32} \xrightarrow{\mathrm{ReLU}} \mathbb{R}^{32}$
are the shared MLP hidden layers for spatial and phonological
features respectively, and
$w_{\mathrm{sp}}^{(h)}, w_{\mathrm{ph}}^{(h)} \in \mathbb{R}^{32}$
are per-head output weights. The spatial features $\mathbf{s}_{ij}$
are computed as in Eq.~\eqref{eq:spatial_feat}; the phonological
features $\mathbf{p}_{ij}$ are described below.

Note that both spatial and phonological biases share the same MLP
architecture ($\mathbb{R}^{\cdot} \to \mathbb{R}^{32}
\to \mathbb{R}^H$) but operate on different input spaces
(4-d geometric vs.\ 8-d linguistic). Each attention head learns
independent output weights $w^{(h)}$, enabling
specialisation---e.g., one head may weight
$\psi_7$ (\texttt{base\_form\_match}) to attend to
likely OCR diacritic confusions, while another weights $\psi_4$
(\texttt{rhyme\_match}) to group phonologically related tokens.
The graph layer update from Eq.~\eqref{eq:graph_update} applies
unchanged, with $A_{ij}^{(h)}$ now given by
Eq.~\eqref{eq:dual_bias} in place of Eq.~\eqref{eq:spatial_attn}.

\paragraph{Pairwise Phonological Features.}
Each OCR token is decomposed into its phonological components
using standard Vietnamese phonotactic constraints~\cite{doan1977}.
The decomposition is non-trivial: it requires Unicode NFD
normalisation to separate base characters from combining
diacritics, followed by rule-based syllable parsing encoding
35+~phonotactic constraints. Algorithm~\ref{alg:phono} formalises
this procedure. For each pair $(t_i, t_j)$, it returns an
8-dimensional binary vector $\mathbf{p}_{ij} \in \{0,1\}^8$.

\begin{algorithm}[htbp]
\caption{Vietnamese Phonological Feature Extraction}
\label{alg:phono}
\small
\begin{algorithmic}[1]
\Require OCR token strings $\{s_i\}_{i=1}^{N_t}$
\Ensure $\mathbf{P} \in \{0,1\}^{N_t \times N_t \times 8}$

\Statex \textit{Stage 1: per-token analysis}
\For{each token $s_i$}
  \State $\hat{s}_i \gets \textsc{NfdNormalize}(s_i)$
  \State $\mathrm{VN}(i) \gets \textsc{IsVietnamese}(\hat{s}_i)$
  \Comment{35+ phonotactic rules~\cite{doan1977}}
  \If{$\mathrm{VN}(i)$}
    \State $\mathrm{tone}(i) \gets \textsc{ExtractTone}(\hat{s}_i)$
    \State $\mathrm{base}(i) \gets \textsc{StripTone}(\hat{s}_i)$
    \State $[\mathrm{C_1},\, w,\, V,\, \mathrm{C_2}] \gets
           \textsc{DecomposeSyllable}(\mathrm{base}(i))$
    \State $\mathrm{rhyme}(i) \gets (w,\, V,\, \mathrm{C_2})$
    \State $\mathrm{tc}(i) \gets \textsc{ToneClass}(\mathrm{tone}(i))$
  \EndIf
\EndFor

\Statex \textit{Stage 2: pairwise feature computation}
\For{each pair $(i,\, j)$}
  \If{$\mathrm{VN}(i)$ \textbf{and} $\mathrm{VN}(j)$}
    \State $p_1 \gets \mathbb{1}[\mathrm{C_1}(i) = \mathrm{C_1}(j)]$
    \State $p_2 \gets \mathbb{1}[V(i) = V(j)]$
    \State $p_3 \gets \mathbb{1}[\mathrm{C_2}(i) = \mathrm{C_2}(j)]$
    \State $p_4 \gets \mathbb{1}[\mathrm{rhyme}(i) = \mathrm{rhyme}(j)]$
    \State $p_5 \gets \mathbb{1}[\mathrm{tone}(i) = \mathrm{tone}(j)]$
    \State $p_6 \gets \mathbb{1}[\mathrm{tc}(i) = \mathrm{tc}(j)]$
    \State $p_7 \gets \mathbb{1}[\mathrm{base}(i) = \mathrm{base}(j)]$
    \Comment{detects tone-only OCR errors}
    \State $p_8 \gets 1$
  \Else
    \State $\mathbf{p}_{ij} \gets \mathbf{0}_8$
  \EndIf
  \State $\mathbf{P}[i,j] \gets [p_1,\ldots,p_8]^\top$
\EndFor
\State \Return $\mathbf{P}$
\end{algorithmic}
\end{algorithm}

Three helper functions have exact implementations:
\textsc{IsVietnamese} encodes 35+~phonotactic constraints (e.g.,
valid onset--nucleus combinations, impermissible codas) derived
from~\cite{doan1977}; \textsc{ExtractTone} reads the combining
tone diacritic via Unicode code-point lookup after NFD
normalisation; \textsc{DecomposeSyllable} parses the tone-stripped
string against a trie of valid Vietnamese syllable components.
Full implementation details are provided in the supplementary
material. All computations operate on OCR text strings only---not
on model weights---and are performed once before training, reused
across all 3~graph layers, incurring negligible overhead.

The 8-dimensional feature vector for pair $(t_i, t_j)$ is thus
$\mathbf{p}_{ij} = [p_1,\ldots,p_8]^\top \in \{0,1\}^8$,
structured in three groups as follows.
Algorithm~\ref{alg:phono} details the full extraction
procedure, including edge-case handling for non-Vietnamese tokens.

\texttt{Group A: Syllable Structure} (4 features). Each valid
Vietnamese token is decomposed into onset, medial, nucleus, and
coda:
\begin{align*}
p_1 &= \mathbb{1}[\mathrm{onset}(t_i) = \mathrm{onset}(t_j)]
\\
p_2 &= \mathbb{1}[\mathrm{nucleus}(t_i) = \mathrm{nucleus}(t_j)]
\\
p_3 &= \mathbb{1}[\mathrm{coda}(t_i) = \mathrm{coda}(t_j)]
\\
p_4 &= \mathbb{1}[\mathrm{rhyme}(t_i) = \mathrm{rhyme}(t_j)],
\quad
\mathrm{rhyme}(t) = \mathrm{med}(t) \oplus
\mathrm{nuc}(t) \oplus \mathrm{coda}(t)
\end{align*}

\texttt{Group B: Tone System} (2 features):
\begin{align*}
p_5 &= \mathbb{1}[\mathrm{tone}(t_i) = \mathrm{tone}(t_j)]
\quad \text{(extracted via Unicode NFD decomposition)}
\\
p_6 &= \mathbb{1}[\mathrm{toneclass}(t_i)
       = \mathrm{toneclass}(t_j)]
\end{align*}
where $\mathrm{toneclass} \in \{\text{\textit{b\`{a}ng}},
\text{\textit{tr\'{a}c}}\}$: \textit{b\`{a}ng} (level register:
ngang, huy\`{e}n) and \textit{tr\'{a}c} (oblique register:
s\'{a}c, h\~{o}i, ng\~{a}, n\d{a}ng).

\texttt{Group C: Diacritic System} (2 features):
\begin{align*}
p_7 &= \mathbb{1}[\mathrm{base}(t_i) = \mathrm{base}(t_j)]
\quad \text{where } \mathrm{base}(t)
\text{ strips all tone marks}
\\
p_8 &= \mathbb{1}[\mathrm{VN}(t_i) \wedge \mathrm{VN}(t_j)]
\end{align*}
$p_7$ is the most critical feature for OCR error detection: it
identifies token pairs that differ only in tone
(e.g., ``Tr\uhorn\`{o}ng'' and ``Tr\uhorn\ohorn ng'' both yield
base ``tr\uhorn ong''). $p_8$ gates the other features:
$p_1$--$p_7$ are meaningful only when both tokens are valid
Vietnamese syllables; for non-Vietnamese tokens, all 8~features
are zero. 
Phonological features depend solely on OCR text strings and are
computed once on CPU, incurring negligible overhead. The
phonological MLP adds only 552~parameters per layer (1,656~total
for 3~layers).

\paragraph{Distinction from PhoBERT.}
A natural question is whether phonological bias is redundant with
PhoBERT, since both encode Vietnamese linguistic knowledge. They
operate at fundamentally different granularities. PhoBERT maps
each token \textit{independently} to a contextual
embedding---it encodes what ``Tr\uhorn\`{o}ng'' means, but
produces separate 768-d vectors for ``Tr\uhorn\`{o}ng'' and
``Tr\uhorn\ohorn ng'' with no guaranteed structural relationship
between them. Phonological bias operates on \textit{token pairs}:
via Eq.~\eqref{eq:dual_bias}, it informs the attention mechanism
that these two tokens share base form ($p_7 = 1$) and rhyme
($p_4 = 1$) but differ in tone ($p_5 = 0$)---precisely the
pairwise structural signal needed to reason about OCR confusions.

Together, the four components form PhonoSTFG's linguistic fusion
pipeline:
\begin{equation}
\underbrace{\mathbf{f}_i}_{\text{gated dual-stream}}
\xrightarrow{+\,\mathrm{bbox}}
\underbrace{\mathbf{t}_i}_{\text{T node}}
\xrightarrow{\text{GraphLayers}(\cdot) + \sigma(\alpha)\mathbf{t}_i}
\underbrace{\mathbf{t}_i^{\mathrm{final}}}_{\text{residual output}}
\end{equation}
where graph layers implement Eq.~\eqref{eq:graph_update} with
dual-bias attention from Eq.~\eqref{eq:dual_bias}.

\section{Benchmark Setup} 
\label{sec:benchmark_setup}

All models are evaluated on ViTextCaps using the standard 7:1:2
train/dev/test split (10{,}996 / 1{,}641 / 3{,}092 images);
hyperparameters are selected on the development set and final
results reported on the test set.

\subsection{Evaluation Metrics}
\label{sec:metrics}

We adopt the standard evaluation protocol for scene-text image
captioning~\cite{sidorov2020textcaps}. All metrics are computed
using syllable-level (space-split) tokenization as established
in Section~\ref{sec:word_segmentation}.

\textbf{BLEU-$n$}~\cite{bleu} measures modified
$n$-gram precision with a brevity penalty:
\begin{equation}
\text{BLEU-}n = \text{BP} \cdot \exp\!\left(\frac{1}{n}
  \sum_{k=1}^{n} \log p_k\right)
\end{equation}
where $p_k$ is the clipped $k$-gram precision and
$\text{BP} = \min(1, e^{1 - r/c})$ penalizes short outputs
($c$ = candidate length, $r$ = reference length).
We report BLEU-1 and BLEU-4.

\textbf{CIDEr}~\cite{cider} computes
TF-IDF-weighted cosine similarity of $n$-gram vectors
between a candidate caption $c$ and reference set $S$:
\begin{equation}
\text{CIDEr}(c, S) = \frac{1}{|S|} \sum_{s \in S}
  \sum_{n=1}^{N} w_n \,
  \frac{\mathbf{g}^n(c) \cdot \mathbf{g}^n(s)}
       {\|\mathbf{g}^n(c)\|\,\|\mathbf{g}^n(s)\|}
\end{equation}
where $\mathbf{g}^n(\cdot)$ is the TF-IDF weighted $n$-gram
vector and $w_n = 1/N$ with $N=4$.
CIDEr rewards image-discriminative tokens and exhibits the
lowest sensitivity to tokenizer choice among standard metrics
(relative variance 31--36\%, Section~\ref{sec:word_segmentation}),
making it the \textbf{primary metric} for ViTextCaps.

\textbf{ROUGE-L}~\cite{rouge} measures longest common
subsequence (LCS) recall:
\begin{equation}
\text{ROUGE-L} = \frac{(1+\beta^2)\,R_{\text{lcs}}\,
  P_{\text{lcs}}}{R_{\text{lcs}} + \beta^2 P_{\text{lcs}}}
\end{equation}
where $R_{\text{lcs}}$ and $P_{\text{lcs}}$ are LCS-based
recall and precision, and $\beta$ is set to favour recall.

\textbf{METEOR}~\cite{meteor} computes unigram
$F$-score with synonym-aware alignment, providing a
complementary view of semantic overlap that is insensitive
to exact surface form.

\subsection{Comparison Baselines}
\label{sec:baselines}

All models in our comparison solve the same conditional
generation problem: at each decoding step $t$, predict
$y_t$ from visual features $\{\mathbf{v}_i\}$, OCR token
representations $\{\mathbf{t}_j\}$, and partial output
$y_{<t}$. We unify them under the encoder--fusion--decoder
framework defined by three components: the OCR encoder
$f_T$, the fusion operator $\mathcal{F}$, and the decoding
distribution:
\begin{equation}
p(y_t \mid y_{<t}, \mathbf{z}) =
  (1-\lambda_t)\,p_{\mathrm{vocab}}(y_t \mid \mathbf{h}_t)
  + \lambda_t\,p_{\mathrm{ocr}}(y_t \mid \mathbf{h}_t,
  \{\mathbf{t}_j\})
\label{eq:copy_mixture_unified}
\end{equation}
where $\mathbf{z} = \mathcal{F}(\{\mathbf{v}_i\},
\{\mathbf{t}_j\})$ is the fused context, $\mathbf{h}_t$
is the decoder state attending over $\mathbf{z}$, and
$\lambda_t$ is a learned copy weight. The baselines differ
exclusively in $(f_T, \mathcal{F}, \lambda_t)$, enabling
controlled attribution of performance differences.
We organize baselines into four groups based on their
text-reading capability and training regime.

\paragraph{No OCR fusion
($\mathcal{T} = \emptyset$, $\lambda_t \equiv 0$).}
\textbf{BUTD}~\cite{butd},
\textbf{AOA}~\cite{aoa}, and
\textbf{M2}~\cite{m2} set $f_T \equiv 0$
and fix $\lambda_t = 0$, reducing
Eq.~\eqref{eq:copy_mixture_unified} to:
\begin{equation}
p(y_t \mid y_{<t}, \mathbf{z}) =
  p_{\mathrm{vocab}}(y_t \mid \mathbf{h}_t), \quad
\mathbf{z} = \mathcal{F}_V(\{\mathbf{v}_i\})
\end{equation}
They differ only in $\mathcal{F}_V$: BUTD uses top-down
additive attention over region features; AOA augments this
with a secondary attention filter that suppresses irrelevant
attended content; M2 replaces single-level cross-attention
with meshed connections across all encoder layers, enriched
by persistent memory vectors. Their inclusion establishes
a necessary condition: scene-text captioning on ViTextCaps
requires $\lambda_t > 0$, i.e., the ability to copy OCR
tokens not in the fixed vocabulary.

\paragraph{Flat cross-modal fusion
($\lambda_t > 0$, $\mathcal{F} = $ full attention).}
\textbf{M4C}~\cite{hu2020iterative} activates the full
copy mechanism with $f_T$ given by
Eq.~\eqref{eq:ocr_feat_m4c} and implements $\mathcal{F}$
as $L$ stacked transformer layers with \textit{full
cross-attention} among all visual tokens, OCR tokens,
and partial answer tokens:
\begin{equation}
\mathbf{z} = \mathrm{Transformer}^L\!\left(
  \bigl[\mathbf{v}_1,\ldots,\mathbf{v}_{N_v},\,
         \mathbf{t}_1,\ldots,\mathbf{t}_{N_t},\,
         \mathbf{e}_{y_{<t}}\bigr]\right)
\end{equation}
This \textit{flat} fusion applies uniform
$O\!\left((N_v+N_t)^2\right)$ attention without
distinguishing intra-modal from cross-modal interactions.
M4C is our primary baseline: our proposed methods share
its decoder and copy mechanism
(Eq.~\eqref{eq:copy_mixture_unified}), differing only
in $f_T$ and $\mathcal{F}$.

\paragraph{Heterogeneous graph fusion.}
HSTFG and PhonoSTFG replace flat attention with a typed
graph fusion operator $\mathcal{F}_\mathcal{G}$ defined
over a heterogeneous graph $\mathcal{G} =
(\mathcal{V} \cup \mathcal{T}, \mathcal{E})$ with
configurable edge types $\mathcal{E} \subseteq
\{V{\to}T,\,T{\to}V,\,T{\to}T\}$
(Section~\ref{sec:hstfg}). Unlike M4C's flat attention,
$\mathcal{F}_\mathcal{G}$ enforces modality-aware message
passing with learned spatial attention bias and confidence
gating. PhonoSTFG further specializes $f_T$ to a
dual-stream PhoBERT encoder and enriches the T$\to$T
attention with Vietnamese phonological
bias---the only model in the comparison where $f_T$
encodes language-specific structural knowledge
(Section~\ref{sec:phonostfg}).

\paragraph{LMMs.}
\textbf{GPT-4o}~\cite{openai2024gpt4o} instantiates a
fundamentally different regime: $\mathcal{F}$ is a frozen
large language model whose copy behavior is governed by
the LM prior rather than learned from ViTextCaps, and
$\lambda_t$ is implicit rather than explicitly trained.
This model tests whether large-scale pre-training can
substitute for task-specific fusion
design---a question whose answer directly characterizes
the fluency--fidelity trade-off reported in
Section~\ref{sec:main_results}.

\subsection{Implementation Details}
\label{sec:implementation}

Both HSTFG and PhonoSTFG build on the MMT
decoder~\cite{hu2020iterative} and share the same training
configuration. Visual features are extracted using a
VinVL-based Faster R-CNN backbone~\cite{zhang2021vinvl},
and OCR tokens are detected by
SwinTextSpotter~\cite{swintextspotter}; both feature extractors
are fixed during training. Models are trained end-to-end
with cross-entropy loss using the Adam
optimizer~\cite{adam} with $\beta_1 = 0.9$,
$\beta_2 = 0.98$, and a Noam learning rate schedule with
10{,}000 warmup steps and a peak learning rate of 1.0.
Training uses a batch size of 64 with early stopping
(patience = 5) based on validation CIDEr. 

In PhonoSTFG, the PhoBERT
encoder~\cite{nguyen2020phobert} (vinai/phobert-base-v2,
$\sim$135M parameters) is frozen throughout training.
Table~\ref{tab:param_summary} details the new trainable
parameters introduced by PhonoSTFG relative to HSTFG;
the net increase is only $\sim$1.55M parameters, as
the removal of the FastText projection ($-$624K) offsets
most of the addition.

\begin{table}[t]
\centering
\caption{New trainable parameters introduced by PhonoSTFG
relative to HSTFG. PhoBERT parameters are frozen and not
counted toward the trainable total.}
\label{tab:param_summary}
\small
\begin{tabular}{lr}
\toprule
\textbf{Component} & \textbf{Parameters} \\
\midrule
\texttt{linear\_ocr\_visual} ($512 \to 768$)       & 393K \\
\texttt{linear\_phobert} ($768 \to 768$)            & 590K \\
\texttt{GatedFusion.gate\_linear} ($1536 \to 768$)  & 1{,}180K \\
LayerNorms (\texttt{ocr\_visual}, \texttt{phobert}) & 3K \\
\texttt{residual\_alpha} (scalar)                   & 1 \\
\texttt{phonological\_mlp} $\times$ 3 layers        & 1.7K \\
\midrule
Total new trainable  & $\sim$2.17M \\
PhoBERT (frozen)     & $\sim$135M \\
Removed \texttt{linear\_ocr\_feat} ($812 \to 768$)  & $-$624K \\
\midrule
\textbf{Net increase} & $\sim$\textbf{1.55M} \\
\bottomrule
\end{tabular}
\end{table}

Detailed feature extraction pipelines for visual and OCR 
representations, together with a word segmentation 
sensitivity analysis, are provided in 
Appendix~\ref{app:setup}.

\section{Experiments} 
\label{sec:experiments}

\subsection{Main Results}
\label{sec:main_results}

\begin{table*}[ht]
\centering
\caption{Results on the ViTextCaps test set. All metrics use
syllable-level tokenization
(Section~\ref{sec:word_segmentation}). Subscripts denote 95\%
bootstrap CI half-widths for trained models. \textbf{Bold}:
best among trained models. $\dagger$: our proposed models.
}
\label{tab:main_results}
    \resizebox{\textwidth}{!}{
    \begin{tabular}{l ccccc}
    \toprule
    \textbf{Model}
      & \textbf{B-1} & \textbf{B-4} & \textbf{MET}
      & \textbf{R-L} & \textbf{CIDEr} \\
    \midrule
    \multicolumn{6}{l}{\textit{Without copy mechanism}} \\
    AOA~\cite{aoa}
      & $0.000_{\pm.000}$ & $0.000_{\pm.000}$ & $0.006$
      & $0.031_{\pm.001}$ & $0.007_{\pm.001}$ \\
    BUTD~\cite{butd}
      & $0.409_{\pm.004}$ & $0.000_{\pm.000}$ & $0.253$
      & $0.196_{\pm.003}$ & $0.048_{\pm.003}$ \\
    M2~\cite{m2}
      & $0.413_{\pm.004}$ & $0.000_{\pm.000}$ & $0.225$
      & $0.151_{\pm.003}$ & $0.026_{\pm.002}$ \\
    \midrule
    \multicolumn{6}{l}{\textit{With copy mechanism}} \\
    M4C~\cite{hu2020iterative}
      & $0.244_{\pm.003}$ & $0.087_{\pm.002}$ & $0.172$
      & $0.221_{\pm.002}$ & $0.594_{\pm.016}$ \\
    HSTFG$^\dagger$
      & $0.222_{\pm.002}$ & $0.085_{\pm.002}$ & $0.164$
      & $0.219_{\pm.002}$ & $0.640_{\pm.017}$ \\
    PhonoSTFG$^\dagger$
      & $\mathbf{0.251}_{\pm.003}$ & $\mathbf{0.098}_{\pm.002}$
      & $\mathbf{0.174}$ & $\mathbf{0.227}_{\pm.002}$
      & $\mathbf{0.646}_{\pm.017}$ \\
    \midrule
    \multicolumn{6}{l}{\textit{Reference points}} \\
    Human
      & $0.300$ & $0.118$ & $0.218$ & $0.272$ & $0.808$ \\
    GPT-4o~\cite{openai2024gpt4o}
      & $0.283$ & $0.106$ & $0.282$ & $0.255$ & $0.140$ \\
    \bottomrule
    \end{tabular}
    }
\end{table*}

\begin{figure*}[ht]
\centering
\includegraphics[width=\textwidth]{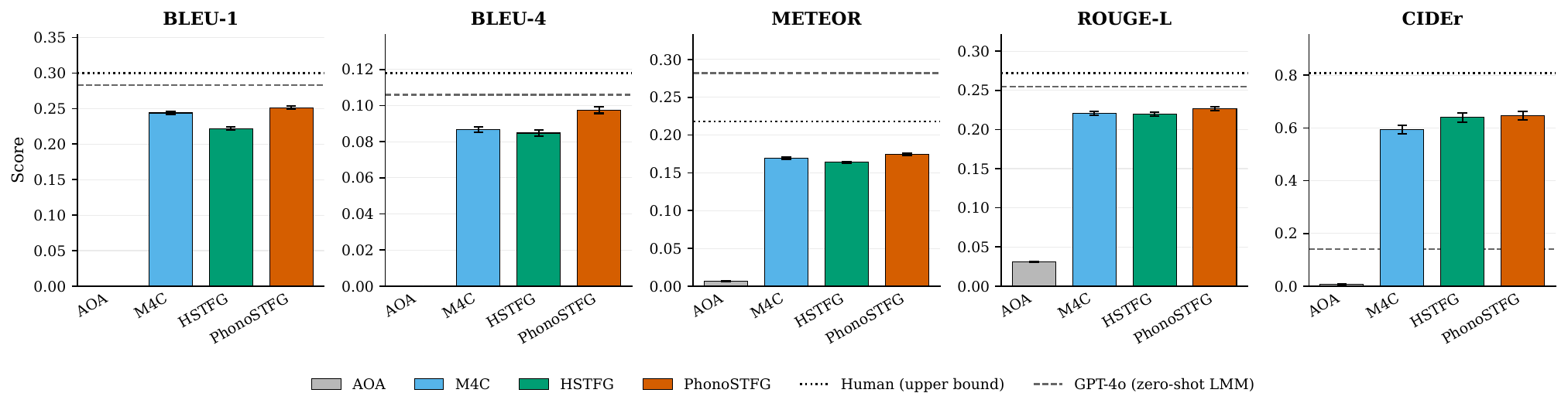}
\caption{Main results on the ViTextCaps test set. Bars show corpus
scores for the four trained copy-mechanism-aware models
(AOA, M4C, HSTFG, PhonoSTFG) with 95\% bootstrap confidence
intervals ($n{=}1{,}000$ resamples) for BLEU-1, BLEU-4, ROUGE-L,
and CIDEr; METEOR is corpus-level. Dotted and dashed horizontal
lines mark the human upper bound and the zero-shot GPT-4o
reference. Non-overlapping intervals between M4C and both HSTFG
and PhonoSTFG on CIDEr indicate statistical significance without
p-value annotation. BUTD and M2 are included in
Table~\ref{tab:main_results} but omitted from the figure
(near-zero BLEU-4 and CIDEr).}
\label{fig:main_results}
\end{figure*}

Table~\ref{tab:main_results} and Figure~\ref{fig:main_results}
report performance on the ViTextCaps test set with 95\% bootstrap
confidence intervals ($n = 1{,}000$ resamples) for trained models. The results establish four following findings.

\textbf{The copy mechanism is a necessary condition for ViTextCaps.}
Models without copy mechanisms (AOA, BUTD, M2) achieve near-zero
CIDEr ($\leq 0.048$) and BLEU-4 ($= 0.000$) despite BUTD and M2
producing high BLEU-1 ($0.41$) and METEOR ($0.25$). This
dissociation---high unigram overlap but zero phrase-level
overlap---indicates that these models generate fluent Vietnamese
descriptive text but fail entirely to reproduce scene-text tokens,
confirming that ViTextCaps requires the pointer-copy mechanism
(Eq.~\ref{eq:copy_mixture_unified}).

\textbf{Graph-based fusion significantly outperforms flat attention.}
Both HSTFG and PhonoSTFG achieve substantially higher CIDEr than
M4C ($0.640$ / $0.646$ vs.\ $0.594$), with paired bootstrap tests
confirming significance at $p < 0.01$ ($\Delta = +0.046$ and
$+0.052$ respectively). PhonoSTFG leads on all surface-form
metrics. Since all three share the same decoder and copy mechanism,
the gains are attributable solely to the heterogeneous graph fusion
operator $\mathcal{F}_\mathcal{G}$ (Section~\ref{sec:hstfg}).

\textbf{PhonoSTFG and HSTFG are not significantly different on
the full test set.}
The CIDEr gap ($\Delta = +0.006$, $p = 0.71$) does not reach
statistical significance. The confidence intervals for PhonoSTFG
($0.646_{\pm.017}$) and HSTFG ($0.640_{\pm.017}$) overlap
substantially, while both are clearly separated from M4C
($0.594_{\pm.016}$). PhonoSTFG's phonological bias provides
condition-specific gains analyzed in \ref{app:stratified}.

\textbf{A striking fluency--fidelity trade-off separates LMMs from
trained models.}
GPT-4o achieves the highest METEOR ($0.282$) among all models,
approaching human-level fluency, yet its CIDEr ($0.140$) is
$4.6\times$ lower than PhonoSTFG. The gap reflects not a failure
of visual understanding or language generation, but a fundamental
absence of text selection capability: GPT-4o describes the visual
scene adequately but lacks the pointer-copy mechanism needed for
selective OCR token integration. 



\subsection{Analysis}
\label{sec:ablation}

\subsubsection{Fusion Topology Ablation}

\begin{table*}[ht]
\centering
\caption{Ablation study on fusion topology
analysis within HSTFG (validation set); delta relative to the
fully connected reference (shaded). \textbf{Bold}: primary metric extremes within each part. Inline deltas:
{\scriptsize\textcolor{teal!80!black}{$\uparrow$}} improvement,
{\scriptsize\textcolor{red!70!black}{$\downarrow$}} degradation.}
\label{tab:ablation_topology}
\small
\resizebox{\textwidth}{!}{
\begin{tabular}{l cccc}
\toprule
\textbf{Configuration}
  & \textbf{CIDEr} & \textbf{B-4}
  & \textbf{METEOR} & \textbf{R-L} \\
\midrule
\multicolumn{5}{l}{\textit{Edge type ablations}} \\
\quad Fully connected
  (V$\!\to\!$T\,+\,T$\!\to\!$V\,+\,T$\!\to\!$T) [ref.]
  & .6399 & .0848 & .1638 & .2194 \\
\quad $-$ T$\to$V\;\;(V$\!\to\!$T\,+\,T$\!\to\!$T)
  & \textbf{.6862}\pos{.046} & \textbf{.0967}\pos{.012}
  & \textbf{.1684}\pos{.005} & \textbf{.2310}\pos{.012} \\
\quad $-$ V$\to$T\;\;(T$\!\to\!$V\,+\,T$\!\to\!$T)
  & .6720\pos{.032} & .0903\pos{.006}
  & .1646\pos{.001} & .2262\pos{.007} \\
\quad $-$ Confidence gate
  & .6563\pos{.016} & .0928\pos{.008}
  & .1646\pos{.001} & .2202\pos{.001} \\
\quad $-$ Spatial bias
  & \textbf{.5664}\nega{.074} & .0785\nega{.006}
  & .1604\nega{.003} & .2190\nega{.000} \\
\quad V$\!\to\!$T\,+\,T$\!\to\!$V\;(no T$\!\to\!$T)
  & .5648\nega{.075} & \textbf{.0846}\nega{.000}
  & .1665\pos{.003} & \textbf{.2107}\nega{.009} \\[2pt]
\multicolumn{5}{l}{\textit{Cross-modal vs.\ intra-modal}} \\
\quad T$\to$T only (no visual in graph)
  & .6419\pos{.002} & .0887\pos{.004}
  & .1583\nega{.006} & .2164\nega{.003} \\[2pt]
\multicolumn{5}{l}{\textit{Graph depth}} \\
\quad 1 layer
  & .6326\nega{.007} & .0935\pos{.009}
  & .1667\pos{.003} & .2210\pos{.002} \\
\quad 2 layers
  & .6154\nega{.025} & \textbf{.0945}\pos{.010}
  & \textbf{.1706}\pos{.007} & .2192\nega{.000} \\
\quad 3 layers \textnormal{[adopted]}
  & .6399 & .0848 & .1638 & .2194 \\
\quad 4 layers
  & .6335\nega{.006} & .0880\pos{.003}
  & .1661\pos{.002} & \textbf{.2227}\pos{.003} \\
\bottomrule
\end{tabular}
}
\end{table*}

\begin{figure*}[htbp]
\centering
\includegraphics[width=\textwidth]{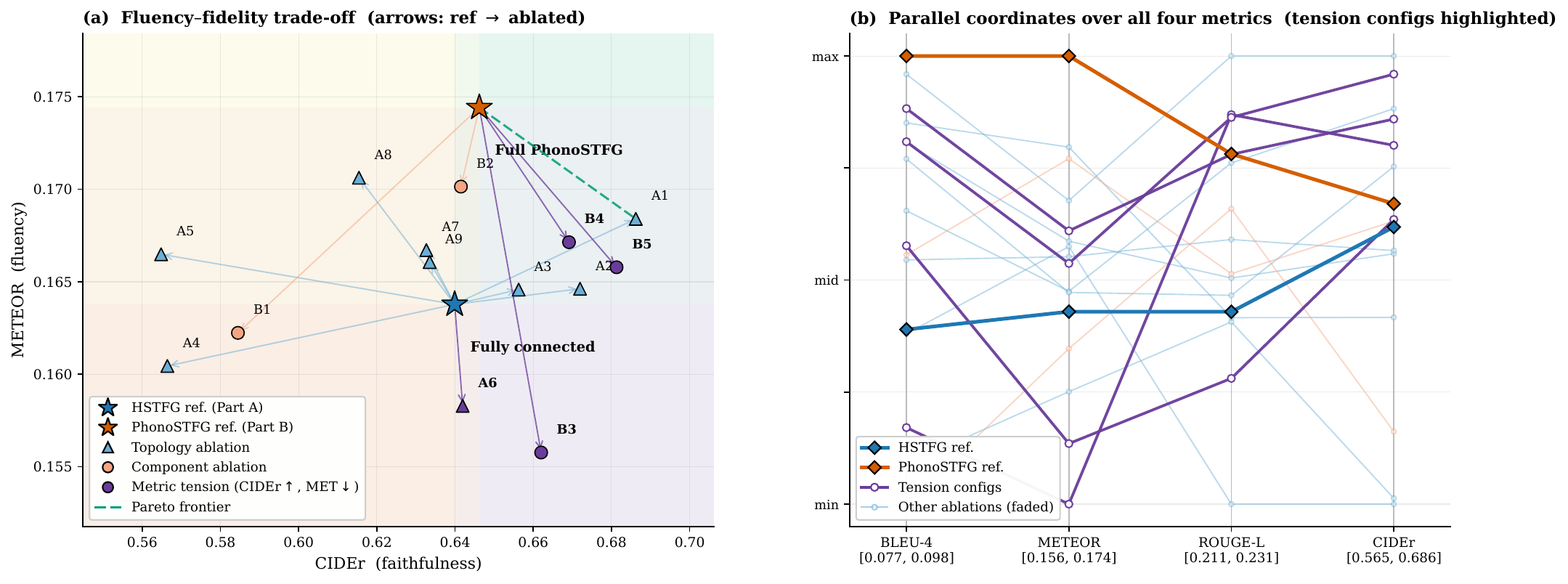}
\caption{Ablation analysis of the fusion architecture.
\textbf{(a)} Each configuration plotted in (CIDEr, METEOR) space;
arrows run from each reference configuration (stars) to each
ablation. Quadrants are anchored at the two reference points;
the lower-right quadrant (purple) marks configurations where
CIDEr increases but METEOR decreases (metric-tension region).
A Pareto frontier (green dashed) traces non-dominated
configurations. \textbf{(b)} Parallel coordinates over the four
metrics; reference polylines are bold, tension-exhibiting Part~B
ablations ($-$PhoBERT, $-$Gated fusion, $-$Residual preserv.) are
highlighted in purple; other ablations are faded. Both panels
read from the same configuration set as
Table~\ref{tab:ablation_topology} and Table \ref{tab:ablation_component}}
\label{fig:ablation_tradeoff}
\end{figure*}

We conduct two analyses to disentangle architectural contributions.
The fusion topology analysis systematically compares graph
connectivity configurations within HSTFG. The component analysis
evaluates each linguistic component introduced in PhonoSTFG.
Both analyses are reported on the test set using cross-entropy
training with early stopping (patience = 5) on validation CIDEr.
Table~\ref{tab:ablation_topology} presents both analyses in a unified view;
deltas are computed relative to the respective reference
configuration of each section, and
Figure~\ref{fig:ablation_tradeoff} visualizes the same configurations
in fluency--fidelity space (CIDEr $\times$ METEOR) together with a
parallel-coordinates view over all four metrics.

\textbf{T$\to$T edges and spatial bias are critical.}
Removing T$\to$T edges causes the largest CIDEr drop ($-0.075$),
consistent across all metrics. T$\to$T edges enable structured
inter-token reasoning between OCR tokens, essential for Vietnamese
scene text where adjacent syllables frequently form multi-word
compounds (Section~\ref{sec:word_segmentation}). Spatial bias
removal causes a similar drop ($-0.074$ CIDEr), confirming that
geometric relationships between text regions are a necessary
inductive bias that content-based attention alone cannot recover.

\textbf{Cross-modal edges are redundant.}
Removing either V$\leftrightarrow$T edge type improves CIDEr
and ROUGE-L consistently: $+0.046$ without T$\to$V, $+0.032$
without V$\to$T. The T$\to$T-only configuration matches the
fully connected topology on CIDEr ($0.642$ vs $0.640$) while
improving BLEU-4, confirming that the MMT decoder handles
cross-modal fusion via its joint self-attention, making
graph-level V$\leftrightarrow$T interaction redundant.
\textbf{Full connectivity in heterogeneous graphs is not always
beneficial}; topology should be validated empirically.

\textbf{The confidence gate provides no benefit under spatial bias.}
Removing the confidence gate improves all metrics, indicating
that spatial bias already encodes sufficient geometric context.
PhonoSTFG therefore adopts the T$\to$T-only topology without
a confidence gate, and 3 graph layers as the optimal depth.

\subsubsection{Component Ablation}
\label{sec:enhanced_ablation}

\begin{table*}[ht]
\centering
\caption{Component ablation study of PhonoSTFG (test set); delta relative to the full model (shaded). \textbf{Bold}: primary metric extremes within
each part. Inline deltas:
{\scriptsize\textcolor{teal!80!black}{$\uparrow$}} improvement,
{\scriptsize\textcolor{red!70!black}{$\downarrow$}} degradation.}
\label{tab:ablation_component}
\small
\resizebox{\textwidth}{!}{
\begin{tabular}{l cccc}
\toprule
\textbf{Configuration}
  & \textbf{CIDEr} & \textbf{B-4}
  & \textbf{METEOR} & \textbf{R-L} \\
\midrule
\multicolumn{5}{l}{\textit{Unanimous impact}} \\
\quad Full PhonoSTFG [ref.]
  & .6462 & .0976 & .1744 & .2266 \\
\quad $-$ Spatial bias
  & \textbf{.5845}\nega{.062} & \textbf{.0766}\nega{.021}
  & \textbf{.1622}\nega{.012} & \textbf{.2241}\nega{.003} \\
\quad $-$ Phonological bias
  & .6415\nega{.005} & .0883\nega{.009}
  & .1701\nega{.004} & .2211\nega{.006} \\[2pt]
\multicolumn{5}{l}{\textit{Metric tension
  (CIDEr\,$\uparrow$,\;fluency\,$\downarrow$)}} \\
\quad $-$ PhoBERT
  & .6620\pos{.016} & .0802\nega{.017}
  & .1558\nega{.019} & .2283\pos{.002} \\
\quad $-$ Gated fusion
  & .6691\pos{.023} & .0951\nega{.003}
  & .1671\nega{.007} & .2265\nega{.000} \\
\quad $-$ Residual preserv.
  & \textbf{.6813}\pos{.035} & .0936\nega{.004}
  & .1658\nega{.009} & .2282\pos{.002} \\
\bottomrule
\end{tabular}
}
\end{table*}

Table~\ref{tab:ablation_component} reveals a clear structural
pattern: the five components split into two distinct groups by
their metric signature.

\paragraph{Spatial bias.}
The only component with unanimous, large-magnitude impact across
all four metrics (CIDEr $-0.062$, BLEU-4 $-0.021$, METEOR
$-0.012$, ROUGE-L $-0.003$). Without spatial bias, the model
cannot distinguish OCR tokens that are semantically similar but
appear in different image regions. Its essentiality holds
regardless of graph topology, as confirmed by both Part A and B.

\paragraph{Phonological bias.}
Also unanimous across all four metrics (all negative), confirming
a genuine contribution despite the modest absolute magnitude
(CIDEr $-0.005$, BLEU-4 $-0.009$, METEOR $-0.004$, ROUGE-L
$-0.006$). The small absolute value is expected: phonological
reasoning targets only Vietnamese syllable pairs at risk of
diacritic confusion, a subset of all token-pair interactions.
Its contribution is more pronounced in diacritic-specific
behavior (Section~\ref{sec:error_propagation}).

\paragraph{PhoBERT, gated fusion, and residual preservation.}
These three components share a consistent metric tension signature:
removing each one \textit{increases} CIDEr ($+0.016$, $+0.023$,
$+0.035$ respectively) while simultaneously \textit{decreasing}
BLEU-4 and METEOR. This pattern is not a sign of architectural
redundancy but reflects a fundamental distinction between what
these metrics reward.

CIDEr's TF-IDF weighting assigns disproportionately high scores
to rare, image-discriminative tokens---store names, phone numbers,
and addresses that appear in few captions and are therefore
upweighted. Without PhoBERT's linguistic prior, the model loses
the capacity to paraphrase and instead copies these rare OCR tokens
more aggressively, boosting CIDEr at the cost of grammatical
coherence. BLEU-4 and METEOR, by contrast, reward $n$-gram
precision and semantic recall across the full caption---both
of which benefit from PhoBERT's contextual representations even
when exact OCR token reproduction decreases. The same mechanism
applies to gated fusion and residual preservation: removing either
forces the model toward a more copy-heavy strategy that is
CIDEr-optimal but linguistically impoverished.

This tension is an empirical demonstration that no single metric
captures the full quality spectrum of Vietnamese scene-text
captioning. We retain all three components because downstream
caption utility---the ability to generate grammatically correct,
semantically coherent Vietnamese---requires linguistic grounding
that CIDEr alone does not reward. The metric tension itself
provides a concrete argument for the multi-metric evaluation
framework introduced in Section~\ref{sec:metrics}, where
BLEU-4, METEOR, and the text-specific diagnostic measures
(THR, NF, OTR) collectively characterize model behavior
beyond what any individual metric captures.

Stratified performance analysis across OCR confidence, 
text density, domain, and visual complexity is provided 
in Appendix~\ref{app:stratified}. Qualitative 
analysis including error taxonomy and cross-model 
comparisons is provided in Appendix~\ref{app:qualitative}.

\subsubsection{Metric Sensitivity Analysis}
\label{sec:metric_sensitivity}

Standard captioning metrics may behave differently on scene-text
data. We analyze cross-metric correlations and sensitivity across
22 model configurations (main models, ablation variants, and LMM
baselines).

\begin{table}[t]
\centering
\caption{Cross-metric correlations across 22 model configurations
including ablation variants. All correlations statistically
significant ($p < 0.05$).}
\label{tab:metric_correlation}
\small
\begin{tabular}{lcc}
\toprule
\textbf{Metric Pair} & \textbf{Pearson $r$} & \textbf{Kendall $\tau$} \\
\midrule
BLEU-4 vs.\ ROUGE-L & 0.895 & 0.397 \\
BLEU-4 vs.\ CIDEr   & 0.765 & 0.467 \\
CIDEr vs.\ ROUGE-L  & 0.626 & 0.371 \\
\bottomrule
\end{tabular}
\end{table}

Table~\ref{tab:metric_correlation} reveals that BLEU-4 and ROUGE-L
are strongly correlated ($r = 0.895$), while CIDEr shows only
moderate correlation with both ($r = 0.765$ and $0.626$
respectively). The weaker Kendall $\tau$ ($0.37$--$0.47$) indicates
that the three metrics frequently disagree on model rankings. This
divergence is most striking for LMMs: GPT-4o ranks 1st in BLEU-4
and ROUGE-L but 20th in CIDEr (out of 22 configurations including
ablation variants), motivating the use of CIDEr as the primary
metric for ViTextCaps.

\begin{table}[htbp]
\centering
\caption{Relative metric change (\%) when removing each component
from PhonoSTFG. ROUGE-L is insensitive to meaningful architectural
changes; only spatial bias removal produces sign-consistent changes
across all four metrics.}
\label{tab:metric_sensitivity}
\small
\begin{tabular}{lcccc}
\toprule
\textbf{Removed Component}
  & $\boldsymbol{\Delta}$\textbf{B-4}
  & $\boldsymbol{\Delta}$\textbf{CIDEr}
  & $\boldsymbol{\Delta}$\textbf{MET}
  & $\boldsymbol{\Delta}$\textbf{R-L} \\
\midrule
$-$ Gated fusion      & $-2.6\%$  & $+3.5\%$  & $-4.2\%$  & $-0.0\%$ \\
$-$ Residual preserv. & $-4.1\%$  & $+5.4\%$  & $-4.9\%$  & $+0.7\%$ \\
$-$ Phonological bias & $-9.5\%$  & $-0.7\%$  & $-2.5\%$  & $-2.4\%$ \\
$-$ PhoBERT           & $-17.8\%$ & $+2.4\%$  & $-10.7\%$ & $+0.8\%$ \\
$-$ Spatial bias      & $-21.5\%$ & $-9.5\%$  & $-7.0\%$  & $-1.1\%$ \\
\bottomrule
\end{tabular}
\end{table}

Table~\ref{tab:metric_sensitivity} highlights three patterns.
BLEU-4 and CIDEr can disagree in sign: removing PhoBERT decreases
BLEU-4 by $17.8\%$ and METEOR by $10.7\%$ but increases CIDEr by
$2.4\%$, because BLEU-4 and METEOR reward $n$-gram precision and
semantic recall (both favoring PhoBERT-enhanced fluency) while
CIDEr rewards image-discriminative tokens (favoring scene-text
selection). ROUGE-L changes by at most $2.4\%$ across all
ablations---making it a poor choice as the sole evaluation metric.
Spatial bias is the only component where all four metrics agree
on the direction of change (all negative), reinforcing the need
for multi-metric evaluation in scene-text captioning.


\section{Conclusion}
\label{sec:conclusion}
We have addressed the problem of multimodal information fusion for
Vietnamese scene-text image captioning---a task that requires
integrating visual, textual, and linguistic information streams
within a language that poses unique fusion challenges.

Our linguistic analysis of \textbf{ViTextCaps}, the first
large-scale Vietnamese scene-text captioning dataset (15{,}729
images, 74{,}970 captions), establishes why language-agnostic
fusion fails for Vietnamese: 52.8\% of the caption vocabulary is
at risk of diacritic collision, and OCR errors are dominated by
tone substitution that standard metrics treat as generic token
mismatches.

We proposed \textbf{HSTFG}, a heterogeneous graph fusion framework
with learned spatial attention bias, and demonstrated through
topology analysis that cross-modal graph edges are harmful for
scene-text fusion---a finding with implications for heterogeneous
graph design beyond Vietnamese. Building on this insight, we
proposed \textbf{PhonoSTFG}, which specializes graph fusion for
Vietnamese through three mechanisms: (1)~dual-stream gated fusion
that dynamically balances visual and PhoBERT linguistic OCR
representations, (2)~Vietnamese phonological attention bias---the
first integration of formal phonological knowledge into
attention-based fusion for scene-text captioning, and
(3)~T$\to$T-only graph topology that focuses linguistic reasoning
between OCR tokens.

Experiments show that both methods significantly outperform
existing baselines on ViTextCaps (HSTFG: CIDEr 0.640; PhonoSTFG:
CIDEr 0.646 vs.\ M4C: 0.594), and substantially outperform GPT-4o
on CIDEr ($0.140$) despite its superior language fluency---a
striking \textbf{fluency--fidelity trade-off} that confirms
task-specific fusion cannot be replaced by general-purpose
pre-training. A human performance baseline on a randomly sampled
subset of the test set (CIDEr 0.808) establishes a 25\% gap to
the best model, concentrated in text hallucination (THR: 0.05
vs.\ 0.244) and numerical accuracy (NF: 0.90 vs.\ 0.686)---
identifying text selection and diacritic handling as the primary
remaining fusion bottlenecks.

\paragraph{Broader implications for multimodal fusion.}
Our findings suggest two general principles beyond Vietnamese:
(1)~\textbf{Language-specific structural knowledge improves fusion
quality} when the task involves text in linguistically complex
languages---the phonological attention bias paradigm could be
extended to other tonal languages (Thai, Chinese, Yoruba) or
morphologically rich languages (Arabic, Turkish) with appropriate
structural features; and (2)~\textbf{fusion topology should be
data-driven}, as our ablation demonstrates that cross-modal graph
edges can hurt rather than help when modality-specific processing
is more effective.

\paragraph{Future work.}
The primary fusion bottlenecks---text selection (39.8\% missing
text), diacritic propagation (15.6\%), and repetition
(18.0\%)---suggest three directions: (1)~explicit text selection
modules that determine \textit{which} OCR tokens to fuse before
generation begins, (2)~diacritic-aware decoding that leverages
phonological constraints during beam search, and (3)~hybrid
architectures that combine the text grounding strength of
task-specific fusion models with the language fluency of LMMs.
We release ViTextCaps, all model code, and evaluation tools to
support future research on linguistically informed multimodal
fusion.


\bibliographystyle{elsarticle-num-names}

\bibliography{main}

\clearpage
\appendix
\section{Dataset Construction Details}
\label{app:app_dataset}

\subsection{Annotation Protocol}
\label{sec:annotation_protocol}

\paragraph{Annotator Recruitment}

The annotation process involved \textbf{30 undergraduate students}
recruited from Vietnam National University Ho Chi Minh City
(VNU-HCM) and affiliated institutions, all native Vietnamese
speakers. Annotators received dedicated training on the annotation
guidelines over approximately \textbf{one month} before the main
annotation phase, comprising guideline familiarization, worked
examples, and iterative feedback. Each annotator was allocated
\textbf{one week} to complete their assigned image set.
All annotators were compensated at standard local rates.

\paragraph{Annotation Tool}

\begin{figure}[ht]
  \centering
  \includegraphics[width=\textwidth]{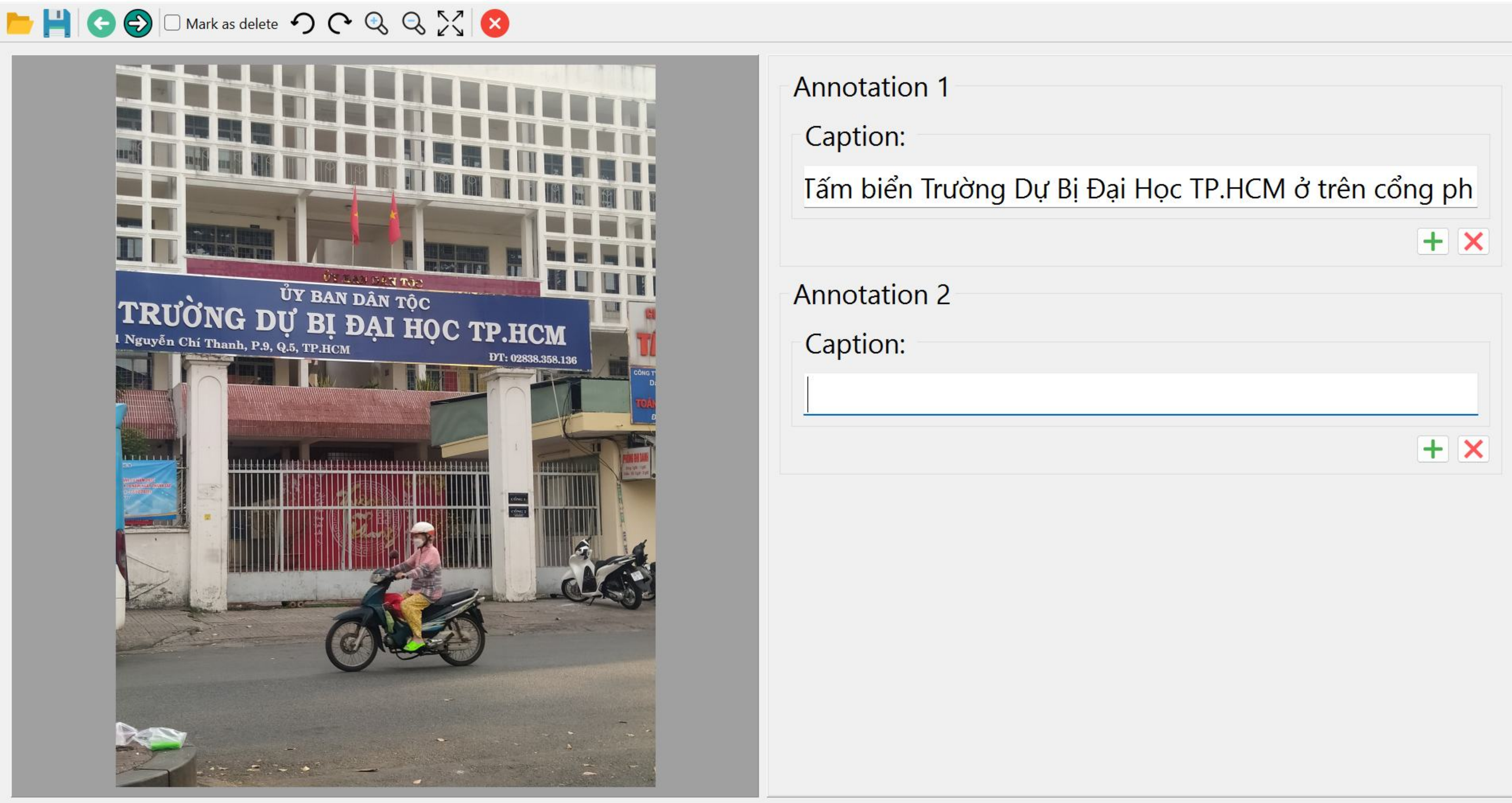}
  \caption{
    The dedicated desktop annotation tool used for 
    ViTextCaps caption collection. The interface 
    presents the source image at full resolution 
    (left panel) alongside independent caption fields 
    (right panel), each equipped with add and delete 
    controls.
  }
  \label{fig:annotation_tool}
\end{figure}

Annotations were collected via a dedicated desktop 
annotation tool (Figure~\ref{fig:annotation_tool}). 
To eliminate anchoring bias, annotators cannot view 
submissions from other annotators for the same image 
during their session. Captions are composed entirely 
from scratch without algorithmic pre-filling or 
OCR assistance.

\paragraph{Annotation Guidelines}

Annotators followed the task definition in
Section~\ref{sec:task_definition} and four operational
principles:

\begin{itemize}
  \item \textbf{(P1) Verbatim transcription.}
  Scene text is copied exactly as it appears in the image,
  preserving typos, missing diacritics, and abbreviations.
  For example, \textit{``ko''} is retained rather than
  corrected to \textit{``không''}, and \textit{``k/c''}
  is retained rather than expanded to
  \textit{``khoảng cách''}.

  \item \textbf{(P2) Omission over guessing.}
  Text that is partially occluded, blurred, or otherwise
  unreadable is omitted entirely rather than inferred,
  preventing hallucinated content from entering the
  reference captions.

  \item \textbf{(P3) Lexical restraint.}
  Color descriptions are restricted to basic terms
  (\textit{đỏ, cam, vàng, tím, trắng, đen}) to ensure
  cross-annotator consistency and avoid subjective
  color interpretation.

  \item \textbf{(P4) Caption variation.}
  The five captions per image must differ in sentence
  structure, vocabulary, or descriptive focus.
  For instance, one caption may lead with the visual
  context (\textit{``Một cửa hàng với biển hiệu
  ghi...''}) while another leads with the text content
  (\textit{``Dòng chữ ... xuất hiện trên...''}).
\end{itemize}


\subsection{Quality Control}
\label{sec:quality_control}

\paragraph{Stage 1 --- Pilot calibration.}
\label{sec:pilot}
Before the main phase, all annotators independently labeled a
shared \textbf{pilot set of 100 images} evaluated against
gold-standard annotations from the research team on three
criteria: scene text matches the image exactly; captions are
grammatically correct; captions demonstrate required variation.
Fewer than \textbf{40\% of annotators} required more than one
submission to meet the acceptance threshold, indicating broad
guideline clarity while identifying annotators needing additional
calibration. All 30 annotators reached acceptable agreement
before proceeding.

\paragraph{Stage 2 --- Batch review.}
Submissions were organized in \textbf{batches of 500 images}.
Five research team members---independent of the annotation
pool---randomly sampled \textbf{10\% of each batch} (50 images,
250 captions) for manual evaluation against grammatical accuracy,
semantic alignment, and scene-text transcription fidelity.
Any batch in which more than 5 of the 50 sampled images
contained errors was \textbf{rejected in its entirety} and
returned for re-annotation. Images flagged via the
\textit{Mark as Delete} mechanism were additionally reviewed
and removed if warranted. Batch rejection was most frequent
during the first two weeks, reflecting annotators' adaptation
to diacritic transcription and caption variation requirements;
rejection rates declined substantially thereafter.

\paragraph{Stage 3 --- Final review.}
Prior to dataset release, each annotator conducted a structured
self-review guided by a checklist covering diacritic accuracy,
verbatim transcription fidelity, and caption variation.
This stage complements the expert audit by addressing systematic
individual errors that batch-level sampling may not capture.

\clearpage
\section{Further Dataset Analysis}
\label{app:further_analysis}

\subsection{Domain Distribution}
\label{sec:domain_distribution}

We analyze the visual domains represented in ViTextCaps using
object categories detected by the VinVL object detector. Each
image is classified based on the distribution of detected objects
across nine domain categories, defined as follows:
\textit{Street/Outdoor} (storefronts, banners, and road signs
viewed from outside);
\textit{Indoor/Furniture} (indoor retail, office, and living
environments);
\textit{Fashion/Clothing} (apparel stores and clothing markets
with text on price tags, labels, and brand displays);
\textit{Food/Restaurant} (menus, food stalls, and packaged food);
\textit{Book/Publication} (book covers, newspapers, and printed
media);
\textit{Shop/Store} (general commercial signage and storefronts);
\textit{Product/Packaging} (product labels, ingredient lists,
and barcodes). Two cross-domain categories---\textit{Signage/Text}
and \textit{Person/Portrait}---appear across all scenes and are
excluded from primary ranking.

Table~\ref{tab:domain_distribution} summarizes the distribution.

\begin{table}[htbp]
\centering
\caption{Domain distribution of ViTextCaps images, classified
using detected objects from VinVL. \textit{Primary}: dominant
scene domain per image. \textit{Presence}: images where the
domain appears as a secondary element (multi-label).
Person/Portrait and Signage/Text are excluded from primary
ranking as they appear across all domains.}
\label{tab:domain_distribution}
\begin{tabular}{lrrrr}
\toprule
\textbf{Domain} & \textbf{Primary} & \textbf{\%}
  & \textbf{Presence} & \textbf{\%} \\
\midrule
Street/Outdoor    & 2{,}923 & 26.6\% & 6{,}518 & 59.3\% \\
Indoor/Furniture  & 2{,}679 & 24.4\% & 8{,}414 & 76.5\% \\
Fashion/Clothing  & 2{,}574 & 23.4\% & 6{,}204 & 56.4\% \\
Food/Restaurant   &   929   &  8.4\% & 3{,}849 & 35.0\% \\
Book/Publication  &   726   &  6.6\% & 2{,}037 & 18.5\% \\
Shop/Store        &   626   &  5.7\% & 5{,}124 & 46.6\% \\
Product/Packaging &   268   &  2.4\% & 3{,}809 & 34.6\% \\
\midrule
\textit{Signage/Text}    & --- & --- & 9{,}664 & 87.9\% \\
\textit{Person/Portrait} & --- & --- & 7{,}447 & 67.7\% \\
\bottomrule
\end{tabular}
\end{table}

Street/Outdoor (26.6\%), Indoor/Furniture (24.4\%), and
Fashion/Clothing (23.4\%) are the three most frequent primary
domains, collectively accounting for 74.4\% of images.
Food/Restaurant (8.4\%), Book/Publication (6.6\%), and Shop/Store
(5.7\%) form the second tier. Signage/Text objects are present
in 87.9\% of images and Person/Portrait in 67.7\%, confirming
that these are ubiquitous elements across domains rather than
domain-defining categories.

\subsection{Multi-Domain Characteristics}

The average image is associated with 4.8 domain categories, and
most images span multiple visual domains. The most frequent
co-occurrences are Indoor/Furniture + Street/Outdoor (5{,}452
images) and Fashion/Clothing + Indoor/Furniture (5{,}183 images).

This multi-domain property has a practical consequence for model
design: images spanning more domains contain text from more
diverse sources at lower average OCR confidence---images with
4+ associated domains have mean OCR confidence 0.63 and mean
text density 44.3 tokens/image, compared to 0.68 and 15.5 for
single-domain images (Section~\ref{sec:domain_eval}). This
gradient directly motivates the confidence gate in HSTFG
(Section~\ref{sec:hstfg_graph}), which down-weights attention
toward low-confidence detections in visually complex,
multi-source scenes. In Section~\ref{sec:domain_eval}, we
examine how model performance varies across domains and domain
complexity levels.


\subsection{OCR--Caption Interaction}
\label{sec:ocr_caption_alignment}

A defining feature of scene-text captioning is the \textit{copy
mechanism}: models can either generate tokens from the language
model vocabulary or copy them directly from the OCR output.
Understanding the balance between copying and generation in the
ground-truth captions is essential for model design.

\paragraph{Copy vs.\ Generate Ratio}

We classify each of the 1{,}150{,}352 caption tokens in the
training set as either \textit{copied} (matching an OCR token
for the same image) or \textit{generated}. A token is considered
an exact copy if it matches an OCR token after normalization,
or a base-form copy if it matches after stripping diacritics
(accounting for OCR diacritic errors identified in
Section~\ref{sec:ocr_error_taxonomy}).
Table~\ref{tab:copy_generate} reports the source classification.

\begin{table}[ht]
\centering
\caption{Source classification of training-set caption tokens.
Despite the annotation requirement to reference scene text,
71.1\% of caption tokens are generated rather than copied from
OCR, reflecting the descriptive nature of Vietnamese captioning.}
\label{tab:copy_generate}
\begin{tabular}{lrr}
\toprule
\textbf{Source} & \textbf{Tokens} & \textbf{\%} \\
\midrule
Exact copy            & 276{,}945 & 24.1\% \\
Base-form copy        &  56{,}074 &  4.9\% \\
\textbf{Total copied} & \textbf{333{,}019} & \textbf{28.9\%} \\
Generated             & 817{,}333 & 71.1\% \\
\midrule
Total                 & 1{,}150{,}352 & 100\% \\
\bottomrule
\end{tabular}
\end{table}

\textbf{71.1\% of caption tokens are generated}, not copied from
OCR. This ratio is notable given the annotation guideline
requiring OCR reference: annotators embed scene text within
descriptive Vietnamese prose, producing captions where the
majority of tokens are structural (``dòng chữ màu đỏ ghi\ldots'',
\textit{red text that reads\ldots}) rather than verbatim copies.
Unlike extractive QA tasks where answers are largely copied,
Vietnamese scene-text captioning requires a capable language
generator---the copy mechanism supplements but does not replace
generation.

The 4.9\% base-form copy rate---tokens matching OCR after
diacritic stripping---provides an independent estimate of
diacritic divergence consistent with the 20.6\% divergence
rate from Section~\ref{sec:diacritic_divergence}, quantifying
the burden of diacritic correction on the copy mechanism.

The mean per-caption copy ratio is 29.3\% (median 25.7\%,
$\sigma = 0.22$), with substantial variance: some captions
directly quote long sign text ($>$60\% copied), while others
describe the scene with minimal verbatim copying ($<$10\%).

\paragraph{Copy Rate by Token Type}

Table~\ref{tab:copy_rate_type} breaks down copy rates by token
language category.

\begin{table}[ht]
\centering
\caption{Copy rate by caption token type. English and numeric
tokens are copied more frequently because they resist natural
paraphrasing into Vietnamese.}
\label{tab:copy_rate_type}
\begin{tabular}{lrrr}
\toprule
\textbf{Token Type} & \textbf{Total} & \textbf{Copied}
  & \textbf{Rate} \\
\midrule
Vietnamese          & 967{,}314 & 257{,}529 & 26.6\% \\
Numeric             &  24{,}301 &   9{,}327 & 38.4\% \\
English / Ambiguous & 158{,}737 &  66{,}163 & 41.7\% \\
\bottomrule
\end{tabular}
\end{table}

English and ambiguous tokens have the highest copy rate (41.7\%),
followed by numeric tokens (38.4\%), with Vietnamese tokens
lowest (26.6\%). This ordering reflects the asymmetry of
paraphrasing difficulty: Vietnamese text can be naturally
rephrased within Vietnamese prose, while English brand names
(``Samsung'', ``coffee shop'') and numbers (``50.000\dj'')
resist paraphrasing and are typically copied verbatim. The
15-point gap between Vietnamese and English copy rates suggests
that effective models should modulate copying behavior based on
token type or language identity---a design implication
consistent with the code-mixing analysis in
Section~\ref{sec:code_mixing}.


\subsection{Vocabulary \& Lexical Distribution}
\label{sec:vocabulary}

We analyze the lexical characteristics of ViTextCaps captions to
understand the vocabulary demands that models must handle.

\subsubsection{Corpus Statistics}

Table~\ref{tab:vocab_stats} reports the vocabulary and corpus
statistics computed across all splits.

\begin{table}[ht]
\centering
\caption{Vocabulary and corpus statistics of ViTextCaps
(all splits combined). The high OOV rate and large proportion
of rare words reflect the open-vocabulary nature of
scene-text captioning.}
\label{tab:vocab_stats}
\begin{tabular}{lr}
\toprule
\textbf{Statistic} & \textbf{Value} \\
\midrule
Total captions                      & 74{,}970 \\
Vocabulary size                     & 30{,}982 \\
Avg caption length                  & 22.0 $\pm$ 13.1 tokens \\
Median caption length               & 19.0 tokens \\
Type-Token Ratio (TTR)              & 0.019 \\
\midrule
Hapax legomena (freq\,=\,1)         & 9{,}983 (32.2\%) \\
Rare words (freq\,$\leq$\,5)        & 18{,}811 (60.7\%) \\
Words containing digits             & 5{,}295 (17.1\%) \\
OOV (test $\setminus$ train)        & 4{,}064 (35.7\%) \\
\bottomrule
\end{tabular}
\end{table}

Table~\ref{tab:vocab_stats} reveals several characteristics that
distinguish ViTextCaps from standard image captioning datasets:

\paragraph{Large, sparse vocabulary.}
The vocabulary contains 30{,}982 unique word types---substantially
larger than typical captioning datasets of similar size (e.g.,
MSCOCO's $\sim$10K vocabulary for 600K captions~\cite{lin2014coco}).
This inflation is driven by scene-text tokens: store names, brand
names, addresses, and abbreviations that appear in OCR output.
The extreme sparsity is evident from the 60.7\% of words appearing
five times or fewer and 32.2\% hapax legomena (single occurrences).

\paragraph{High OOV rate.}
35.7\% of test-set vocabulary items do not appear in the training
set. This is a direct consequence of the open-vocabulary nature of
scene text: each image may contain novel proper nouns, phone numbers,
or domain-specific terms that no training set can fully cover. This
OOV rate motivates the use of copy mechanisms that can directly
reproduce OCR tokens rather than relying solely on the learned
vocabulary.

\paragraph{Digit-bearing words are prevalent.}
17.1\% of the vocabulary contains digits (phone numbers, dates,
prices, addresses), reflecting the quantitative richness of
Vietnamese scene text. These tokens are particularly challenging for
generation models because they must be reproduced exactly---an
approximate phone number is worse than no phone number at all.

\subsubsection{Lexical Distribution}

Table~\ref{tab:top_words} lists the 15 most frequent words in
ViTextCaps captions, revealing the dual vocabulary structure
of scene-text annotations.

\begin{table}[ht]
\centering
\caption{Top 15 most frequent words in ViTextCaps captions.
Scene-text-specific terms (\textit{chữ}, \textit{dòng},
\textit{tấm}, \textit{bảng}) are highlighted in bold.}
\label{tab:top_words}
\begin{tabular}{rlrr}
\toprule
\textbf{Rank} & \textbf{Word} & \textbf{Count} & \textbf{\%} \\
\midrule
1  & màu (color)             & 35{,}569 & 2.16 \\
2  & có (have/exist)         & 33{,}135 & 2.01 \\
3  & \textbf{chữ} (text)     & 26{,}827 & 1.63 \\
4  & một (one/a)             & 24{,}555 & 1.49 \\
5  & \textbf{dòng} (line)    & 20{,}715 & 1.26 \\
6  & trên (on/above)         & 19{,}713 & 1.20 \\
7  & được (be/get)           & 17{,}053 & 1.03 \\
8  & là (is)                 & 16{,}162 & 0.98 \\
9  & và (and)                & 13{,}663 & 0.83 \\
10 & của (of)                & 13{,}196 & 0.80 \\
11 & bên (side)              & 12{,}992 & 0.79 \\
12 & hàng (row/store)        & 12{,}883 & 0.78 \\
13 & \textbf{tấm} (sheet)    & 12{,}289 & 0.75 \\
14 & ở (at)                  & 11{,}401 & 0.69 \\
15 & \textbf{bảng} (board)   & 10{,}881 & 0.66 \\
\bottomrule
\end{tabular}
\end{table}

The most frequent words form two distinct groups:
\begin{itemize}[nosep]
    \item \textbf{Scene-text vocabulary}: \textit{chữ} (``text'',
    rank 3), \textit{dòng} (``line of text'', rank 5), \textit{tấm}
    (``sheet/board'', rank 13), and \textit{bảng} (``signboard'',
    rank 15) are domain-specific terms that appear far more
    frequently than in general Vietnamese text.
    \item \textbf{Descriptive vocabulary}: \textit{màu} (``color'',
    rank 1), \textit{có} (``have'', rank 2), \textit{trên}
    (``on'', rank 6) are common Vietnamese function and descriptive
    words used to situate text within the visual scene.
\end{itemize}

The co-occurrence of these two groups reflects the dual nature of
scene-text captions: annotators must both \textit{describe} the
visual context and \textit{transcribe} the textual content.

\subsubsection{Caption Length Distribution}

Table~\ref{tab:length_dist} reports the distribution of caption
lengths across all splits.

\begin{table}[ht]
\centering
\caption{Caption length distribution. The majority of captions
(66.3\%) fall within 11--25 tokens, while a long tail ($>$40
tokens) captures complex scenes with extensive text.}
\label{tab:length_dist}
\begin{tabular}{lrr}
\toprule
\textbf{Length (tokens)} & \textbf{Count} & \textbf{\%} \\
\midrule
1--10  &  6{,}941 &  9.3 \\
11--15 & 15{,}321 & 20.4 \\
16--20 & 20{,}354 & 27.1 \\
21--25 & 14{,}060 & 18.8 \\
26--30 &  6{,}896 &  9.2 \\
31--40 &  6{,}189 &  8.3 \\
41--50 &  2{,}725 &  3.6 \\
51+    &  2{,}483 &  3.3 \\
\bottomrule
\end{tabular}
\end{table}

Caption length follows a right-skewed distribution centered around
16--20 tokens. The long tail (6.9\% captions with $>$40 tokens)
corresponds to text-heavy images where annotators transcribe
extensive scene text content. This length variability poses a
challenge for autoregressive decoders, which must learn when to
stop generating without truncating important text content or
entering repetition loops (Section~\ref{sec:error_analysis}).


\subsection{Text Density and Layout}
\label{sec:text_density}

We characterize the amount and spatial arrangement of text in
ViTextCaps images.

\subsubsection{Text Density}
\label{sec:text_density_1}
Table~\ref{tab:text_density} reports the OCR token density
distribution across all three splits.

\begin{table}[ht]
\centering
\caption{Text density distribution across splits, measured by
the number of OCR tokens per image. The distribution is
consistent across train, dev, and test.}
\label{tab:text_density}
\begin{tabular}{lrrr}
\toprule
\textbf{Statistic} & \textbf{Train} & \textbf{Dev} & \textbf{Test} \\
\midrule
Mean tokens/image   & 31.8  & 31.0  & 31.3  \\
Median tokens/image & 23    & 23    & 23    \\
Std                 & 27.7  & 26.3  & 26.8  \\
\midrule
Sparse ($\leq$3)    & 5.8\%  & 4.9\%  & 5.3\%  \\
Medium (4--10)      & 19.0\% & 19.1\% & 19.7\% \\
Dense ($>$10)       & 75.1\% & 76.0\% & 75.0\% \\
\bottomrule
\end{tabular}
\end{table}

75\% of images contain more than 10 OCR tokens, with a mean of
31.8 tokens per image. The distribution is right-skewed (median
23, mean 31.8), indicating a long tail of very text-heavy images.
The density distribution is consistent across splits, confirming
that the train/dev/test partition does not introduce density bias.

\subsubsection{Layout Complexity}

We classify text layout based on the spatial arrangement of OCR
bounding boxes. Table~\ref{tab:layout_complexity} reports the
results on the training set.

\begin{table}[t]
\centering
\caption{Text layout complexity distribution (training set).}
\label{tab:layout_complexity}
\begin{tabular}{lr}
\toprule
\textbf{Layout Type} & \textbf{\%} \\
\midrule
Multi-block (multiple spatial regions)     & 74.3\% \\
Multi-line (multiple lines, single region) & 18.9\% \\
Single-line                                &  6.8\% \\
\bottomrule
\end{tabular}
\end{table}

74.3\% of images have multi-block layouts, where text appears in
multiple spatially separated regions (e.g., multiple signs, labels,
or text blocks). This spatial heterogeneity directly contributes
to the text selection challenge analyzed in
Section~\ref{sec:text_dependency}---models must choose among
multiple text sources when generating captions---and motivates
the spatial attention bias in HSTFG
(Section~\ref{sec:hstfg_graph}) as a necessary inductive prior
for distinguishing geometrically distant token groups.

\subsubsection{OCR Confidence Distribution}

The mean OCR confidence is 0.638 ($\sigma = 0.231$), with 73.0\%
of images in the medium confidence range (0.33--0.67), 15.1\% in
high ($>$0.67), and 11.9\% in low ($<$0.33). This confidence
distribution provides the basis for the stratified evaluation in
\ref{app:stratified}, where we show that model
performance varies 2.9$\times$ between low and high confidence
subsets.


\subsection{OCR Error Taxonomy}
\label{sec:ocr_error_taxonomy}

Section~\ref{sec:tone_diacritic_ambiguity} established that diacritic
divergence between OCR and human reading occurs in 20.6\% of matched
token pairs. We now ask: \textit{how} do these divergences manifest?
Understanding the structure of OCR errors is essential both for
designing robust captioning models and for interpreting benchmark
results.

We propose a fine-grained taxonomy of five error types specific to
Vietnamese orthography, derived from character-level alignment of
89{,}730 caption--OCR divergence pairs:

\begin{itemize}[nosep]
    \item \textbf{T1 --- Tone Drop.} A toned vowel is recognized
    as its unmarked base form: \textit{hóa} (chemistry) $\to$
    \textit{hoa} (flower). The diacritical mark is lost entirely.
    \item \textbf{T2 --- Tone Substitution.} One tone mark is
    replaced by another: \textit{tài} (talent) $\to$ \textit{tải}
    (to download). Both forms carry a diacritic, but the wrong one.
    \item \textbf{T3 --- Vowel Variant Confusion.} The base vowel
    character is swapped within its family
    (\textit{a}$\leftrightarrow$\textit{â}$\leftrightarrow$\textit{ă},
    \textit{o}$\leftrightarrow$\textit{ô}$\leftrightarrow$\textit{ơ},
    \textit{u}$\leftrightarrow$\textit{ư}): \textit{nghiệm} $\to$
    \textit{nghiẹm}, where the circumflex on \textit{ê} is lost,
    mapping \textit{ệ} to \textit{ẹ}.
    \item \textbf{T4 --- Đ/D Confusion.} The stroke on \textit{đ}
    is added or removed: \textit{dẫn} (to lead) $\to$ \textit{đẫn}.
    This is unique to Vietnamese among Latin-script languages.
    \item \textbf{T5 --- Tone Insertion.} A diacritical mark is
    hallucinated on an unmarked vowel: \textit{hoa} $\to$
    \textit{hóa}. The inverse of T1.
\end{itemize}

When multiple error types co-occur in a single word (e.g.,
\textit{nguyễn} $\to$ \textit{nguyên}, involving both a vowel
variant shift and a tone change), the instance is additionally
labeled as \textit{compound}.

\subsubsection{Error Distribution}
\label{sec:error_distribution}

Table~\ref{tab:ocr_error_taxonomy} reports the distribution across
the five error types, derived from 141{,}526 error instances across
the 89{,}730 divergence pairs identified in
Section~\ref{sec:diacritic_divergence}.

\begin{table}[ht]
\centering
\caption{Vietnamese OCR error taxonomy and distribution.
A single word pair may contribute to multiple error type counts
when compound errors are present.}
\label{tab:ocr_error_taxonomy}
\begin{tabular}{clrrl}
\toprule
\textbf{ID} & \textbf{Error Type} & \textbf{Count}
  & \textbf{\%} & \textbf{Example} \\
\midrule
T3 & Vowel Variant     & 34{,}384 & 24.3\%
  & nghiệm $\to$ nghiẹm \\
T1 & Tone Drop         & 31{,}878 & 22.5\%
  & hóa $\to$ hoa \\
T2 & Tone Substitution & 31{,}396 & 22.2\%
  & tài $\to$ tải \\
-- & Compound          & 24{,}520 & 17.3\%
  & nguyễn $\to$ nguyên \\
T5 & Tone Insertion    & 11{,}201 &  7.9\%
  & hoa $\to$ hóa \\
T4 & Đ/D Confusion     &  8{,}147 &  5.8\%
  & dẫn $\to$ đẫn \\
\midrule
   & \textbf{Total}    & \textbf{141{,}526} & \textbf{100\%} & \\
\bottomrule
\end{tabular}
\end{table}

The three dominant types---T3 (24.3\%), T1 (22.5\%), and T2
(22.2\%)---account for nearly 70\% of all errors, each contributing
roughly equally. This balance indicates that Vietnamese OCR errors
are not dominated by a single failure mode; rather, the diacritical
system is vulnerable along multiple orthographic dimensions
simultaneously. Compound errors (17.3\%) confirm that real-world
OCR errors are frequently more complex than isolated single-character
mistakes.

\subsubsection{Error Patterns by OCR Confidence}
\label{sec:error_by_confidence}

Table~\ref{tab:error_by_confidence} breaks down error type
proportions across three OCR confidence buckets.

\begin{table}[ht]
\centering
\caption{Error type proportions stratified by OCR confidence.
Values indicate the percentage of each error type within each
confidence bucket.}
\label{tab:error_by_confidence}
\begin{tabular}{lccc}
\toprule
\textbf{Error Type} & \textbf{Low} ($<$0.5) & \textbf{Med.} (0.5--0.8) & \textbf{High} ($\geq$0.8) \\
\midrule
T1 Tone Drop        & \textbf{28.9\%} & 23.5\% & 18.1\% \\
T2 Tone Sub.        & 14.9\% & 22.2\% & \textbf{26.4\%} \\
T3 Vowel Variant    & 25.5\% & 24.3\% & 23.6\% \\
T4 Đ/D Confusion   & 6.3\%  & 5.3\%  & 5.8\%  \\
T5 Tone Insert      & 5.3\%  & 7.5\%  & 9.7\%  \\
Compound            & 19.1\% & 17.2\% & 16.4\% \\
\bottomrule
\end{tabular}
\end{table}

A striking shift emerges across confidence levels. At low
confidence, tone drop (T1) is the dominant error at
28.9\%---the OCR system fails to detect the diacritical mark and
defaults to the unmarked base form. At high confidence, tone drop
decreases to 18.1\% while tone substitution (T2) rises to 26.4\%.
This pattern suggests that when the OCR system is uncertain, it
tends to \textit{omit} diacritics (a conservative failure); when
it is confident, it more often selects the \textit{wrong} diacritic
(an assertive failure). The latter case is arguably more dangerous
for downstream captioning, as a confidently incorrect tone mark is
more likely to be propagated through the copy mechanism without
correction.

Vowel variant confusion (T3) remains stable across all confidence
levels at 23--26\%, indicating that base vowel identity errors
(\textit{o}$\leftrightarrow$\textit{ô},
\textit{a}$\leftrightarrow$\textit{â}) are a systematic challenge
independent of recognition certainty.

\subsubsection{Tone Confusion Structure}
\label{sec:tone_confusion}

To understand which tones are most frequently confused, we construct
a tone confusion matrix from all T1, T2, and T5 error instances.
The most frequent confusion is
\textit{sắc}~$\to$~\textit{ngang} (10{,}593 instances), followed
by \textit{huyền}~$\to$~\textit{ngang} (9{,}221). Together, these
two tone-to-ngang confusions account for the majority of T1 errors,
confirming that the acute accent (\textit{sắc}) and grave accent
(\textit{huyền})---the two most visually subtle marks---are the most
vulnerable to OCR misrecognition.

The asymmetry between tones is also informative. The confusion
\textit{hỏi}~$\to$~\textit{sắc} (4{,}151 instances) is 36\% more
frequent than its reverse \textit{sắc}~$\to$~\textit{hỏi} (3{,}045),
reflecting the visual similarity between the hook (\textit{hỏi})
and acute (\textit{sắc}) diacritics, which differ by only a small
curvature at the mark's tip.

\subsubsection{Vowel Variant Confusion}
\label{sec:vowel_confusion}

Among vowel variant errors, the \textit{o}$\leftrightarrow$\textit{ô}
pair dominates with 11{,}537 instances (8{,}833 in the
\textit{o}~$\to$~\textit{ô} direction and 2{,}704 in reverse),
followed by \textit{ă}$\leftrightarrow$\textit{a} (3{,}179) and
\textit{ư}$\leftrightarrow$\textit{u} (2{,}984 + 2{,}093). These
confusions arise because the distinguishing marks---the circumflex
on \textit{ô}, the breve on \textit{ă}, and the horn on
\textit{ư}---are small diacritical additions easily lost or
hallucinated at typical scene-text resolutions.

Notably, the Đ/D confusion (T4) is strongly directional: 79\% of
cases involve \textit{d}~$\to$~\textit{đ} (the OCR system adds
the stroke), while only 21\% involve \textit{đ}~$\to$~\textit{d}
(stroke removal), suggesting a learned OCR bias toward the stroked
variant, possibly because \textit{đ} is more frequent in Vietnamese
text than the unstroked \textit{d}.

The structured nature of these errors has three direct consequences
for model design. First, T1 and T3 errors---where the correct
diacritic can often be inferred from lexical context---could
potentially be addressed through strong Vietnamese language model
priors; however, T2 errors at high confidence represent cases where
visual evidence actively misleads the system, requiring the model
to weigh linguistic plausibility against OCR confidence rather than
trusting it naively. Second, this confidence--error type interaction
argues against simple confidence thresholding, supporting the
graph-based fusion design in HSTFG
(Section~\ref{sec:hstfg_graph}) which considers spatial and
phonological relationships between tokens rather than relying
solely on confidence scores. Third, the systematic orthographic
patterns underlying all five error types confirm that standard
captioning metrics treating diacritic mismatches as generic token
errors will underestimate their semantic impact, further motivating
the THR and NF metrics proposed in Section~\ref{sec:main_results}.


\subsection{Code-Mixing Analysis}
\label{sec:code_mixing}

Vietnamese urban signage frequently mixes Vietnamese and English text,
reflecting the bilingual nature of commercial and public spaces in
Vietnam. This code-mixing introduces a challenge absent from English
scene-text datasets: a captioning model must simultaneously process
two languages with different morphological properties and decide how
to integrate foreign-language tokens into Vietnamese captions.

We analyze the extent and structure of code-mixing in ViTextCaps at
three levels: OCR tokens, images, and captions.

\subsubsection{Token-Level Language Distribution}
\label{sec:token_language}

We classify each of the 349{,}998 OCR tokens in the training set into
five language categories based on orthographic features:
\textit{Vietnamese} (contains Vietnamese-specific diacritics),
\textit{English} (recognized English words or brands),
\textit{Numeric} (numbers, prices, codes), \textit{Ambiguous}
(short ASCII words without diacritics that could be either Vietnamese
or English), and \textit{Other/Mixed}.
Table~\ref{tab:token_language_dist} reports the distribution.

\begin{table}[ht]
\centering
\caption{Language distribution of OCR tokens in ViTextCaps
(training split). \textit{Ambiguous} tokens are short words
without diacritics whose language cannot be determined from
orthography alone.}
\label{tab:token_language_dist}
\begin{tabular}{lrr}
\toprule
\textbf{Category} & \textbf{Tokens} & \textbf{\%} \\
\midrule
Vietnamese  & 190{,}888 & 54.5\% \\
Ambiguous   &  97{,}238 & 27.8\% \\
Numeric     &  39{,}910 & 11.4\% \\
English     &  10{,}858 &  3.1\% \\
Other/Mixed &  11{,}104 &  3.2\% \\
\midrule
\textbf{Total} & \textbf{349{,}998} & \textbf{100\%} \\
\bottomrule
\end{tabular}
\end{table}

Two notable findings emerge. First, over a quarter of all OCR tokens
(27.8\%) are \textit{ambiguous}: short words without diacritics whose
language cannot be resolved from orthography alone. Vietnamese has many
common monosyllabic words written without diacritics
(e.g., \textit{con}, \textit{ban}, \textit{can}) that are
orthographically indistinguishable from English words. This ambiguity
layer, unique to Vietnamese among scene-text datasets, means that even
language identification---a trivial task in English-only
settings---becomes a non-trivial challenge.

Second, English tokens constitute 3.1\% of all OCR output. While this
proportion appears modest, its distribution across images tells a
different story.

\subsubsection{Image-Level Mixing Profiles}
\label{sec:image_profiles}

Table~\ref{tab:image_mixing_profiles} classifies images by their
language mixing profile based on the proportion of Vietnamese, English,
and numeric tokens in the OCR output.

\begin{table}[ht]
\centering
\caption{Language mixing profiles of ViTextCaps images. An image
is classified based on the proportion of Vietnamese, English,
and numeric tokens in its OCR output (with ambiguous tokens
split equally between VN and EN).}
\label{tab:image_mixing_profiles}
\begin{tabular}{lrr}
\toprule
\textbf{Profile} & \textbf{Images} & \textbf{\%} \\
\midrule
Vietnamese only       &   469   &  4.3\% \\
VN dominant (with EN) & 7{,}672 & 69.8\% \\
Balanced (VN + EN)    & 2{,}583 & 23.5\% \\
English dominant      &   170   &  1.5\% \\
Numeric dominant      &    95   &  0.9\% \\
\midrule
\textit{Any EN present} & \textit{4{,}722} & \textit{42.9\%} \\
\textit{Both VN and EN} & \textit{4{,}447} & \textit{40.4\%} \\
\bottomrule
\end{tabular}
\end{table}

\textbf{42.9\% of images contain at least one English token}, and
\textbf{40.4\% contain both Vietnamese and English text on the same
image}. The dominant profile (69.8\%) is \textit{VN dominant with
English}---Vietnamese signs that incorporate English brand names,
loanwords, or decorative text. Only 4.3\% of images contain
exclusively Vietnamese text.

This pervasive bilingualism means that a model trained on ViTextCaps
must handle code-mixed input as the norm, not the exception. English
datasets such as TextCaps do not exhibit this property, as English
signage in Western countries rarely contains substantial text in other
languages.



The most frequent English tokens in the OCR output reflect the
commercial nature of Vietnamese street scenes: \textit{cafe} (252),
\textit{sale} (195), \textit{shop} (168), \textit{coffee} (164),
\textit{hotel} (75), and brand names such as \textit{Samsung} (58).
Among English subtypes, long words (53.6\%) and generic English
vocabulary (41.3\%) dominate, with recognized brand names comprising
5.1\%.

\subsubsection{Caption-Level Code-Switching}
\label{sec:caption_code_switching}

Annotators writing Vietnamese captions must decide how to handle
English text visible in the image. Our analysis of 52{,}359 training
captions reveals:

\begin{itemize}[nosep]
    \item 84.0\% of caption tokens are Vietnamese, 1.7\% are English,
    and the remainder are numeric or ambiguous.
    \item \textbf{23.0\% of captions contain at least one English
    token}, confirming that annotators naturally code-switch when
    describing bilingual scenes.
\end{itemize}

This code-switching rate is substantially higher than what is observed
in English captioning datasets, where captions are almost exclusively
monolingual. The presence of English tokens in Vietnamese captions
creates challenges for evaluation: metrics must handle cross-lingual
token matching, and a model that paraphrases ``coffee shop'' as
``quán cà phê'' should not be penalized more heavily than one that
copies the English text verbatim.

\subsubsection{Copy Rate by Language}
\label{sec:copy_rate}

Table~\ref{tab:copy_rate_language} reports how copy rates differ by
language category, revealing systematic differences in how annotators
handle Vietnamese versus foreign-language OCR tokens.

\begin{table}[ht]
\centering
\caption{Copy rate of OCR tokens into captions by language category.
A token is considered copied if it or its base-form-stripped variant
appears in at least one caption for the same image.}
\label{tab:copy_rate_language}
\begin{tabular}{lrrr}
\toprule
\textbf{Category} & \textbf{OCR Tokens} & \textbf{Copied}
  & \textbf{Rate} \\
\midrule
Vietnamese  & 190{,}888 & 97{,}153 & 50.9\% \\
English     &  10{,}858 &  3{,}366 & 31.0\% \\
Ambiguous   &  97{,}238 & 28{,}560 & 29.4\% \\
Numeric     &  39{,}910 &  5{,}684 & 14.2\% \\
\bottomrule
\end{tabular}
\end{table}

Vietnamese tokens appear verbatim in captions at a rate of 50.9\%,
while English tokens appear at only 31.0\%---a gap of nearly 20
percentage points. This indicates that when annotators reference
English text, they tend to \textit{paraphrase it into Vietnamese}
for natural flow (e.g., writing ``quán cà phê'' instead of copying
``coffee shop''), while Vietnamese text is more often quoted directly.
Numeric tokens have the lowest copy rate (14.2\%), indicating that
annotators frequently summarize numerical information (prices, phone
numbers, addresses) rather than transcribing it verbatim.

These findings have three direct consequences for model design.
With 42.9\% of images containing English text and 23.0\% of captions
code-switching, monolingual approaches to Vietnamese scene-text
captioning face a systematic blind spot---models must process
cross-lingual input as the norm, a challenge that English-only
benchmarks such as TextCaps do not capture. The 27.8\% ambiguous
token layer interacts directly with the diacritic vulnerability
identified in Section~\ref{sec:tone_diacritic_ambiguity}: a
Vietnamese word that has lost its diacritics through OCR error
(T1, T5) becomes orthographically identical to an English word,
compounding the identification difficulty. This motivates the gating
design of the phonological attention bias in PhonoSTFG
(Section~\ref{sec:phonological_bias}): features $p_1$--$p_7$ are
activated only when both tokens are confirmed Vietnamese syllables
($p_8 = 1$), preventing phonological reasoning from being applied
to ambiguous or non-Vietnamese tokens where it is meaningless.
Finally, the 20-point gap between Vietnamese and English copy rates
suggests that effective models should learn language-dependent
copying behavior---favoring direct copying for Vietnamese text
and paraphrasing for English---rather than applying a uniform
copy probability across all token types.


\subsection{Linguistic Structure Analysis}
\label{sec:linguistic_analysis}

Beyond lexical-level phenomena
(Sections~\ref{sec:tone_diacritic_ambiguity}--\ref{sec:text_dependency}),
we analyze the syntactic structure of ViTextCaps captions to
characterize the linguistic complexity that models must handle.
We apply POS tagging to all 74{,}970 reference captions and
dependency parsing to a random sample of 2{,}000 captions using
Underthesea~\cite{underthesea}.

\subsubsection{Part-of-Speech Distribution}

\begin{figure*}[t]
    \centering
    \includegraphics[width=\textwidth]{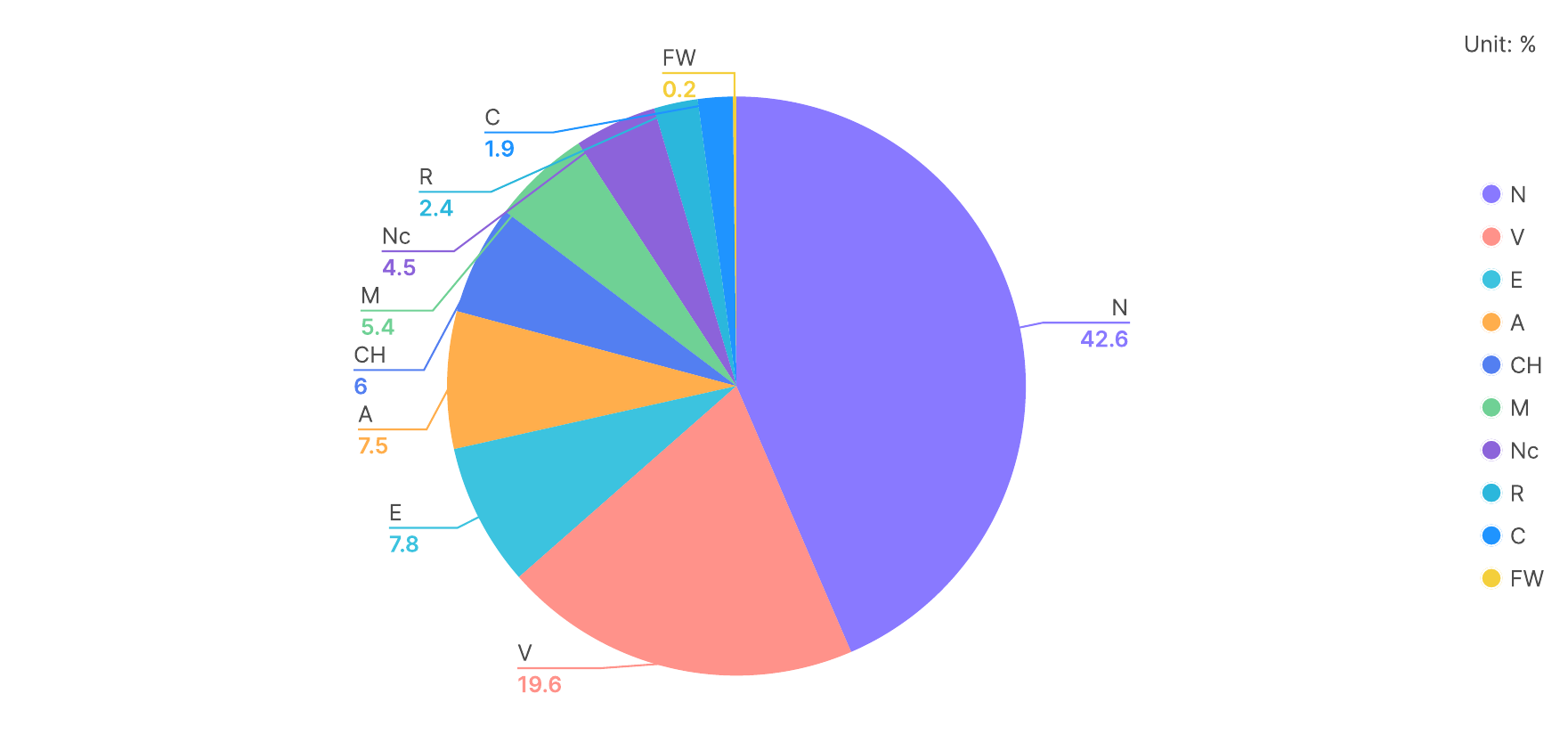}
    \caption{POS tag distribution of ViTextCaps reference captions
    (74{,}970 captions). Nouns dominate at 42.6\%---substantially
    above typical image captioning datasets ($\sim$30\% in
    MSCOCO)---driven by OCR-extracted proper nouns, store names,
    and brand names.}
    \label{fig:pos_distribution}
\end{figure*}

Figure~\ref{fig:pos_distribution} shows the POS tag distribution
across all 74{,}970 reference captions.

Nouns dominate at 42.6\%, substantially higher than typical image
captioning datasets ($\sim$30\% in MSCOCO~\cite{lin2014coco}). This
reflects the nature of scene-text captioning: captions frequently
reference store names, brand names, addresses, and other proper nouns
extracted from OCR. The high numeral proportion (M, 5.4\%) further
confirms the prevalence of phone numbers, dates, and prices in
Vietnamese scene text. Foreign words (FW, 0.2\%) correspond to the
code-mixing phenomenon analyzed in Section~\ref{sec:code_mixing},
appearing as English brand names and loanwords.

\subsubsection{Vietnamese Classifier Usage}
\label{sec:classifier_analysis}

Vietnamese employs a rich system of classifier nouns (loại từ) that
must precede common nouns in many syntactic contexts. Correct
classifier usage is essential for natural Vietnamese text generation.
Table~\ref{tab:classifier_distribution} reports the ten most frequent
classifiers in ViTextCaps captions.

\begin{figure}[htbp]
    \centering
    \includegraphics[width=\textwidth]{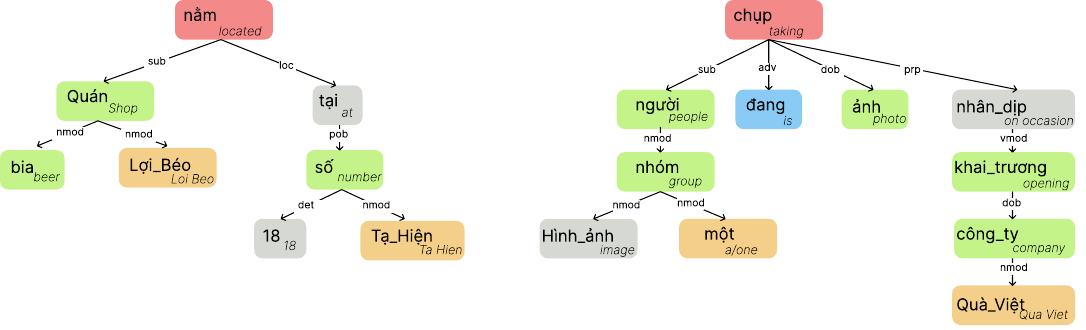}
    \caption{Syntactic dependency trees for two captions. Left: A simple sentence structure (\textit{"Quán bia Lợi Béo nằm tại số 18 Tạ Hiện"}). Right: A structurally complex sentence (\textit{"Hình ảnh một nhóm người đang chụp ảnh nhân dịp khai trương công ty Quà Việt"}) demonstrating deeper modifier nesting (\textit{nmod, vmod}) to capture specific reading comprehension elements like the event and company name.}
    \label{fig:dependency_tree}
\end{figure}

\begin{table}[ht]
\centering
\caption{Top 10 Vietnamese classifiers in ViTextCaps captions.
Scene-text-specific classifiers (\textit{dòng}, \textit{tấm})
dominate, reflecting the domain focus on text-bearing objects.}
\label{tab:classifier_distribution}
\begin{tabular}{llrp{3.8cm}}
\toprule
\textbf{Classifier} & \textbf{Meaning} & \textbf{Count}
  & \textbf{Typical usage} \\
\midrule
dòng  & line/flow & 29{,}707 & \textit{dòng chữ} (line of text) \\
tấm   & flat obj. & 17{,}418 & \textit{tấm bảng} (signboard) \\
bức   & picture   & 10{,}868 & \textit{bức ảnh} (photograph) \\
chiếc & general   &  6{,}245 & \textit{chiếc xe} (vehicle) \\
cuốn  & book      &  5{,}368 & \textit{cuốn sách} (book) \\
cái   & general   &  3{,}494 & \textit{cái bàn} (table) \\
con   & animal    &  3{,}040 & \textit{con đường} (road) \\
quyển & book      &  2{,}820 & \textit{quyển sách} (book) \\
ngôi  & building  &  1{,}402 & \textit{ngôi nhà} (house) \\
hộp   & box       &  1{,}016 & \textit{hộp sữa} (milk box) \\
\bottomrule
\end{tabular}
\end{table}

\textit{Dòng} (``line of text'') is by far the most frequent
classifier, appearing in 29{,}707 captions (39.6\%). Together with
\textit{tấm} (``flat surface object'') at 17{,}418 occurrences,
these two scene-text-specific classifiers account for over 63\% of
all classifier usage. This distribution is highly domain-specific:
in general Vietnamese text, \textit{cái} and \textit{con} are the
most frequent classifiers. The dominance of \textit{dòng} and
\textit{tấm} reflects the centrality of text-bearing objects
(signboards, banners, books) in ViTextCaps, and has direct
implications for caption generation models, which must learn to
produce grammatically correct classifier--noun combinations.

\subsubsection{Syntactic Dependency Structure}

Table~\ref{tab:dep_stats} summarizes the syntactic complexity of
2{,}000 randomly sampled captions parsed with Underthesea.

\begin{table}[ht]
\centering
\caption{Dependency structure statistics of ViTextCaps captions
(2{,}000 random samples). Captions exhibit moderate syntactic
complexity with short dependency distances, consistent with
Vietnamese's preference for local syntactic dependencies.}
\label{tab:dep_stats}
\begin{tabular}{lcc}
\toprule
\textbf{Metric} & \textbf{Mean} & \textbf{Std} \\
\midrule
Tokens per caption      & 19.4 & 13.4 \\
Tree depth              & 4.89 & 2.16 \\
Avg dependency distance & 2.39 & 0.87 \\
\bottomrule
\end{tabular}
\end{table}

The average tree depth of 4.89 indicates moderately complex sentence
structures---deeper than simple subject--verb--object constructions
but shallower than literary text. The short average dependency
distance (2.39 tokens) reflects Vietnamese's tendency toward local
syntactic dependencies, with modifiers typically adjacent to their
heads.

\paragraph{Root verb distribution.}

Table~\ref{tab:root_verbs} lists the ten most frequent syntactic
root verbs across the parsed sample.

\begin{table}[ht]
\centering
\caption{Top 10 root verbs in dependency trees of ViTextCaps
captions. The existential verb \textit{có} (``to have / there is'')
dominates at 28.2\%, reflecting Vietnamese descriptive sentence
patterns.}
\label{tab:root_verbs}
\begin{tabular}{llrr}
\toprule
\textbf{Root word} & \textbf{Meaning} & \textbf{Count} & \textbf{\%} \\
\midrule
có   & have / exist     & 564 & 28.2 \\
in   & print            &  81 &  4.0 \\
đặt  & place            &  69 &  3.5 \\
dòng & line (of text)   &  64 &  3.2 \\
chụp & photograph       &  57 &  2.9 \\
nằm  & lie / be located &  47 &  2.4 \\
treo & hang             &  47 &  2.4 \\
đứng & stand            &  33 &  1.7 \\
bán  & sell             &  30 &  1.5 \\
viết & write            &  30 &  1.5 \\
\bottomrule
\end{tabular}
\end{table}

The existential verb \textit{có} serves as syntactic root in 28.2\%
of captions (e.g., ``\textit{ảnh có dòng chữ...}''---``the image has
a line of text...''), characteristic of Vietnamese descriptive
sentences where \textit{có} functions as both a possessive verb and
an existential marker. The remaining top roots are action verbs
related to text display (\textit{in} ``print'', \textit{viết}
``write'', \textit{treo} ``hang'') and spatial description
(\textit{đặt} ``place'', \textit{nằm} ``lie''), reflecting the
dual visual--textual nature of scene-text captions.

\paragraph{Dependency relation distribution.}

The most frequent dependency relations are \texttt{nmod} (nominal
modifier, 14.4\%) and \texttt{compound} (11.4\%). The high
\texttt{compound} rate reflects the prevalence of Vietnamese
multi-word expressions in scene text: compound proper nouns
(``\textit{công ty TNHH}'', ``\textit{nhà xuất bản}''), compound
common nouns (``\textit{bảo tàng}'', ``\textit{cửa hàng}''), and
compound verbs (``\textit{chào mừng}'', ``\textit{quảng cáo}'').
The frequent \texttt{nummod} relation (4.8\%) corresponds to the
numerical content pervasive in scene text (phone numbers, addresses,
prices). Notably, \texttt{det:clf} (determiner--classifier) appears
in 3.9\% of all dependency relations, confirming that classifier
constructions are a syntactically significant feature that caption
generation models must learn to produce correctly.

\clearpage
\section{Experimental Setup Details}
\label{app:setup}

\subsection{Feature Extraction}
\label{sec:features}

All models in our comparison share the same visual and OCR
feature extractors, both fixed during training, ensuring
that performance differences are attributable solely to
fusion architecture rather than input representation.

\paragraph{Visual features.}
Visual region features are extracted using a Faster R-CNN
detector with a VinVL backbone~\cite{zhang2021vinvl},
which jointly models object--attribute--relationship
semantics through cross-modal pre-training. For each image,
the detector produces $N_v$ region proposals with
2048-dimensional feature vectors and bounding boxes
$\mathbf{b}^v_i \in \mathbb{R}^4$:
\begin{equation}
x^v_i \in \mathbb{R}^{2048}, \quad
\mathbf{b}^v_i = (c^x_i, c^y_i, w_i, h_i) \in \mathbb{R}^4
\end{equation}
Both HSTFG and PhonoSTFG project these to $d=768$ via
$\mathbf{v}_i = \mathrm{LN}(W_v x^v_i) +
\mathrm{LN}(W_b \mathbf{b}^v_i)$, matching the hidden
dimension of the MMT decoder.

\paragraph{OCR features.}
Scene text is detected and recognized by
SwinTextSpotter~\cite{swintextspotter}, which produces for each
detection $j$: a text string $s_j$, a 256-dimensional
recognition feature vector $r_j \in \mathbb{R}^{256}$
encoding character-level visual appearance, a 256-dimensional
detection feature vector $d_j \in \mathbb{R}^{256}$ encoding
the text region appearance, a bounding box
$\mathbf{b}^t_j \in \mathbb{R}^4$, and a scalar confidence
score $c_j \in [0,1]$.

For \textbf{M4C and HSTFG}, the OCR token representation
follows~\cite{hu2020iterative}: the string $s_j$ is encoded
via a pre-trained FastText embedding $e^{\mathrm{ft}}_j
\in \mathbb{R}^{300}$, which is then concatenated with recognition
and detection features:
\begin{equation}
x^t_j = [e^{\mathrm{ft}}_j\,;\,r_j\,;\,d_j]
  \in \mathbb{R}^{812}
\label{eq:ocr_feat_m4c}
\end{equation}
and projected to $d=768$ via $\mathbf{t}_j =
\mathrm{LN}(W_t x^t_j) + \mathrm{LN}(W_b \mathbf{b}^t_j)$.

For \textbf{PhonoSTFG}, the FastText embedding is replaced
by a dual-stream architecture that captures both visual OCR
appearance and Vietnamese linguistic context
(Section~\ref{sec:phonostfg}): the recognition and detection
streams are L2-normalized and projected jointly as the visual
stream, while the PhoBERT encoder provides the linguistic
stream. A learned gate dynamically balances the two streams
per dimension.

\subsection{Word Segmentation Analysis}
\label{sec:word_segmentation}

\begin{table*}[ht]
\centering
\caption{Impact of word segmentation on evaluation metrics.
The same predictions are scored with five tokenizers applied
to both hypotheses and references. $\Delta$ is the maximum
absolute difference across tokenizers per metric. Relative
variance ($\Delta / \max$) reaches 76\% for BLEU-4,
demonstrating that tokenizer choice can alter scores by more
than the typical improvement gap between model generations.}
\label{tab:segmentation_impact}
\resizebox{\textwidth}{!}{
\small
\begin{tabular}{lccccc|c}
\toprule
\textbf{Model / Metric} & \textbf{Space} & \textbf{PyVi}
  & \textbf{Underthesea} & \textbf{Character}
  & \textbf{Syllable} & $\boldsymbol{\Delta}$ \\
\midrule
\multicolumn{7}{l}{\textit{PhonoSTFG}} \\
\quad BLEU-1  & 0.2513 & 0.2331 & 0.2298
  & \textbf{0.5246} & 0.2517 & 0.2948 \\
\quad BLEU-4  & 0.0976 & 0.0718 & 0.0664
  & \textbf{0.2638} & 0.0979 & 0.1974 \\
\quad CIDEr   & \textbf{0.6462} & 0.5781 & 0.5382
  & 0.4146 & 0.6492 & 0.2345 \\
\quad ROUGE-L & 0.2266 & 0.2090 & 0.2042
  & \textbf{0.3788} & 0.2268 & 0.1746 \\
\addlinespace
\multicolumn{7}{l}{\textit{HSTFG}} \\
\quad BLEU-1  & 0.2218 & 0.2019 & 0.1973
  & \textbf{0.4744} & 0.2221 & 0.2771 \\
\quad BLEU-4  & 0.0848 & 0.0607 & 0.0561
  & \textbf{0.2373} & 0.0852 & 0.1812 \\
\quad CIDEr   & 0.6399 & 0.5681 & 0.5268
  & 0.4389 & \textbf{0.6432} & 0.2043 \\
\quad ROUGE-L & 0.2194 & 0.2003 & 0.1948
  & \textbf{0.3726} & 0.2197 & 0.1779 \\
\addlinespace
\multicolumn{7}{l}{\textit{M4C}} \\
\quad BLEU-1  & 0.2439 & 0.2266 & 0.2221
  & \textbf{0.5178} & 0.2443 & 0.2957 \\
\quad BLEU-4  & 0.0868 & 0.0646 & 0.0600
  & \textbf{0.2535} & 0.0871 & 0.1935 \\
\quad CIDEr   & 0.5943 & 0.5311 & 0.4879
  & 0.4075 & \textbf{0.5978} & 0.1904 \\
\quad ROUGE-L & 0.2207 & 0.2031 & 0.1980
  & \textbf{0.3734} & 0.2210 & 0.1754 \\
\bottomrule
\end{tabular}
}
\end{table*}

Vietnamese is an analytic language in which multi-syllable words
are written with spaces between syllables: \textit{thương mại}
(commerce) is written as two space-separated tokens, unlike
English \textit{commerce}. Word segmentation---the task of
grouping syllables into words---is therefore a non-trivial
preprocessing step that directly affects how evaluation metrics
count $n$-gram matches. Unlike English, where whitespace
tokenization reliably produces words, Vietnamese requires an
explicit segmentation model, and different segmenters produce
different word boundaries.

We demonstrate that this ambiguity has a severe impact on metric
reliability by evaluating the same model predictions under five
tokenization strategies applied uniformly to both hypotheses
and references:
\begin{enumerate}[nosep]
    \item \textbf{Space split}---whitespace tokenization
    (syllable-level; the current default in most Vietnamese
    captioning work).
    \item \textbf{PyVi}---rule-based Vietnamese word
    segmenter~\cite{pyvi}.
    \item \textbf{Underthesea}---neural Vietnamese NLP
    toolkit~\cite{underthesea}.
    \item \textbf{Character}---each character as a token
    (language-agnostic baseline).
    \item \textbf{Syllable}---space split with Unicode
    normalization and punctuation removal.
\end{enumerate}

Table~\ref{tab:segmentation_impact} reports the results across
three representative models.

BLEU-4 varies by up to \textbf{76\% relative to its maximum
value}: for HSTFG, BLEU-4 ranges from 0.0561 (Underthesea) to
0.2373 (character-level)---a 4.2$\times$ difference from
identical predictions. Character-level tokenization inflates
BLEU through trivially shared substrings while simultaneously
deflating CIDEr, confirming that no single metric is robust
to tokenization granularity. Word-level segmenters (PyVi,
Underthesea) disagree by 0.04--0.05 CIDEr absolute points
due to different compound word boundary decisions. Space split
and syllable produce near-identical results, confirming that
prior Vietnamese captioning work operates at the syllable level.

\begin{figure*}[htbp]
  \centering
  \includegraphics[width=\textwidth]{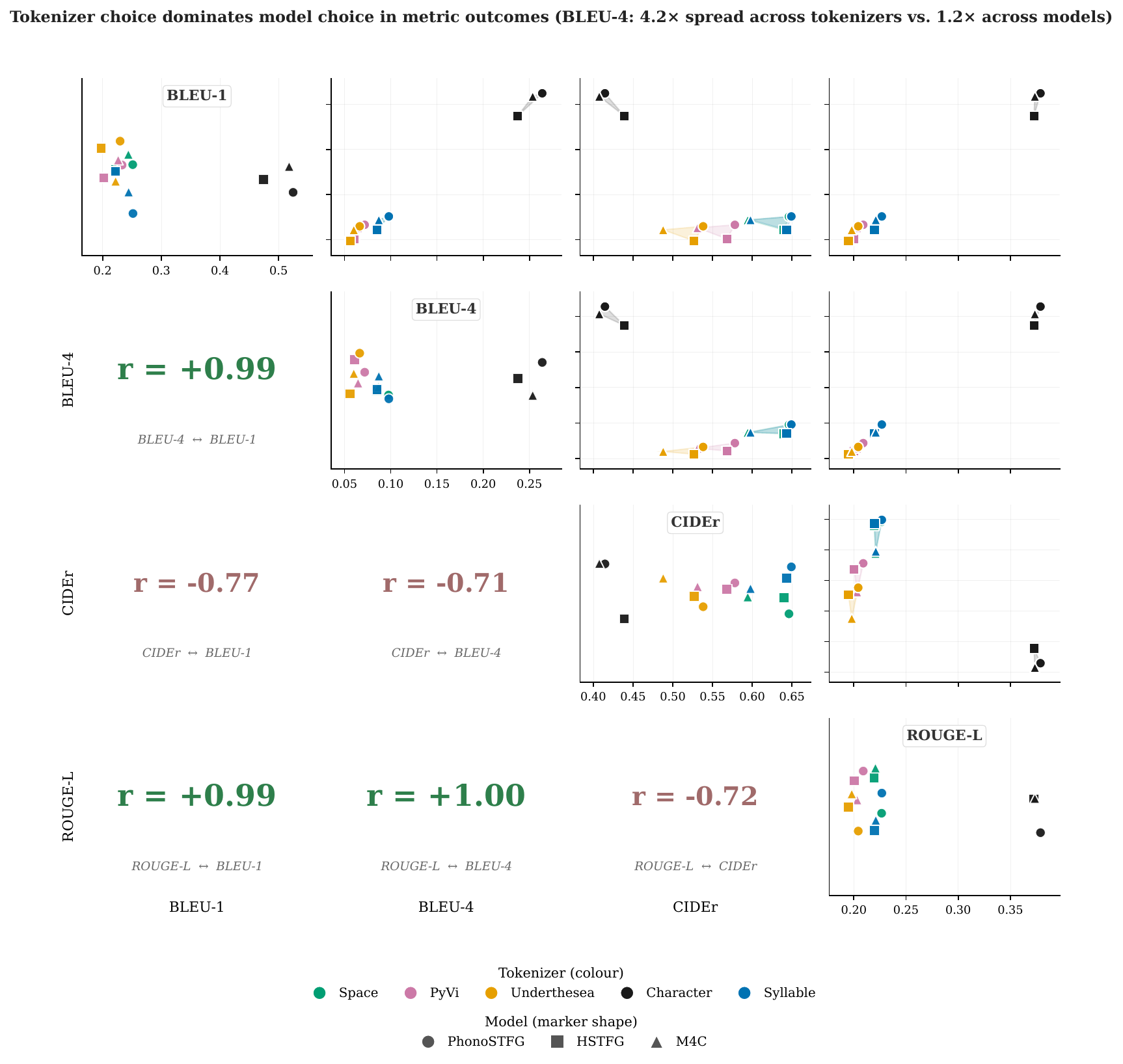}
  \caption{%
    Scatter-plot matrix (SPLOM) of 4 metrics $\times$ 15
    model--tokenizer configurations (Table~\ref{tab:segmentation_impact}).
    Upper triangle: scatter plots with per-tokenizer convex hulls
    showing that tokenizer choice clusters configurations more
    tightly than model choice (4.2$\times$ metric spread across
    tokenizers vs.\ 1.2$\times$ across models).
    Lower triangle: Pearson correlations.
    Diagonal: marginal score distributions per metric.
    Space-split and syllable tokenizers (overlapping hulls) produce
    near-identical scores, confirming the benchmark choice.%
  }
  \label{fig:segmentation_splom}
\end{figure*}

Figure~\ref{fig:segmentation_splom} makes this tokenizer-dominance effect geometrically explicit
across all four metrics simultaneously.

Based on these findings, we adopt \textbf{syllable-level
(space-split) tokenization} as the ViTextCaps benchmark
standard. This choice requires no external tool (ensuring
reproducibility), operates at a linguistically meaningful
granularity (Vietnamese syllables are the minimal prosodic
unit), and avoids inter-segmenter disagreement. CIDEr is
adopted as the \textbf{primary comparison metric}, owing to
its lowest relative variance across tokenizers (31--36\%)
and its sensitivity to image-discriminative scene-text tokens
via TF-IDF weighting.
\clearpage
\section{Additional Experimental Analysis}
\label{app:exp_ana}

\subsection{Result Analysis}
\label{sec:result_analysis}

\subsubsection{Stratified Performance Analysis}
\label{app:stratified}

We evaluate model performance on test subsets defined by three
image properties---OCR confidence, text density, and language
profile---to characterize when graph-based fusion provides the
largest gains.

\begin{table}[htbp]
\centering
\caption{CIDEr stratified by OCR confidence (F2), text density (F3), and language profile (F4).}
\label{tab:stratified}
\resizebox{\textwidth}{!}{%
\begin{tabular}{llcccc}
\toprule
\textbf{Factor} & \textbf{Stratum} & \textbf{$n$} & \textbf{PhonoSTFG} & \textbf{HSTFG} & \textbf{M4C} \\
\midrule
\multirow{3}{*}{F2: OCR Conf.}
  & Low ($<$0.5)      & 1{,}728  & 0.336 & 0.326 & 0.345 \\
  & Medium (0.5--0.8) & 10{,}822 & 0.612 & 0.608 & 0.575 \\
  & High ($\geq$0.8)  & 2{,}178  & \textbf{0.988} & 0.965 & 0.823 \\
\midrule
\multirow{3}{*}{F3: Text Density}
  & Sparse ($\leq$5)  & 1{,}568 & 0.732 & 0.726 & 0.731 \\
  & Medium (6--20)    & 4{,}999 & 0.687 & 0.679 & 0.626 \\
  & Dense ($>$20)     & 8{,}161 & 0.581 & 0.577 & 0.528 \\
\midrule
\multirow{2}{*}{F4: Language}
  & Pure Vietnamese   & 8{,}386 & 0.704 & 0.703 & 0.650 \\
  & Mixed VN+EN       & 6{,}039 & 0.554 & 0.544 & 0.503 \\
\bottomrule
\end{tabular}%
}
\end{table}

OCR confidence shows the largest variation in CIDEr ($2.9\times$
range for PhonoSTFG: $0.336$ to $0.988$), followed by text density
($1.3\times$) and language profile ($1.3\times$). Three patterns
emerge from Table~\ref{tab:stratified}.

The PhonoSTFG advantage is \textbf{largest at high OCR confidence}.
The gap between PhonoSTFG and M4C grows from near-zero at low
confidence ($0.336$ vs.\ $0.345$, M4C slightly leads) to $+0.165$
at high confidence ($0.988$ vs.\ $0.823$). This is consistent with
the phonological bias being most effective when OCR tokens are
reliable enough to serve as anchors for diacritic disambiguation;
at low confidence, noisy OCR undermines phonological reasoning
regardless of architecture.

\textbf{All text-aware models degrade on dense images.}
CIDEr drops by $0.151$ for PhonoSTFG (sparse to dense), $0.149$
for HSTFG, and $0.203$ for M4C. M4C shows a larger relative drop
($27.8\%$) than PhonoSTFG ($20.6\%$), suggesting that graph-based
inter-token reasoning partially mitigates the text selection problem
on dense images. Dense images constitute $54\%$ of the test set,
making this the primary difficulty gradient for ViTextCaps.

\textbf{Mixed-language images are consistently harder.}
PhonoSTFG drops from $0.704$ (pure Vietnamese) to $0.554$ (mixed,
$21.3\%$ relative decrease), consistent with the code-mixing
analysis (Section~\ref{sec:code_mixing}): $42.9\%$ of images
contain both Vietnamese and English text, and the phonological bias
is only activated for Vietnamese syllables ($p_8 = 1$), providing
no benefit for English tokens.

\subsubsection{Domain and Visual Complexity Analysis}
\label{sec:domain_eval}

\begin{table*}[htbp]
\centering
\caption{Per-domain CIDEr scores and image characteristics, sorted by PhonoSTFG performance. \textit{Density}: mean OCR tokens per image. \textit{Conf.}: mean OCR confidence. \textbf{Bold}: best CIDEr per domain.}
\label{tab:domain_performance}
\resizebox{\textwidth}{!}{%
\begin{tabular}{l rrc ccc}
\toprule
\textbf{Domain} & \textbf{$n$} & \textbf{Density} & \textbf{Conf.} & \textbf{PhonoSTFG} & \textbf{HSTFG} & \textbf{M4C} \\
\midrule
Book/Publication  & 2{,}097 & 35.4 & 0.67 & 0.749 & \textbf{0.769} & 0.702 \\
Service           & 453     & 35.9 & 0.70 & \textbf{0.704} & 0.667 & 0.531 \\
Education         & 852     & 36.0 & 0.69 & 0.645 & \textbf{0.708} & 0.690 \\
Product/Packaging & 637     & 37.5 & 0.65 & \textbf{0.598} & 0.529 & 0.589 \\
Shop/Store        & 4{,}357 & 33.8 & 0.63 & \textbf{0.588} & 0.525 & 0.486 \\
Advertisement     & 263     & 34.7 & 0.66 & 0.536 & \textbf{0.583} & 0.528 \\
Street/Outdoor    & 1{,}944 & 23.1 & 0.65 & 0.504 & \textbf{0.576} & 0.484 \\
Food/Restaurant   & 2{,}369 & 33.0 & 0.67 & 0.500 & 0.504 & \textbf{0.515} \\
Fashion/Clothing  & 1{,}312 & 24.5 & 0.64 & \textbf{0.470} & 0.377 & 0.428 \\
\bottomrule
\end{tabular}%
}
\end{table*}

Table~\ref{tab:domain_performance} reveals that no single model
dominates across all domains, with wins distributed between
PhonoSTFG (5 domains) and HSTFG (3 domains). The split follows a
domain-type pattern rather than a simple confidence gradient:
HSTFG leads on domains with formal, structured text (Book,
Education, Street/Outdoor, Advertisement), while PhonoSTFG leads
on domains with informal commercial text (Shop/Store, Fashion,
Product/Packaging, Service). Formal text---book titles, school
signs, road signage---is typically printed in standard fonts with
minimal diacritic ambiguity, making the additional visual context
from HSTFG's V$\leftrightarrow$T edges beneficial. Commercial
text, by contrast, frequently uses creative fonts and color-heavy
designs that degrade OCR diacritic accuracy, precisely the
condition where PhonoSTFG's phonological bias provides the most
value. Food/Restaurant is the only domain where M4C matches
graph-based models (CIDEr $0.515$ vs.\ $0.504$/$0.500$), possibly
because menu text---prices and dish names---follows rigid
formatting that flat attention handles adequately without
linguistic specialization.

\begin{table}[htbp]
\centering
\caption{CIDEr by number of domains per image. Images with more
domains have higher text density and lower OCR confidence.}
\label{tab:ndomain}
\small
\begin{tabular}{l rrr ccc}
\toprule
\textbf{\#Dom.} & \textbf{$n$}
  & \textbf{Density} & \textbf{Conf.}
  & \textbf{PhonoSTFG} & \textbf{HSTFG} & \textbf{M4C} \\
\midrule
1  & 2{,}080 & 15.5 & 0.68 & 0.881 & \textbf{1.011} & 0.865 \\
2  & 3{,}056 & 20.9 & 0.66 & \textbf{0.680} & 0.629 & 0.615 \\
3  & 3{,}239 & 27.4 & 0.64 & \textbf{0.647} & 0.629 & 0.583 \\
4+ & 6{,}353 & 44.3 & 0.63 & \textbf{0.511} & 0.489 & 0.468 \\
\bottomrule
\end{tabular}
\end{table}

CIDEr decreases monotonically with domain complexity: from $0.881$
(single-domain) to $0.511$ (4+ domains) for PhonoSTFG---a $42\%$
relative drop---co-occurring with increasing text density ($15.5$
to $44.3$ tokens/image) and decreasing OCR confidence ($0.68$ to
$0.63$).
The single-domain subset reveals a notable reversal: HSTFG
substantially outperforms PhonoSTFG ($1.011$ vs.\ $0.881$, gap
$= 0.130$). Single-domain images are characterised by low text
density ($15.5$ tokens/image) and high OCR confidence ($0.68$),
meaning that scene text is sparse and reliably recognised. In
this regime, HSTFG's V$\leftrightarrow$T graph edges---removed in
PhonoSTFG---provide additional visual context that proves valuable
when few OCR tokens are available to reason over. Conversely, the
linguistic components added by PhonoSTFG (PhoBERT, phonological
bias) offer little benefit when OCR is already accurate, and the
absent V$\leftrightarrow$T edges represent a net loss of signal.
PhonoSTFG recovers and leads from 2+ domains onward, where text
density increases and OCR diacritic errors become more prevalent.
This analysis suggests a practical deployment guideline: HSTFG is
preferable for document-like images with sparse, high-confidence
text (e.g., book covers, formal signage), while PhonoSTFG provides
superior performance on the commercially dominant case of complex,
multi-source Vietnamese scene text.

\begin{figure*}[htbp]
  \centering
  \includegraphics[width=\textwidth]{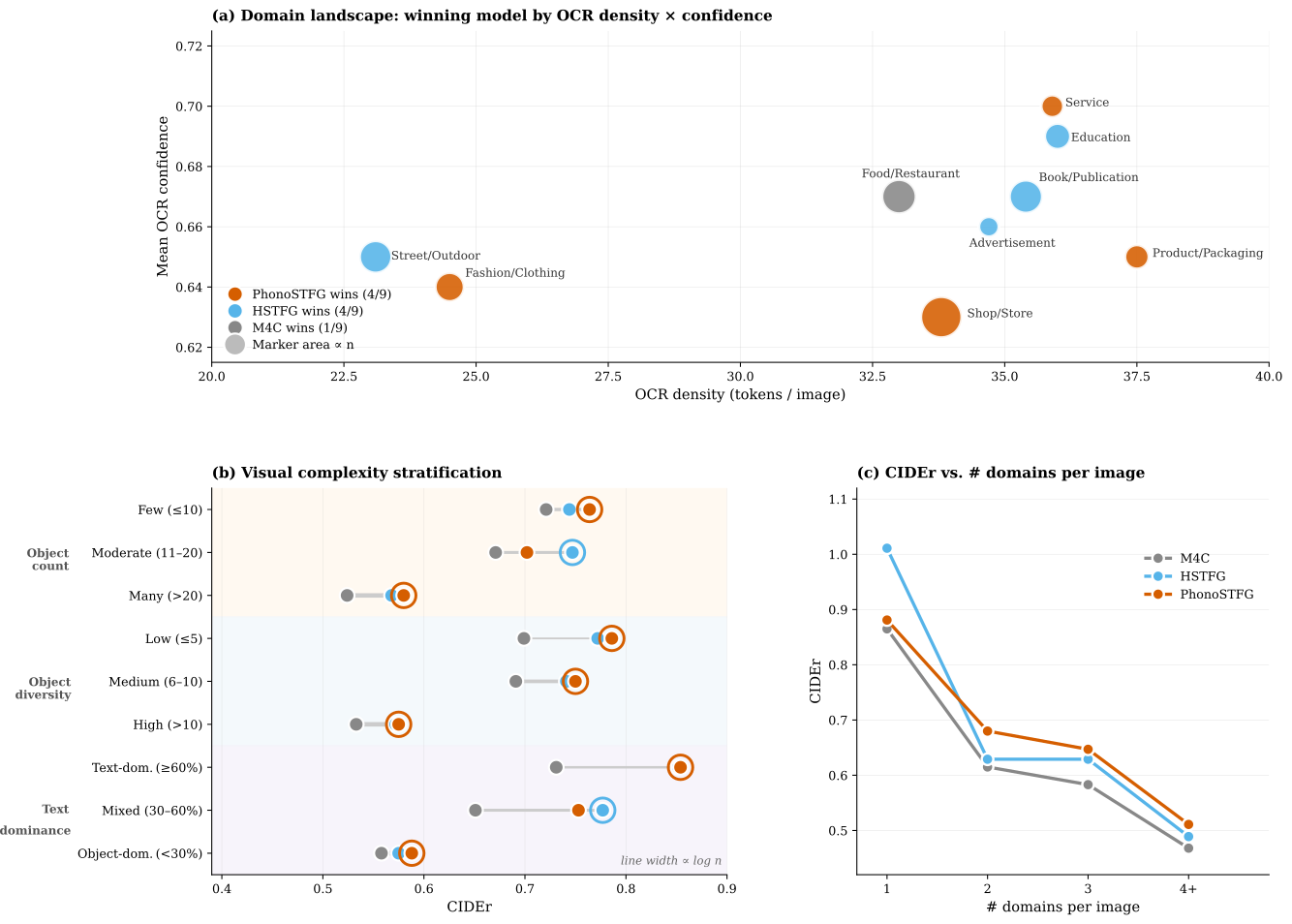}
  \caption{%
    CIDEr performance across scene-text complexity dimensions.
    \textbf{(a)} Domain scatter: OCR density vs.\ confidence with
    bubble area proportional to image count and color indicating the
    winning model per domain.
    \textbf{(b)} Visual-complexity dumbbell plot: 9 strata grouped by
    object count, object diversity, and text-object dominance---showing
    per-model CIDEr and the PhonoSTFG advantage at high text dominance.
    \textbf{(c)} CIDEr by number of distinct domains per image, showing
    HSTFG parity at single-domain and PhonoSTFG recovery from two
    domains onward as text density and diacritic errors increase.%
  }
  \label{fig:complexity_landscape}
\end{figure*}

Figure~\ref{fig:complexity_landscape} consolidates all three complexity stratifications into a single panel.

\begin{table}[htbp]
\centering
\caption{CIDEr stratified by visual complexity. Text dominance
measures the fraction of detected objects that are text-related.}
\label{tab:visual_complexity}
\small
\resizebox{\textwidth}{!}{
\begin{tabular}{llr ccc}
\toprule
\textbf{Dimension} & \textbf{Stratum}
  & \textbf{$n$}
  & \textbf{PhonoSTFG} & \textbf{HSTFG} & \textbf{M4C} \\
\midrule
\multirow{3}{*}{Object count}
  & Few ($\leq$10)    & 2{,}809 & 0.764 & 0.744 & 0.721 \\
  & Moderate (11--20) & 2{,}205 & 0.702 & \textbf{0.747} & 0.671 \\
  & Many ($>$20)      & 9{,}714 & \textbf{0.580} & 0.568 & 0.524 \\
\midrule
\multirow{3}{*}{Object diversity}
  & Low ($\leq$5)   &    680  & \textbf{0.786} & 0.772 & 0.699 \\
  & Medium (6--10)  &  4{,}178 & \textbf{0.750} & 0.741 & 0.691 \\
  & High ($>$10)    &  9{,}870 & \textbf{0.575} & 0.572 & 0.533 \\
\midrule
\multirow{3}{*}{Text dominance}
  & Text-dom.\ ($\geq$60\%) & 1{,}239  & \textbf{0.854} & 0.852 & 0.731 \\
  & Mixed (30--60\%)        & 2{,}167  & 0.753 & \textbf{0.777} & 0.651 \\
  & Object-dom.\ ($<$30\%)  & 11{,}322 & \textbf{0.588} & 0.575 & 0.558 \\
\bottomrule
\end{tabular}
}
\end{table}

Performance decreases consistently with visual complexity across
all three dimensions. Text-dominant images ($\geq 60\%$ text
objects) are the easiest subset for all models (CIDEr
$0.731$--$0.854$), as the captioning task aligns closely with
text reading. The PhonoSTFG advantage is largest on text-dominant
images ($+0.123$ over M4C) but only $+0.030$ on object-dominant
images, consistent with the T$\to$T attention design that focuses
on inter-text relationships. Object-dominant images ($<30\%$ text
objects), constituting $76.8\%$ of the test set, remain the hardest
subset for all models.

\subsubsection{Error Propagation}
\label{sec:error_propagation}

\begin{figure}[htbp]
  \centering
  \includegraphics[width=\linewidth]{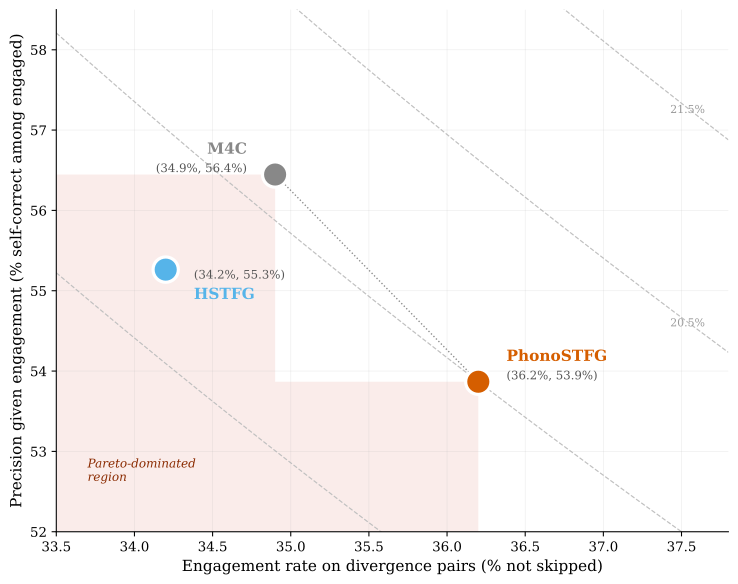}
  \caption{%
    Pareto scatter of engagement rate vs.\ precision on
    $9{,}066$ diacritic-divergence pairs for three models.
    Dashed iso-lines show constant self-correct rates; the shaded
    L-region marks Pareto-dominated configurations.
    All models cluster near the same engagement--precision
    frontier, indicating that higher engagement produces both
    more self-corrections \emph{and} more copy errors proportionally.
    PhonoSTFG shifts slightly toward higher engagement, reflecting
    the phonological bias encouraging the model to commit to a
    diacritic rather than skipping.%
  }
  \label{fig:error_propagation}
\end{figure}

Figure~\ref{fig:error_propagation} plots the engagement--precision
Pareto frontier for all three models on the $9{,}066$
diacritic-divergence pairs identified in the test set.

Sections~\ref{sec:tone_diacritic_ambiguity}--\ref{sec:ocr_error_taxonomy}
established that OCR diacritic errors are pervasive and structured.
We now characterize what models do when OCR output disagrees with
ground truth on diacritics, using the $9{,}066$ caption--OCR
divergence pairs identified on the test set
(Section~\ref{sec:diacritic_divergence}). Each model's behavior
per divergence pair is classified as: \textbf{copy error} (output
contains OCR token with incorrect diacritic), \textbf{self-correct}
(output contains reference token with correct diacritic),
\textbf{skip} (output contains neither), or \textbf{other variant}
(base form appears with a third diacritical variant).

\begin{table}[t]
\centering
\caption{Model behavior on $9{,}066$ diacritic divergence pairs
(test set). Models predominantly skip ambiguous tokens rather
than risking diacritic errors.}
\label{tab:error_propagation}
\small
\begin{tabular}{lccc}
\toprule
\textbf{Behavior}
  & \textbf{PhonoSTFG} & \textbf{HSTFG} & \textbf{M4C} \\
\midrule
Copy error    & 13.2\% & 12.1\% & 11.6\% \\
Self-correct  & 19.5\% & 18.9\% & 19.7\% \\
Skip          & \textbf{63.8\%} & \textbf{65.8\%} & \textbf{65.1\%} \\
Other variant &  3.5\% &  3.2\% &  3.6\% \\
\bottomrule
\end{tabular}
\end{table}

Table~\ref{tab:error_propagation} reveals a striking pattern: all
models skip approximately two-thirds of divergence-prone tokens,
choosing to omit ambiguous text rather than attempting reproduction.
When models do engage, they self-correct more often than they copy
errors (19--20\% vs.\ 12--13\%), indicating that the language model
prior provides partial diacritic correction. The behavior
distribution is consistent across all three architectures,
suggesting this is a property of the task and training data rather
than a model-specific phenomenon.

The copy error rates nonetheless reveal a meaningful contrast in
copying strategy. PhonoSTFG adopts a more aggressive approach: its
higher OTR ($0.399$ vs.\ $0.377$ for HSTFG) indicates greater OCR
token coverage, but this comes at the cost of a higher copy error
rate ($13.2\%$ vs.\ $12.1\%$) and higher THR ($0.244$ vs.\
$0.230$). HSTFG, by contrast, copies more selectively---producing
fewer text references but with higher fidelity. This
precision--coverage trade-off between the two models is internally
consistent across three independent measures (OTR, copy error rate,
THR), providing converging evidence that the architectural
differences between HSTFG and PhonoSTFG manifest as a copying
strategy difference rather than an overall quality difference.

\begin{table}[t]
\centering
\caption{Copy error rate stratified by OCR confidence.
High-confidence errors are propagated $5\times$ more frequently
than low-confidence ones.}
\label{tab:copy_error_by_conf}
\small
\begin{tabular}{lccc}
\toprule
\textbf{Confidence}
  & \textbf{PhonoSTFG} & \textbf{HSTFG} & \textbf{M4C} \\
\midrule
Low ($<$0.5)      &  4.4\% &  4.0\% &  3.9\% \\
Medium (0.5--0.8) &  9.1\% &  7.5\% &  7.5\% \\
High ($\geq$0.8)  & \textbf{20.6\%} & \textbf{19.6\%} & \textbf{18.5\%} \\
\bottomrule
\end{tabular}
\end{table}

Table~\ref{tab:copy_error_by_conf} reveals that the copy error rate
increases $5$-fold from low to high OCR confidence: PhonoSTFG
propagates errors for only $4.4\%$ of low-confidence divergences
but $20.6\%$ of high-confidence ones. This is the behavioral
consequence of the confidence--error type interaction identified in
Section~\ref{sec:error_by_confidence}: high-confidence tokens are
more likely to contain tone substitution errors (T2), and models
trust confident-but-wrong tokens into the output. \textbf{OCR
confidence is therefore not a reliable proxy for correctness} in
Vietnamese scene text---a naive strategy of trusting high-confidence
tokens would paradoxically increase error propagation, further
motivating graph-based fusion that weighs linguistic plausibility
alongside confidence.



The 65\% skip rate represents substantial information loss: models
avoid two-thirds of diacritic-ambiguous tokens, meaning much
scene-text content is never surfaced in generated captions.
Reducing this skip rate---by improving models' ability to handle
diacritic ambiguity---remains a clear avenue for future improvement.


\subsection{Qualitative Analysis}
\label{app:qualitative}

\subsubsection{Error Analysis}
\label{sec:error_analysis}

\begin{table}[htbp]
\centering
\caption{Error type distribution of PhonoSTFG on the test set
($14{,}728$ predictions). Categories are not mutually exclusive.}
\label{tab:error_distribution}
\small
\begin{tabular}{lrr}
\toprule
\textbf{Error Type} & \textbf{Count} & \textbf{\%} \\
\midrule
Missing important text     & 5{,}861 & 39.8\% \\
Very low CIDEr ($<$0.05)  & 4{,}146 & 28.2\% \\
Repetition                 & 2{,}652 & 18.0\% \\
Diacritic copy error       & 2{,}301 & 15.6\% \\
Text hallucination         &   558   &  3.8\% \\
\midrule
No detected errors         & 1{,}453 &  9.9\% \\
\bottomrule
\end{tabular}
\end{table}

The most frequent error is \textbf{missing important text} (39.8\%):
the model fails to mention OCR tokens that appear in reference
captions. This is consistent with the mean text coverage rate of
35.9\% (Section~\ref{sec:text_dependency}), confirming that the text
selection problem remains largely unsolved for the best model.
\textbf{Repetition} (18.0\%) arises from autoregressive decoding
loops exacerbated by the copy mechanism when multiple similar OCR
tokens are present. \textbf{Diacritic copy errors} (15.6\%) are
consistent with the 13.2\% copy error rate from the error propagation
analysis, concentrated at high OCR confidence. \textbf{Text
hallucination} (3.8\%) is rare, confirming that the copy mechanism
effectively grounds text generation in OCR input. Only 9.9\% of
predictions have no detected errors, indicating substantial room for
improvement across all categories.

\subsubsection{Phonological Attention Visualization}
\label{sec:attention_viz}

\begin{figure}[htbp]
\centering
\includegraphics[width=\linewidth]{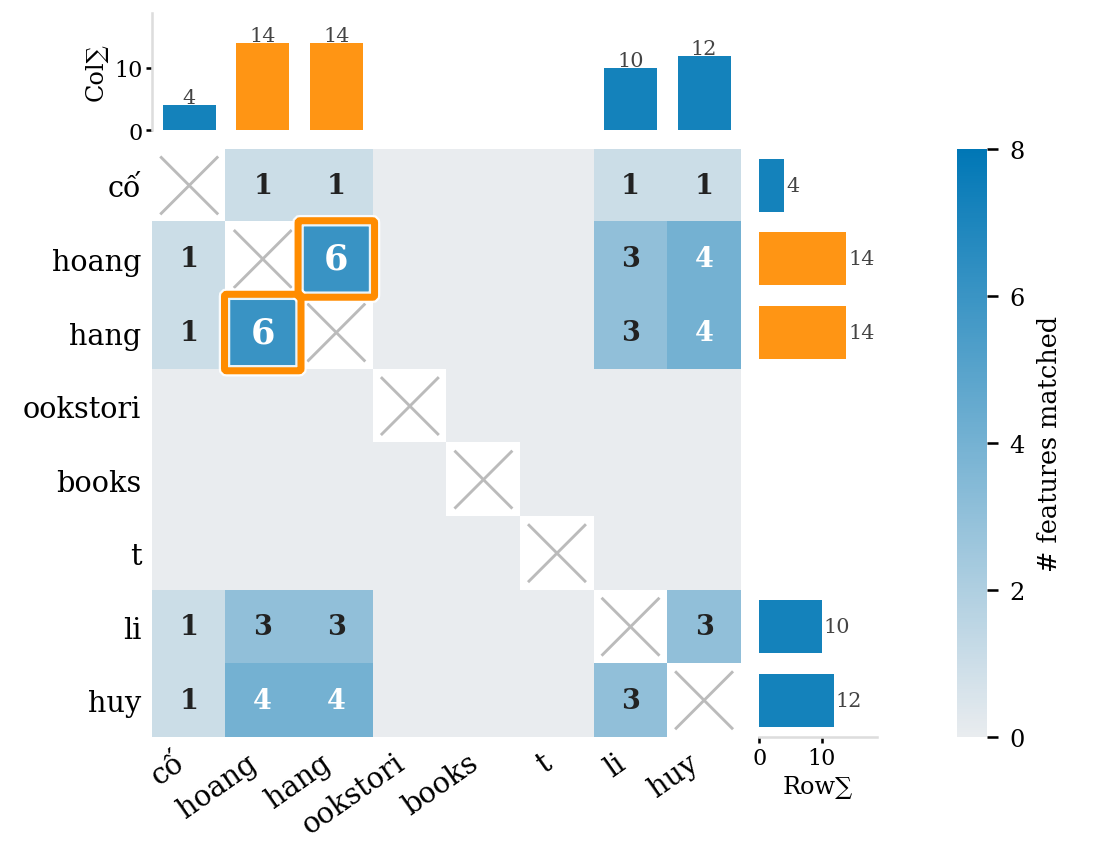}
\caption{Aggregate phonological bias visualization for PhonoSTFG on
image 526291. Circle area is proportional to the number of matched
features across all 8 pairwise bias features. The top phonologically
similar pair (\textit{hoang}\,$\leftrightarrow$\,\textit{hang},
orange) achieves 6/8 matched features and is annotated with its
feature breakdown.}
\label{fig:attention_viz_b}
\end{figure}

Figure~\ref{fig:attention_viz_b} visualizes the aggregate
phonological bias mechanism of PhonoSTFG on image~526291, which
contains eight OCR tokens including the phonologically similar pair
\textit{hoang}~$\leftrightarrow$~\textit{hang}.
The aggregate bias matrix sums all 8 pairwise features into a dot
matrix where circle area encodes the number of matched features.
The top pair receives a score of 6/8, reflecting matched onset,
nucleus, coda, rhyme, base-form, and bilingual membership, and is
highlighted in orange.

\begin{figure}[htbp]
\centering
\includegraphics[width=\linewidth]{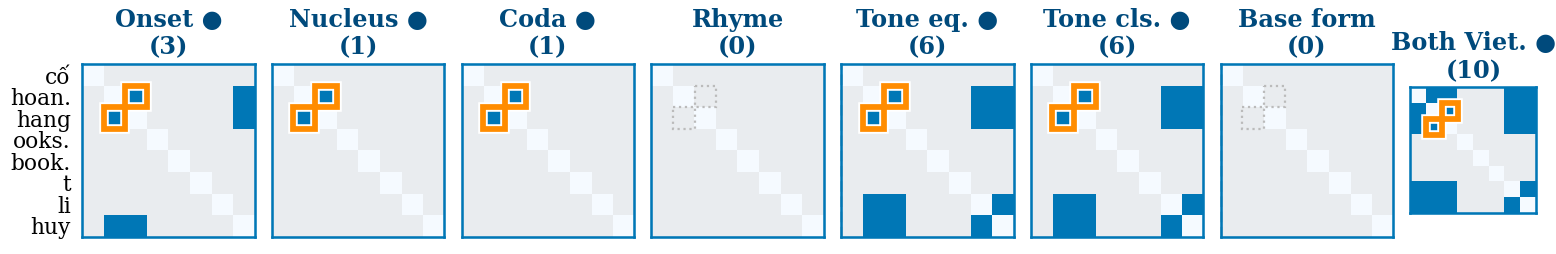}
\caption{Per-feature decomposition of phonological bias for
PhonoSTFG on image 526291. Features are grouped into three
linguistic classes: Phoneme, Tone, and Meta/Diacritics. Orange
dots~($\star$) indicate the features activated for the top pair
(\textit{hoang}\,$\leftrightarrow$\,\textit{hang}).}
\label{fig:attention_viz_c}
\end{figure}

Figure~\ref{fig:attention_viz_c} further decomposes this aggregate
bias into its eight constituent features grouped by linguistic class
(Phoneme, Tone, Meta/Diacritics). Orange dots~($\star$) mark the
features active for the top pair.

\subsubsection{Cross-model Qualitative Comparison}
\label{sec:cross_model}

\begin{figure}[t]
\centering
\includegraphics[width=\linewidth]{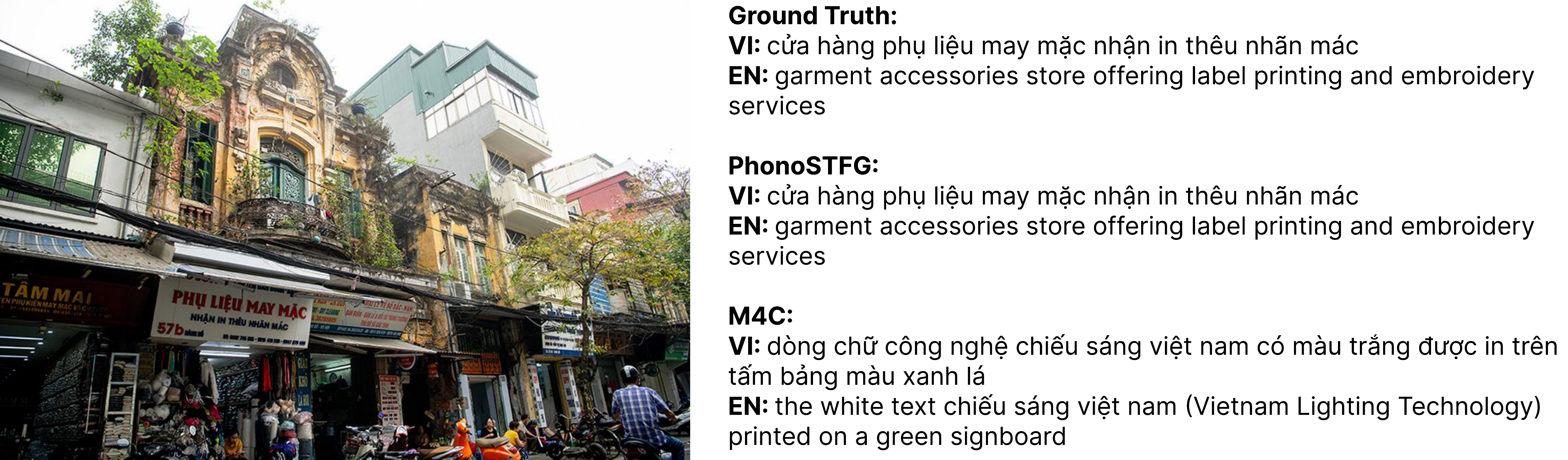}

\vspace{0.5em}

\includegraphics[width=\linewidth]{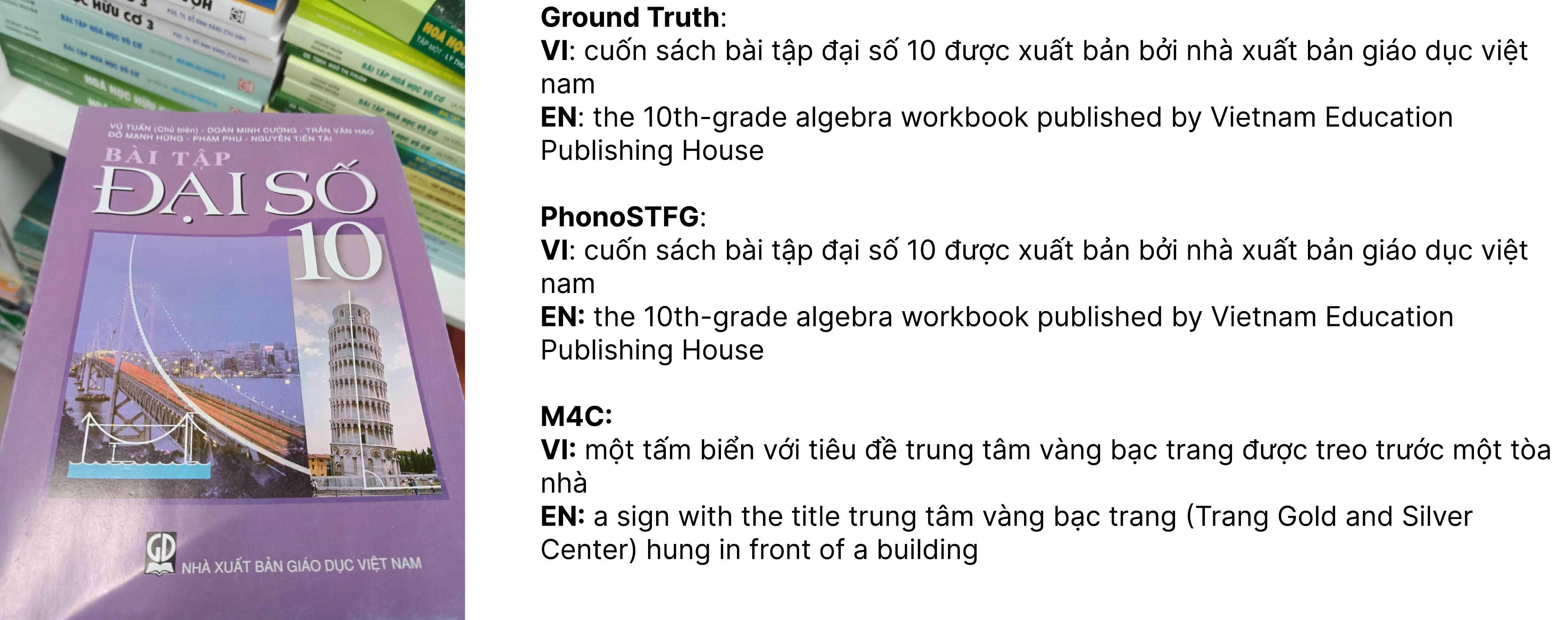}

\caption{Success cases of PhonoSTFG: the model produces captions
that fully align with the ground truth (CIDEr = 10.0), whereas
HSTFG (base) and M4C fail to capture the correct scene text.
The improvement is attributed to PhoBERT dual-stream embeddings
and phonological attention bias, which facilitate more precise
OCR token selection and effective fusion of textual and visual
information.}
\label{fig:qualitative_success}
\end{figure}

Figure~\ref{fig:qualitative_success} illustrates cases where
PhonoSTFG achieves perfect alignment with ground truth (CIDEr = 10.0).
PhonoSTFG demonstrates a superior ability to filter irrelevant OCR
tokens in complex scenes: in the garment store example, it correctly
identifies the business type (\textit{phụ liệu may mặc}) whereas
M4C suffers from noise-leakage, generating tokens from unrelated
signboards. This success reflects the PhoBERT-based dual-stream
embeddings and phonological attention bias enabling better
association of visual regions with semantically relevant text tokens.


\begin{figure*}[htbp]
    \centering
    \includegraphics[width=\textwidth]{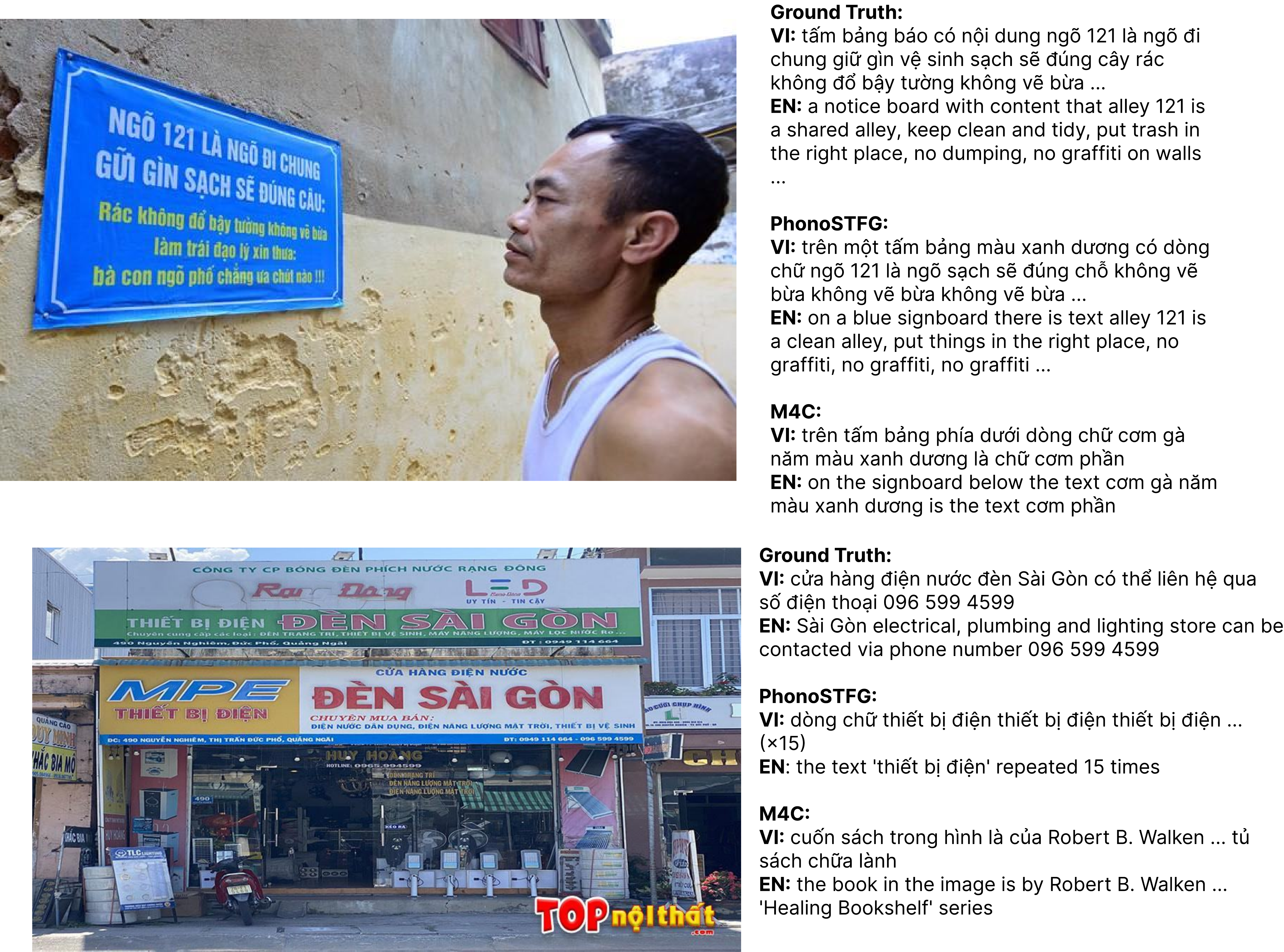}
    \caption{Representative failure cases of PhonoSTFG (Part 2). \textbf{(Top)} Diacritic propagation error: Enhanced reads the sign partially but copies OCR diacritics and enters repetition (‘không vẽ bừa’ $\times$ 3).  \textbf{(Bottom)} Repetition: Enhanced enters a decoding loop on ‘thiết bị điện’; all models fail to capture the store name and phone number.}
    \label{fig:qualitative_failure_1}
\end{figure*}

\begin{figure*}[htbp]
    \centering
    \includegraphics[width=\textwidth]{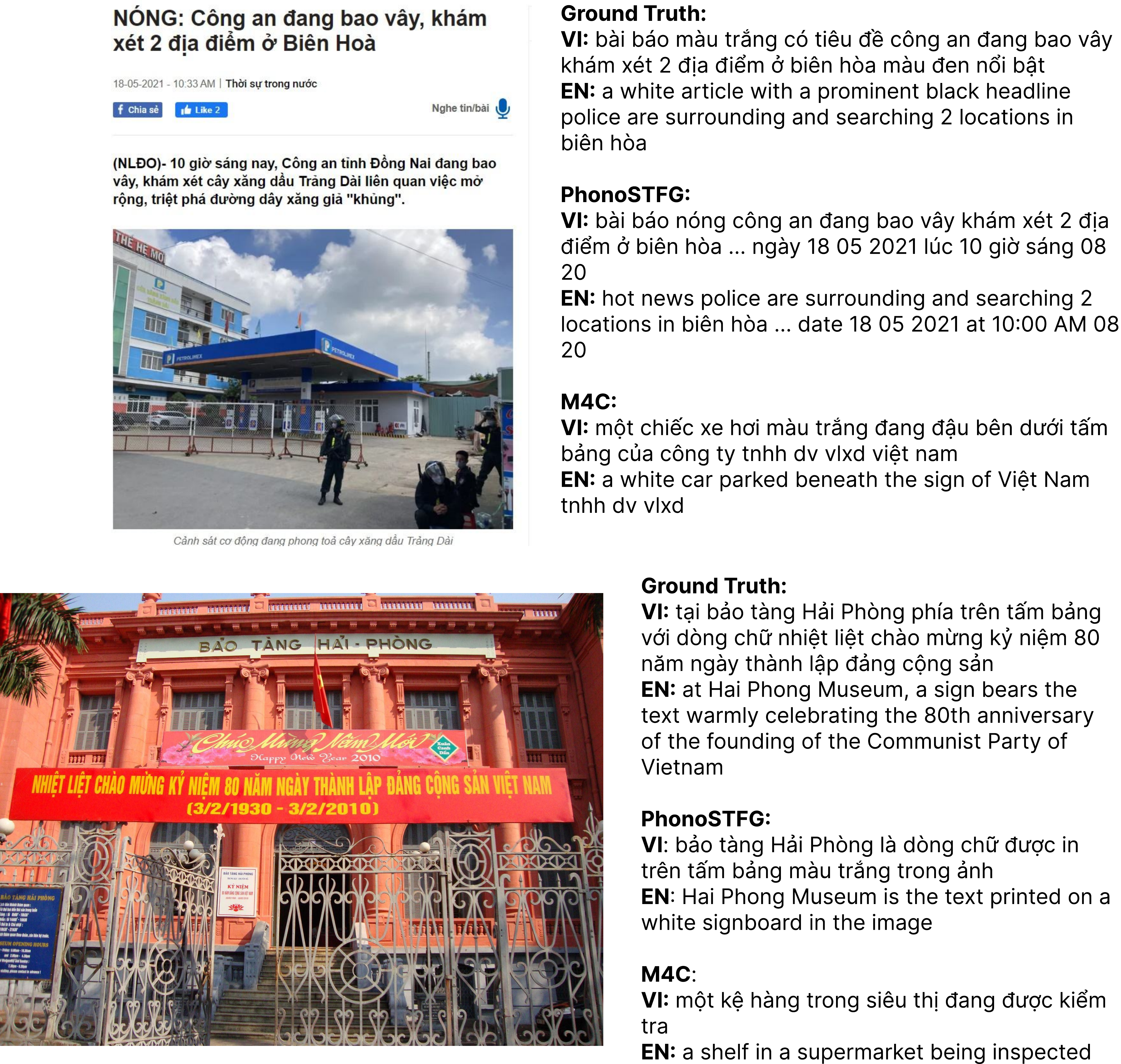}
    \caption{Representative failure cases of PhonoSTFG (Part 1). \textbf{(Top)} Hallucination: Enhanced correctly reads the headline but hallucinates dates (‘18 05 2021’) not present in OCR. \textbf{(Bottom)} Missing or incorrect OCR extraction: Models miss critical text (‘nhiệt liệt chào mừng’, ‘80 năm’); enhanced captures partial context (‘bảo tàng hải phòng’).}
    \label{fig:qualitative_failure_2}
\end{figure*}


Figure~\ref{fig:qualitative_failure_1} and Figure~\ref{fig:qualitative_failure_2} illustrate the four dominant failure patterns:

\begin{itemize}

\item \textbf{Diacritic propagation} (Fig.~\ref{fig:qualitative_failure_1}a): the model inherits incorrect diacritics from high-confidence OCR tokens and propagates them into the output, consistent with the 5$\times$ amplification of copy error rate at high confidence identified in Section~\ref{sec:error_propagation}. 

\item \textbf{Repetition} (Fig.~\ref{fig:qualitative_failure_1}b): the decoder enters a loop repeating \textit{thiết bị điện} 15 times, suggesting that the copy mechanism amplifies repetition when multiple similar OCR tokens are present. 

\item \textbf{Hallucination} (Fig.~\ref{fig:qualitative_failure_2}a): though rare (3.8\%), the model generates specific dates not present in the OCR stream, likely driven by linguistic priors in the pre-trained language model. \textbf{Missing OCR extraction} (Fig.~\ref{fig:qualitative_failure_2}b): even clearly legible text (e.g., \textit{Bảo tàng Hải Phòng}) may be omitted if OCR confidence is low or the text selection mechanism prioritizes other features.

\end{itemize}

\end{document}